\documentclass[10pt,twocolumn,letterpaper]{article}

\usepackage[pagenumbers]{iccv} 

\definecolor{iccvblue}{rgb}{0.21,0.49,0.74}
\usepackage[pagebackref,breaklinks,colorlinks,allcolors=iccvblue]{hyperref}
\newcommand\blfootnote[1]{%
  \begingroup
  \renewcommand\thefootnote{}\footnote{#1}%
  \addtocounter{footnote}{-1}%
  \endgroup
}
\usepackage{multirow}
\usepackage{multicol}
\usepackage{algorithm}
\usepackage{algorithmic}
\usepackage{dirtytalk}
\usepackage{breqn}
\usepackage{pifont}
\def\paperID{5983} 
\def\confName{ICCV}
\def\confYear{2025}

\title{Test-Time Modality Generalization for Medical Image Segmentation}

\author{Ju-Hyeon Nam \qquad Sang-Chul Lee$^{*}$ \\
Department of Electrical and Computer Engineering, Inha University\\
100, Inha-ro, Michuhol-gu, Incheon, Republic of Korea\\
{\tt\small \{jhnam0514, sclee\}@inha.edu}
}

\begin{document}
\maketitle
\begin{abstract}
Generalizable  medical image segmentation is essential for ensuring consistent performance across diverse unseen clinical settings. However, existing methods often overlook the capability to generalize effectively across arbitrary unseen modalities. In this paper, we introduce a novel \textbf{Test-Time Modality Generalization (TTMG)} framework, which comprises two core components: Modality-Aware Style Projection (MASP) and Modality-Sensitive Instance Whitening (MSIW), designed to enhance generalization in arbitrary unseen modality datasets. The MASP estimates the likelihood of a test instance belonging to each seen modality and maps it onto a distribution using modality-specific style bases, guiding its projection effectively. Furthermore, as high feature covariance hinders generalization to unseen modalities, the MSIW is applied during training to selectively suppress modality-sensitive information while retaining modality-invariant features. By integrating MASP and MSIW, the TTMG framework demonstrates robust generalization capabilities for medical image segmentation in unseen modalities—a challenge that current methods have largely neglected. We evaluated TTMG alongside other domain generalization techniques across eleven datasets spanning four modalities (colonoscopy, ultrasound, dermoscopy, and radiology), consistently achieving superior segmentation performance across various modality combinations. These results highlight TTMG’s effectiveness in addressing diverse medical imaging challenges, outperforming existing approaches without additional training.
\end{abstract}   
\blfootnote{$*$ denotes the corresponding author.}
\section{Introduction}
\label{sec:intro}

\begin{figure}
    \centering
    \includegraphics[width=0.48\textwidth]{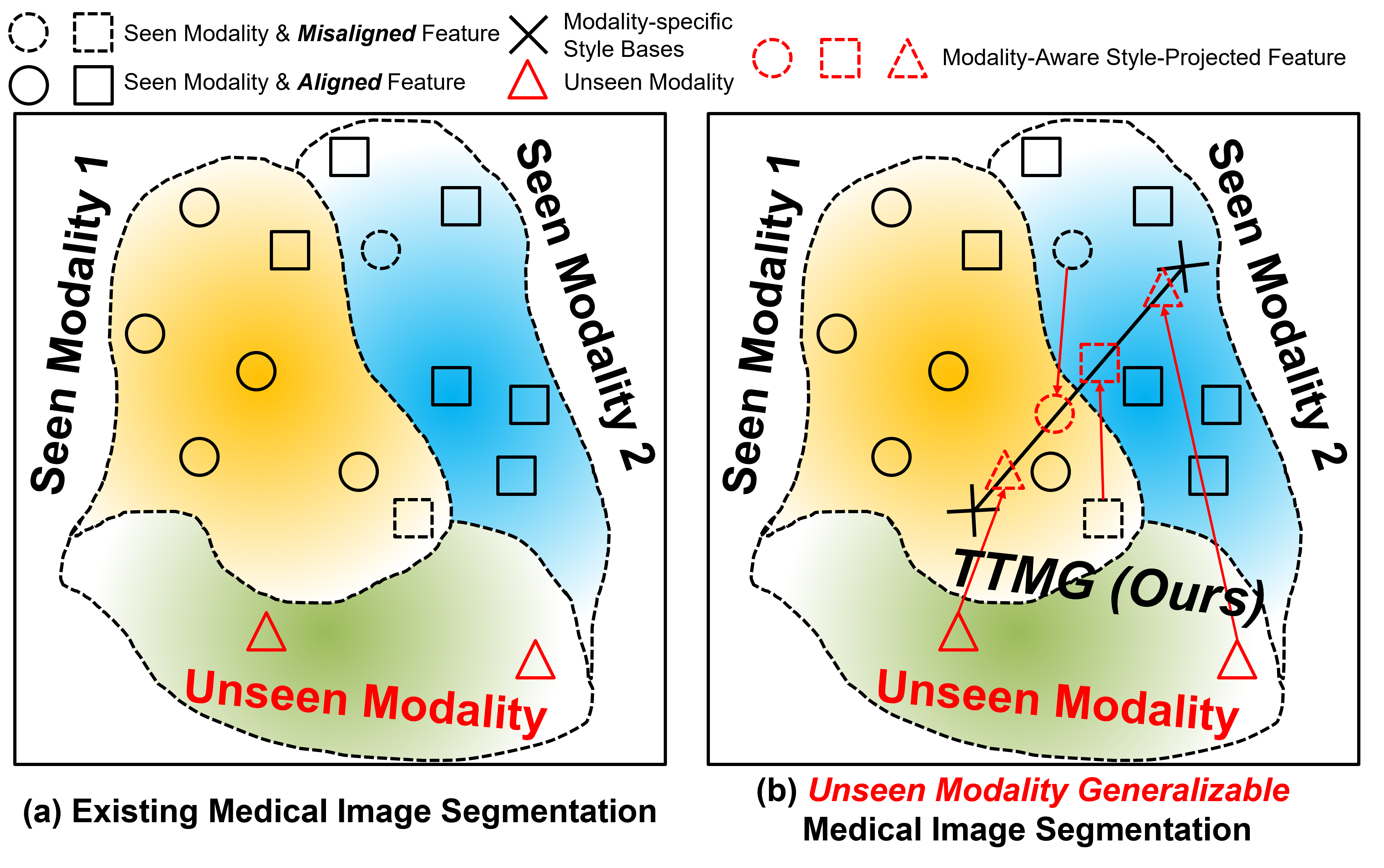}
    \caption{(a) Existing medical image segmentation methods: Performance evaluation on unseen modalities is challenging, often requiring retraining and hyperparameter optimization for each unseen modality dataset, which incurs substantial costs. (b) Schematic diagram of the proposed TTMG framework for unseen modality generalizable medical image segmentation: When an arbitrary unseen modality instance is input, TTMG projects the misaligned, unseen modality features into a well-aligned, seen modality feature distribution using modality-specific style bases, guided by the probability of the instance belonging to each modality \textbf{without additional training}.}
    \label{fig:unseen_modality_generalization}
\end{figure}

Medical image segmentation has been considered a core technology for the early detection of cancer and abnormal tissues that significantly impact human life \cite{coates2015tailoring}. Therefore, medical experts have widely employed traditional automatic segmentation algorithms \cite{otsu1979threshold, kass1988snakes, tizhoosh2005image, haralick1987image}. However, these algorithms are vulnerable to various artifacts and complex morphological structures in medical images, making them hard to generalize to new patients’ cases \cite{riccio2018new}. Consequently, these problems have weakened the reliability of computer-based diagnostic analysis and hindered the development of digital healthcare, which is the leading future industry field \cite{park2019treatment, gaube2021ai}.

Due to advancements in deep learning \cite{krizhevsky2012imagenet, he2016deep, dosovitskiy2021an, wang2022pvt, han2022vision}, recent medical image segmentation models have focused on specific types of medical images, including polyp \cite{fan2020pranet, zhao2021automatic, hu2023ppnet, liu2024cafe}, skin cancer \cite{sarker2018slsdeep, he2023co, din2024lscs}, breast tumor \cite{qi2023mdf, he2024multi}, or lung infection segmentation \cite{fan2020inf, liu2024automated}. These segmentation tasks are often tied to medical imaging modalities, including colonoscopy, dermoscopy, ultrasound, radiology, etc \cite{ma2024segment, guo2019deep}. However, this approach typically necessitates training separate models for each segmentation task, posing significant practical challenges and incurring high costs related to dataset curation, model retraining, and hyperparameter tuning \cite{zhang2022understanding, chen2024versatile}. While some studies \cite{zhao2023m, Rahman_2023_WACV, rahman2024g, nam2024modality} have aimed to improve generalization to unseen clinical settings, no research has explicitly focused on enhancing generalization to entirely unseen modalities—such as \textit{training on colonoscopy datasets and evaluating ultrasound datasets}—using a \textbf{single model}.

Generalizing across various segmentation tasks is particularly challenging due to the inherent heterogeneity of medical image characteristics, where different modalities exhibit diverse feature distributions, noise patterns, and resolution levels \cite{darzi2024review}. To tackle these challenges, we leverage foundational concepts from domain generalization techniques \cite{choi2021robustnet, huang2023style, zhou2023instance, zhou2024test}, which aim to improve generalization ability on arbitrary unseen domains. However, conventional domain generalization methods often struggle to capture complex anatomical structures and retain modality-invariant features in medical images, as they are primarily designed for visual domains with more uniform characteristics.

\begin{figure*}[t]
    \centering
    \includegraphics[width=\textwidth]{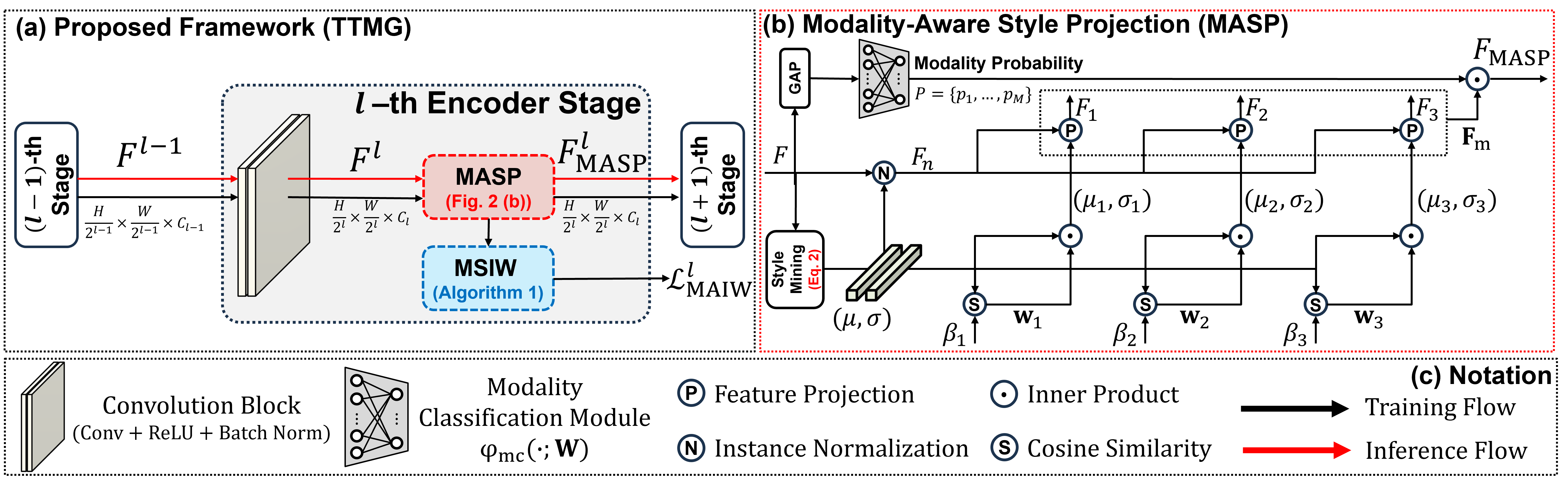}
    \caption{(a) The overall architecture of the proposed framework, called \textbf{Test-Time Modality Generalization (TTMG)}, which mainly comprises \textit{MASP (Figure \ref{fig:TTMG}.(b))} and \textit{MSIW (Algorithm \ref{alg_MSIW})}. (b) MASP stage ($M = 3$ in this figure). (c) Notation description used in this paper.}
    \label{fig:TTMG}
\end{figure*}

To address these challenging issues, we propose a novel technique called \textit{Modality-Aware Style Projection (MASP, Figure \ref{fig:TTMG}.(b))}, leveraging prototype learning by representing the style bases of each accessible modality during training as learnable parameters. Our approach begins by identifying the modality to which the input instance belongs and then projecting the instance into the most similar style space, as illustrated in Figure \ref{fig:unseen_modality_generalization}. While MASP effectively projects instances into the most similar style distribution among seen modalities, it does not mitigate feature covariance, which can lead to overfitting to specific modalities and ultimately hinder the model's ability to generalize effectively to unseen modalities \cite{li2021simple, zhang2024domain}. To address this limitation, we introduced \textit{Modality-Sensitive Instance Whitening (MSIW, Algorithm \ref{alg_MSIW})}, which identifies and removes modality-specific information while preserving modality-invariant features. MSIW  computes the covariance matrix variance between the original and the modality style-projected feature maps, whitening only the covariance matrix elements with high variance. The resulting \textbf{Test-Time Modality Generalization (TTMG)}, which integrates the MASP and MSIW, achieved the highest generalization performance for unseen modality generalizable medical image segmentation \textbf{without additional training}. The contributions of this paper can be summarized as follows:

To address these challenging issues, we propose a novel technique called \textit{Modality-Aware Style Projection (MASP, Figure \ref{fig:TTMG}.(b))}, leveraging prototype learning by representing the style bases of each accessible modality during training as learnable parameters. Our approach begins by identifying the modality to which the input instance belongs and then projecting the instance into the most similar style space, as illustrated in Figure \ref{fig:unseen_modality_generalization}. While MASP effectively projects instances into the most similar style distribution among seen modalities, it does not mitigate feature covariance, which can lead to overfitting to specific modalities and ultimately hinder the model's ability to generalize effectively to unseen modalities \cite{li2021simple, zhang2024domain}. To address this limitation, we introduced \textit{Modality-Sensitive Instance Whitening (MSIW, Algorithm \ref{alg_MSIW})}, which identifies and removes modality-specific information while preserving modality-invariant features. MSIW  computes the covariance matrix variance between the original and the modality style-projected feature maps, whitening only the covariance matrix elements with high variance. The resulting \textbf{Test-Time Modality Generalization (TTMG)}, which integrates the MASP and MSIW, achieved the highest generalization performance for unseen modality generalizable medical image segmentation \textbf{without additional training}. The contributions of this paper can be summarized as follows:

\begin{itemize}
    \item We propose a TTMG framework, mainly comprising MASP and MSIW, for unseen modality generalizable medical image segmentation without additional training. 
    
    \item MASP integrates modality classification modules to project the feature map of a test instance into the most similar style space among available modalities. Meanwhile, MSIW removes only modality-sensitive information while preserving modality-invariant features.
    
    \item We have achieved considerably higher performance than the existing DGSS technique by conducting experiments on a total of eleven datasets, including four modalities that are used as core diagnostic modalities in medical image segmentation.
\end{itemize}

\section{Related Works}

The advancements in deep learning have significantly influenced automatic medical image segmentation \cite{ronneberger2015u, chen2021transunet, zhao2023m, nam2024modality}. However, these models are inherently constrained by the unique challenges posed by the distinct characteristics of each modality, such as variations in artifacts and feature distributions. Consequently, these approaches require expensive retraining for each new modality, limiting their scalability and hindering their practical application. Although some studies \cite{zhou2022generalizable, bastico2023simple, gu2024train} have explored modality generalization within a specific segmentation task, such as brain tumor segmentation, generalization across different tasks remains largely unexplored. To address this unexplored task, we focus on domain generalization (DG) techniques, which aim to enhance generalization performance on arbitrary unseen domains. Recent DG approaches primarily focus on three strategies: 1) Instance Normalization (IN) \cite{ulyanov2017improved, luo2018differentiable, pan2018two}, 2) Instance Whitening (IW) \cite{huang2018decorrelated, pan2019switchable}, and 3) Style Projection \cite{zhou2023instance, huang2023style, ahn2024style}. IN and IW work by standardizing feature maps and decorrelating the covariance of feature maps to reduce the domain-specific variations and improve the model's robustness to unseen domains, respectively. Additionally, SAN-SAW \cite{peng2022semantic} and RobustNet \cite{choi2021robustnet} adopted the strategies of IN and IW, respectively, by selectively applying normalization and whitening: SAN-SAW performs IN only on semantically similar regions, while RobustNet identifies domain-specific information to apply partial whitening rather than whitening the whole feature map. Additionally, BlindNet \cite{ahn2024style}  mitigates style influence through covariance alignment in the encoder with semantic consistency contrastive learning in the decoder. Recently, SPCNet \cite{huang2023style}, IAIW \cite{zhou2023instance}, and TTDG \cite{ahn2024style} have learned style bases from the seen domain and projected unseen domain data onto these learned style bases, facilitating domain generalization at test time. Inspired by these approaches, we developed the TTMG framework for unseen modality generalizable medical image segmentation, integrating domain-specific style recognition algorithm (RobustNet) and prototype learning-based style projection (SPCNet). To the best of our knowledge, prior research has not addressed generalization across diverse modalities beyond specific clinical settings. Our framework bridges this gap by enabling robust performance on unseen modalities without costly retraining, improving scalability, and supporting practical implementation in real-world medical applications.
\section{Method}

In this section, we outline the problem setup (Section \ref{ss31_problem_setup}) for unseen modality generalizable medical image segmentation. Subsequently, we introduce a novel framework, called \textbf{Test-time Modality Generalization (TTMG)}, which integrates two main components: \textit{Modality-Aware Style Projection (MASP, Section \ref{ss32_modality_aware_style_projection})} and \textit{Modality-Sensitive Instance Whitening (MSIW, Section \ref{ss33_modality_sensitive_instance_whitening})}. Our novel framework is designed to improve generalization ability on arbitrary unseen modality segmentation tasks \textbf{without additional training} by projecting arbitrary unseen modality instances to seen modality style distribution. We summarized the technical novelty and social impact of TTMG in the Appendix.

\subsection{Problem Setup}
\label{ss31_problem_setup}
We assume the available of $M$ distinct source modality datasets, denoted as $\mathcal{S} = \{ S_{1}, S_{2}, \dots, S_{M} \}$. The $m$-th source modality dataset $S_{m}$ consists of image-label pairs $(x^{i}_{m}, y^{i}_{m})$, where $x^{i}_{m} \in \mathbb{R}^{H \times W \times C}$ is the image and $y^{i}_{m} \in \mathbb{R}^{H \times W \times K}$ is the corresponding pixel-wise label with $K$ classes in $m$-th source modality. Thus, $S_{m} = \{ (x^{i}_{m}, y^{i}_{m}) \}_{i = 1}^{N_{m}}$, where $N_{m}$ represents the number of samples in the $m$-th modality dataset. And, $(H, W, C)$ denotes the height, width, and number of channels of the input image, respectively. This paper aims to train a binary medical image segmentation model $\varphi_{\text{seg}}$ using a multi-source modality scheme ($M > 1$, $K = 1$) that can generalize to arbitrary unseen modality datasets, which are inaccessible during the training phase.

\subsection{Modality-Aware Style Projection}
\label{ss32_modality_aware_style_projection}

\noindent \textit{Motivation:} We observe that each modality dataset exhibits a distinct feature distribution (Figure \ref{fig:FeatureProjection}.(a)), significantly contributing to performance degradation in deep learning models \cite{gao2022towards, bayram2023domain}. This result is due to learned features often overfitting to the specific characteristics of the training modality, limiting their ability to transfer and generalize effectively to unseen modalities. However, if we can estimate the most likely modality of a test instance and effectively leverage this information, the model's generalization ability to unseen modalities can be significantly improved. To fully leverage this modality information, we propose the \textit{Modality-Aware Style Projection (MASP)}, which is composed of four key steps: \textit{1) Modality Classification Module}, \textit{2) Style Mining and IN}, \textit{3) Style Bases Recalibration}, and \textit{4) Feature Projection}. For clarity, we assume all operations are performed in the $l$-th encoder stage. The overall architecture of the TTMG and MASP stage is illustrated on Figure \ref{fig:TTMG}.

\noindent \textbf{Modality Classification Module.} Let $F \in \mathbb{R}^{H_{l} \times W_{l} \times C_{l}}$ be the feature map at $l$-th encoder stage where $(H_{l}, W_{l}, C_{l})$ denotes the height, width, and number of channels of the input feature map $F$, respectively. The feature map $F$ is then passed through $\varphi_{\text{mc}} (\cdot; \textbf{W})$, a module comprising a Global Average Pooling (GAP) layer followed by a fully connected layer, parameterized by $\textbf{W} \in \mathbb{R}^{C_{l} \times M}$. Finally, we can obtain the probability $p_{m}$ that the input feature map belongs to the $m$-th modality for $m = 1, 2, \dots, M$ by applying the Softmax operation as follows:
\begin{equation}
    p_{m} = \frac{\textbf{exp}(o_{m})}{\sum_{m^{'} = 1}^{M} \textbf{exp} (o_{m^{'}}) }
\end{equation}
\noindent where $o_{m}$ is an $m$-th node output of $\varphi_{\text{mc}} (F; \textbf{W}) \in \mathbb{R}^{M}$. Note that $P = \{ p_{1}, p_{2}, \dots, p_{M} \}$ satisfies the both conditions $\sum_{m = 1}^{M} p_{m} = 1$ and $p_{m} \in [0, 1]$ for all $m = 1, 2, \dots, M$.

\noindent \textbf{Style Mining and IN.} Previous DG methods \cite{choi2021robustnet, lee2022wildnet, huang2023style, zhou2023instance, zhou2024test} have shown that deep layers in segmentation networks capture content information from input images, while shallow layers focus on style features. Moreover, these approaches model the style distribution space using the channel-wise mean and standard deviation of feature maps. Therefore, we obtain the style vector of the feature map as follows:
\begin{equation}
    \begin{cases}
        \mu &= \frac{1}{H_{l}W_{l}} \sum_{h = 1}^{H_{l}} \sum_{w = 1}^{W_{l}} (F)_{h, w, :} \\
        \sigma &= \sqrt{\frac{1}{H_{l}W_{l}} \sum_{h = 1}^{H_{l}} \sum_{w = 1}^{W_{l}} (F - \mu)^{2}_{h, w, :}}
    \end{cases}
\end{equation}
\noindent where $\mu \in \mathbb{R}^{C_{l}}$ and $\sigma \in \mathbb{R}^{C_{l}}$ represent the channel-wise mean and standard deviation of feature map $F$, respectively. Since we assume that the characteristics of different modalities are reflected in their distinct feature distributions, we apply IN \cite{ulyanov2017improved} to remove modality-specific style information by standardizing the features to a normalized distribution (i.e., zero mean and unit variance) as $F_{n} = \frac{F - \mu}{\sigma + \epsilon} \in \mathbb{R}^{H_{l} \times W_{l} \times C_{l}}$ where $\epsilon$ denotes a small value to prevent zero division ($\epsilon = 10^{-6}$).

\noindent \textbf{Style Bases Recalibration.} Now, we recalibrate the style bases $(\mu, \sigma)$ of the input feature map $F$ using a style bases bank that represents the styles of each accessible modality during the training phase. To implement this style recalibration  process, we defined the modality-specific style representation bases bank $\beta_{m} = \{ (\mu^{k}_{m}, \sigma^{k}_{m}) \}_{k = 1}^{K_{m}}$, where $K_{m}$ denotes the number of style bases defining the $m$-th modality style space, inspired by prototype learning-based DG approaches \cite{huang2023style, zhou2023instance, zhou2024test}. Then, we compute the cosine similarity $s^{k}_{m}$ between the input feature map style $(\mu, \sigma)$ and $k$-th style $(\mu^{k}_{m}, \sigma^{k}_{m})$ in $m$-th modality-specific style bases bank $\beta_{m}$ for $k = 1, 2, \dots, K_{m}$ and $m = 1, 2, \dots, M$ as follows:
\begin{equation}
    s^{k}_{m} = \frac{\mu \cdot \mu^{k}_{m}}{|| \mu || \cdot || \mu^{k}_{m} ||} + \frac{\sigma \cdot \sigma^{k}_{m}}{|| \sigma || \cdot || \sigma^{k}_{m} ||}
\end{equation}
\noindent Subsequently, we apply the Softmax operation to cosine similarity values $s^{k}_{m}$ to obtain weight value $\textbf{w}_{m} = \{ w^{k}_{m} \}_{k = 1}^{K_{m}}$ satisfying $\sum_{k = 1}^{K_{m}} w^{k}_{m} = 1$ and $w^{k}_{m} \in [0, 1]$ for $k = 1, 2, \dots, K_{m}$. This process ensures that the cosine similarity scores $s^{k}_{m}$ are normalized into a probabilistic distribution, allowing for an adaptive weighting of different styles from the style bank based on their relevance to the input feature map. Finally, using this weight value, we can obtain recalibrated modality-specific style bases $(\mu_{m}, \sigma_{m})$ for each $m = 1, 2, \dots, M$ as follows:
\begin{equation}
        \mu_{m} = \sum_{k = 1}^{K_{m}} w^{k}_{m} \cdot \mu^{k}_{m}, \sigma_{m} = \sum_{k = 1}^{K_{m}} w^{k}_{m} \cdot \sigma^{k}_{m}
\end{equation}

\noindent \textbf{Feature Projection.} Finally, we project the normalized features $F_{n}$ onto the style distribution of each modality by using the corresponding modality-specific style bases $(\mu_{m}, \sigma_{m})$ as $\mathbf{F}_{m} = \{ F_{m} \}_{m = 1}^{M}$ where $F_{m} = F_{n} \cdot \sigma_{m} + \mu_{m} \in \mathbb{R}^{H_{l} \times W_{l} \times C_{l}}$ for each $m = 1, 2, \dots, M$. Therefore, a $F_{m}$ represents a feature map with modality-specific style. Subsequently, the modality classification output $P$ is used to map the feature map to the most similar modality style space as follows:
\begin{equation}
    F_{\text{MASP}} = \sum_{m = 1}^{M} p_{m} \cdot F_{m} \in \mathbb{R}^{H_{l} \times W_{l} \times C_{l}}
\end{equation}
\noindent By utilizing the modality probability $P = \{ p_{m} \}_{m = 1}^{M}$, we can more precisely capture the most representative feature distribution of each modality. Additionally, this probabilistic information allows the model to adjust its feature extraction process based on the likelihood of each modality, enabling a more accurate projection of the distinct characteristics of the seen modalities. Subsequently, the output feature map $F_{\text{MASP}}$ forwards to the next encoder block.

\begin{algorithm}[t]
\caption{Modality-Sensitive Instance Whitening}
\label{alg_MSIW}
\textbf{Input}: Original feature map $F$, Modality-aware Style-projected feature map $F_{\text{MASP}}$, and Modality-specific style-projected feature maps $F_{m}$ for $m = 1, 2, \dots, M$\\
\textbf{Output}: MSIW loss $\mathcal{L}_{\text{MSIW}}$
\begin{algorithmic}[1] 
\STATE Initialize $\mathcal{L}_{\text{MSIW}} = 0$
\STATE Compute covariance matrix $\Lambda_{0}$ from $F$
\STATE Compute covariance matrix $\Lambda_{M + 1}$ from $F_{\text{MASP}}$
\FOR{$F_{m}$ for $m = 1, 2, \dots, M$}
    \STATE Compute covariance matrix $\Lambda_{m}$ from $F_{m}$
\ENDFOR
\STATE Compute the variance matrix $\mathbf{V}$ of $\Lambda_{m}$ for $m = 0, 1, \dots, M, M + 1$
\STATE Apply $k$-means clustering on the strict upper triangular element $\mathbf{V}_{\text{U}}$ of the variance matrix $\mathbf{V}$
\STATE Separate each cluster into two groups $\mathbf{M}_{\text{low}}$ and $\mathbf{M}_{\text{high}}$
\FOR{$\Lambda_{m}$ for $m = 0, 1, \dots, M, M + 1$}
    \STATE $\mathcal{L}_{\text{MSIW}} = \mathcal{L}_{\text{MSIW}} + || \Lambda_{m} \odot \mathbf{M}_{\text{high}} ||_{1}$
\ENDFOR
\STATE $\mathcal{L}_{\text{MSIW}} = \frac{1}{M + 2} \mathcal{L}_{\text{MSIW}} $
\RETURN $\mathcal{L}_{\text{MSIW}}$
\end{algorithmic}
\end{algorithm}

\subsection{Modality-Sensitive Instance Whitening}
\label{ss33_modality_sensitive_instance_whitening}

\noindent \textit{Motivation:} Although $F_{\text{MASP}}$ is aligned with the most similar modality style distribution via the MASP stage, high covariance between feature channels still limits generalization \cite{li2021simple, zhang2024domain}. While IW effectively addresses this issue, completely whitening covariance degrades performance on seen modalities \cite{choi2021robustnet, peng2022semantic}. To address this issue, we propose \textit{Modality-Sensitive Instance Whitening (MSIW)}, which separates modality-sensitive and modality-invariant information and then selectively whitening only the covariance of modality-sensitive information.

To separate the modality-sensitive information, we use a total $M + 2$ feature maps including original feature $F$, modality-specific style-projected feature $F_{m}$ for $m = 1, 2, \dots, M$, and modality-aware style-projected feature $F_{\text{MASP}}$. Inspired by previous works \cite{choi2021robustnet, jung2023local}, we define modality-sensitive information as the channels that exhibit high variance of the covariance matrix $\Lambda_{m}$ obtained from original, modality-specific style-projected, and modality-aware style-projected feature map. Subsequently, the variance matrix of covariance matrices is calculated as follows:
\begin{equation}
    \mathbf{V} = \frac{1}{M + 2} \sum_{m = 0}^{M + 1} \Lambda_{m}
\end{equation}
\noindent where $\Lambda_{0}$, and $\Lambda_{M + 1}$ denote the covariance matrices of the original and modality-aware style projected feature maps, respectively. And, $\Lambda_{m}$ denotes the covariance matrix of modality-specific style projected feature map for $m = 1, 2, \dots, M$. Since the variance matrix $\mathbf{V}$ is a symmetric matrix, we only consider the strict upper triangular element $\mathbf{V}_{\text{U}} = \{\mathbf{V}_{i, j} | i < j \text{ for } i = 1, \dots, C_{l} \text{ and } j = 1, \dots, C_{l}\}$ to eliminate modality-sensitive information. Then, $\mathbf{V}_{\text{U}}$ is divided into $k$ clusters using $k$-means clustering \cite{lloyd1982least}, and each cluster is separated into two groups, called $\mathbf{M}_{\text{low}} = \{ v_{1}, \dots, v_{s} \} $ and $\mathbf{M}_{\text{high}}  = \{ v_{s + 1}, \dots, v_{k} \}$ with low and high variance, respectively. Therefore, we assume that $\mathbf{M}_{\text{low}}$ and $\mathbf{M}_{\text{high}}$ have modality-invariant and modality-sensitive information, respectively. Finally, we define MSIW loss as $\mathcal{L}_{\text{MSIW}} = \mathbb{E} \left[ || \Lambda_{m} \odot \mathbf{M}_{\text{high}} ||_{1} \right]$. Algorithm \ref{alg_MSIW} describes the detailed training algorithm for MSIW.

\subsection{Network Training}
During training, we define $\mathcal{L}^{l}_{\text{con}}$ to preserve the contents information of style-projected features by maximizing the cosine similarities between normalized and modality-specific style-projected features as follows:
\begin{equation}
    \mathcal{L}_{\text{con}} = -\frac{1}{M} \sum_{m = 1}^{M} \log\left( \frac{\textbf{exp} (z_{nm})}{\sum_{m^{'} = 1}^{M} \textbf{exp} (z_{nm^{'}})} \right)
\end{equation}
\noindent where $z_{nm} = \frac{F_{n}}{||F_{n}||_{2}} \cdot \frac{F_{m}}{||F_{m}||_{2}}$. Additionally, we compute $\mathcal{L}^{l}_{\text{cls}}$ to classify seen modality during training as follows:
\begin{equation}
    \mathcal{L}_{\text{cls}} = \sum_{m = 1}^{M} y^{\text{cls}}_{m} \log(p_{m})
\end{equation}
\noindent where $y^{\text{cls}}$ is the label for the available modality to train modality classification network $\varphi_{\text{mc}} (\cdot; \textbf{W})$. Finally, we combine four different loss terms for the end-to-end training as follows:
\begin{equation}
    \mathcal{L}_{\text{total}} = \mathcal{L}_{\text{seg}} + \frac{1}{L} \sum_{l \in L} \left( \mathcal{L}^{l}_{\text{cls}} + \mathcal{L}^{l}_{\text{con}} + \mathcal{L}^{l}_{\text{MSIW}} \right)
\end{equation}
\noindent where $L$ denotes the set of layers to which MASP and MSIW are applied. And, the loss function for segmentation was defined as $\mathcal{L}_{\text{seg}} = \mathcal{L}^{w}_{IoU} + \mathcal{L}^{w}_{bce}$, where $\mathcal{L}^{w}_{IoU}$ and $\mathcal{L}^{w}_{bce}$ are the weighted IoU and binary cross entropy (BCE) loss functions, respectively. This loss function is identically defined in previous studies \cite{fan2020pranet, zhao2023m, nam2024modality}. For each training phase, the parameter of segmentation network $\varphi_{\text{seg}}$, classification network $\varphi_{\text{mc}}$, and style bases $\beta_{m}$ for $m = 1, 2, \dots, M$ are updated in end-to-end manner. After the training phase, all parameters are not updated during the inference phase.

\begin{table*} [t]
    \centering
    \small
    \setlength\tabcolsep{2.5pt} 
    \begin{tabular}{c|cc|cc|cc|cc|cc|cc|cc|cc} 
    \hline
    \multicolumn{1}{c|}{\multirow{3}{*}{Method}} & \multicolumn{4}{c|}{Training Modalities (C, U, D)} & \multicolumn{4}{c|}{Training Modalities (C, U, R)} & \multicolumn{4}{c|}{Training Modalities (C, D, R)} & \multicolumn{4}{c}{Training Modalities (U, D, R)} \\\cline{2-17}
     & \multicolumn{2}{c|}{Seen (C, U, D)} & \multicolumn{2}{c|}{Unseen (R)} & \multicolumn{2}{c|}{Seen (C, U, R)} & \multicolumn{2}{c|}{Unseen (D)} & \multicolumn{2}{c|}{Seen (C, D, R)} & \multicolumn{2}{c|}{Unseen (U)} & \multicolumn{2}{c|}{Seen (U, D, R)} & \multicolumn{2}{c}{Unseen (C)} \\\cline{2-17}
     & DSC & mIoU & DSC & mIoU & DSC & mIoU & DSC & mIoU & DSC & mIoU & DSC & mIoU & DSC & mIoU & DSC & mIoU \\
     \hline
     baseline \cite{chen2018encoder} & 80.98 & 73.65 & 14.30 & 9.12 & \textcolor{blue}{\textbf{\textit{76.20}}} & \textcolor{blue}{\textbf{\textit{68.67}}} & 42.43 & 36.24 & 77.95 & 70.62 & 13.55 & 8.16 & 76.32 & 67.99 & 19.36 & 12.78 \\
     \hline
     \multicolumn{1}{c|}{\multirow{2}{*}{IN \cite{ulyanov2017improved}}} & \textcolor{red}{\textbf{\underline{82.24}}} & \textcolor{red}{\textbf{\underline{75.00}}} & 13.92 & 8.79 & 74.23 & 66.48 & 48.15 & \textcolor{blue}{\textbf{\textit{39.85}}} & \textcolor{red}{\textbf{\underline{78.17}}} & \textcolor{blue}{\textbf{\textit{70.84}}} & 26.82 & 17.94 & 74.49 & 65.32 & 14.22 & 9.06 \\
      & \textcolor{ForestGreen}{\scriptsize{\textbf{(+1.26)}}}  & \textcolor{ForestGreen}{\scriptsize{\textbf{(+1.35)}}} 
      & \textcolor{red}{\scriptsize{\textbf{(-0.38)}}}          & \textcolor{red}{\scriptsize{\textbf{(-0.33)}}} 
      & \textcolor{red}{\scriptsize{\textbf{(-1.97)}}}          & \textcolor{red}{\scriptsize{\textbf{(-2.19)}}} 
      & \textcolor{ForestGreen}{\scriptsize{\textbf{(+5.72)}}}  & \textcolor{ForestGreen}{\scriptsize{\textbf{(+3.61)}}} 
      & \textcolor{ForestGreen}{\scriptsize{\textbf{(+0.22)}}}  & \textcolor{ForestGreen}{\scriptsize{\textbf{(+0.22)}}} 
      & \textcolor{ForestGreen}{\scriptsize{\textbf{(+13.27)}}} & \textcolor{ForestGreen}{\scriptsize{\textbf{(+9.78)}}} 
      & \textcolor{red}{\scriptsize{\textbf{(-1.83)}}}          & \textcolor{red}{\scriptsize{\textbf{(-2.67)}}}
      & \textcolor{red}{\scriptsize{\textbf{(-5.14)}}}          & \textcolor{red}{\scriptsize{\textbf{(-3.72)}}} \\ 
    \hline
     \multicolumn{1}{c|}{\multirow{2}{*}{IW \cite{huang2018decorrelated}}} & 72.93 & 64.74 & 9.85 & 6.21 & 60.36 & 52.12 & \textcolor{blue}{\textbf{\textit{50.11}}} & 38.79 & 65.98 & 58.45 & 34.65 & 25.30 & 68.64 & 61.26 & 15.01 & 9.35 \\
      & \textcolor{red}{\scriptsize{\textbf{(-8.05)}}}          & \textcolor{red}{\scriptsize{\textbf{(-8.91)}}} 
      & \textcolor{red}{\scriptsize{\textbf{(-4.45)}}}          & \textcolor{red}{\scriptsize{\textbf{(-2.91)}}} 
      & \textcolor{red}{\scriptsize{\textbf{(-15.84)}}}         & \textcolor{red}{\scriptsize{\textbf{(-16.55)}}} 
      & \textcolor{ForestGreen}{\scriptsize{\textbf{(+7.68)}}}  & \textcolor{ForestGreen}{\scriptsize{\textbf{(+2.55)}}} 
      & \textcolor{red}{\scriptsize{\textbf{(-11.97)}}}         & \textcolor{red}{\scriptsize{\textbf{(-12.17)}}} 
      & \textcolor{ForestGreen}{\scriptsize{\textbf{(+21.10)}}} & \textcolor{ForestGreen}{\scriptsize{\textbf{(+17.14)}}} 
      & \textcolor{red}{\scriptsize{\textbf{(-7.68)}}}          & \textcolor{red}{\scriptsize{\textbf{(-6.73)}}} 
      & \textcolor{red}{\scriptsize{\textbf{(-4.35)}}}          & \textcolor{red}{\scriptsize{\textbf{(-3.43)}}} \\
      \hline
     \multicolumn{1}{c|}{\multirow{2}{*}{IBN \cite{pan2018two}}} & 80.98 & 73.89 & \textcolor{blue}{\textbf{\textit{16.43}}} & \textcolor{blue}{\textbf{\textit{10.93}}} & \textcolor{blue}{\textbf{\textit{76.20}}} & 68.64 & 43.50 & 36.59 & 78.00 & 70.62 & 30.74 & 20.97 & \textcolor{blue}{\textbf{\textit{77.87}}} & \textcolor{blue}{\textbf{\textit{69.54}}} & 20.51 & 13.73 \\
      & \scriptsize{\textbf{(+0.00)}}                           & \textcolor{ForestGreen}{\scriptsize{\textbf{(+0.24)}}} 
      & \textcolor{ForestGreen}{\scriptsize{\textbf{(+2.13)}}}  & \textcolor{ForestGreen}{\scriptsize{\textbf{(+1.81)}}} 
      & \scriptsize{\textbf{(+0.00)}}                           & \textcolor{red}{\scriptsize{\textbf{(-0.03)}}} 
      & \textcolor{ForestGreen}{\scriptsize{\textbf{(+1.07)}}}  & \textcolor{ForestGreen}{\scriptsize{\textbf{(+0.35)}}}
      & \textcolor{ForestGreen}{\scriptsize{\textbf{(+0.05)}}}  & \scriptsize{\textbf{(+0.00)}} 
      & \textcolor{ForestGreen}{\scriptsize{\textbf{(+17.19)}}} & \textcolor{ForestGreen}{\scriptsize{\textbf{(+12.81)}}} 
      & \textcolor{ForestGreen}{\scriptsize{\textbf{(+1.55)}}}  & \textcolor{ForestGreen}{\scriptsize{\textbf{(+1.55)}}} 
      & \textcolor{ForestGreen}{\scriptsize{\textbf{(+1.15)}}}  & \textcolor{ForestGreen}{\scriptsize{\textbf{(+0.95)}}} \\
      \hline
     \multicolumn{1}{c|}{\multirow{2}{*}{RobustNet \cite{choi2021robustnet}}} & 79.84 & 73.35 & 7.17 & 4.62 & 70.20 & 62.40 & 46.89 & 38.44 & 70.78 & 63.59 & \textcolor{red}{\textbf{\underline{43.63}}} & \textcolor{red}{\textbf{\underline{33.81}}} & 72.87 & 64.72 & 17.59 & 11.74 \\
      & \textcolor{red}{\scriptsize{\textbf{(-1.14)}}}          & \textcolor{red}{\scriptsize{\textbf{(-0.30)}}}
      & \textcolor{red}{\scriptsize{\textbf{(-7.13)}}}          & \textcolor{red}{\scriptsize{\textbf{(-4.50)}}} 
      & \textcolor{red}{\scriptsize{\textbf{(-6.00)}}}          & \textcolor{red}{\scriptsize{\textbf{(-6.27)}}} 
      & \textcolor{ForestGreen}{\scriptsize{\textbf{(+4.46)}}}  & \textcolor{ForestGreen}{\scriptsize{\textbf{(+2.20)}}}
      & \textcolor{red}{\scriptsize{\textbf{(-7.17)}}}          & \textcolor{red}{\scriptsize{\textbf{(-7.03)}}} 
      & \textcolor{ForestGreen}{\scriptsize{\textbf{(+30.08)}}} & \textcolor{ForestGreen}{\scriptsize{\textbf{(+25.65)}}} 
      & \textcolor{red}{\scriptsize{\textbf{(-3.45)}}}          & \textcolor{red}{\scriptsize{\textbf{(-3.27)}}} 
      & \textcolor{red}{\scriptsize{\textbf{(-1.77)}}}          & \textcolor{red}{\scriptsize{\textbf{(-1.04)}}} \\
      \hline
     \multicolumn{1}{c|}{\multirow{2}{*}{SAN-SAW \cite{peng2022semantic}}} & 80.29 & 73.12 & 14.24 & 8.58 & 75.14 & 67.85 & 40.04 & 30.35 & 77.11 & 70.04 & 32.67 & 23.48 & 77.63 & 69.52 & 20.91 & 13.89 \\
      & \textcolor{red}{\scriptsize{\textbf{(-0.69)}}}          & \textcolor{red}{\scriptsize{\textbf{(-0.53)}}} 
      & \textcolor{red}{\scriptsize{\textbf{(-0.06)}}}          & \textcolor{red}{\scriptsize{\textbf{(-0.54)}}} 
      & \textcolor{red}{\scriptsize{\textbf{(-1.06)}}}          & \textcolor{red}{\scriptsize{\textbf{(-0.82)}}} 
      & \textcolor{red}{\scriptsize{\textbf{(-2.39)}}}          & \textcolor{red}{\scriptsize{\textbf{(-5.89)}}} 
      & \textcolor{red}{\scriptsize{\textbf{(-0.84)}}}          & \textcolor{red}{\scriptsize{\textbf{(-0.58)}}} 
      & \textcolor{ForestGreen}{\scriptsize{\textbf{(+19.12)}}} & \textcolor{ForestGreen}{\scriptsize{\textbf{(+15.32)}}} 
      & \textcolor{ForestGreen}{\scriptsize{\textbf{(+1.31)}}}  & \textcolor{ForestGreen}{\scriptsize{\textbf{(+1.53)}}} 
      & \textcolor{ForestGreen}{\scriptsize{\textbf{(+1.55)}}}  & \textcolor{ForestGreen}{\scriptsize{\textbf{(+1.11)}}} \\
      \hline
     \multicolumn{1}{c|}{\multirow{2}{*}{SPCNet \cite{huang2023style}}} & \textcolor{blue}{\textbf{\textit{81.23}}} & \textcolor{blue}{\textbf{\textit{74.13}}} & 12.86 & 8.36 & 74.95 & 67.47 & 47.70 & 39.60 & \textcolor{blue}{\textbf{\textit{78.11}}} & \textcolor{red}{\textbf{\underline{70.86}}} & 20.12 & 12.83 & 77.63 & 69.49 & 19.87 & 13.47 \\
      & \textcolor{ForestGreen}{\scriptsize{\textbf{(+0.25)}}} & \textcolor{ForestGreen}{\scriptsize{\textbf{(+0.48)}}} 
      & \textcolor{red}{\scriptsize{\textbf{(-1.44)}}}         & \textcolor{red}{\scriptsize{\textbf{(-0.76)}}} 
      & \textcolor{red}{\scriptsize{\textbf{(-1.25)}}}         & \textcolor{red}{\scriptsize{\textbf{(-1.20)}}} 
      & \textcolor{ForestGreen}{\scriptsize{\textbf{(+5.27)}}} & \textcolor{ForestGreen}{\scriptsize{\textbf{(+3.36)}}} 
      & \textcolor{ForestGreen}{\scriptsize{\textbf{(+0.16)}}} & \textcolor{ForestGreen}{\scriptsize{\textbf{(+0.24)}}} 
      & \textcolor{ForestGreen}{\scriptsize{\textbf{(+6.57)}}} & \textcolor{ForestGreen}{\scriptsize{\textbf{(+4.67)}}} 
      & \textcolor{ForestGreen}{\scriptsize{\textbf{(+1.31)}}} & \textcolor{ForestGreen}{\scriptsize{\textbf{(+1.50)}}} 
      & \textcolor{ForestGreen}{\scriptsize{\textbf{(+0.51)}}} & \textcolor{ForestGreen}{\scriptsize{\textbf{(+0.69)}}} \\
      \hline
     \multicolumn{1}{c|}{\multirow{2}{*}{BlindNet \cite{ahn2024style}}} & 79.87 & 72.75 & 4.54 & 3.04 & 72.69 & 65.16 & 27.63 & 22.65 & 73.07 & 65.56 & 33.06 & 25.55 & 76.49 & 68.36 & \textcolor{blue}{\textbf{\textit{22.73}}} & \textcolor{blue}{\textbf{\textit{17.58}}} \\
      & \textcolor{red}{\scriptsize{\textbf{(-1.11)}}}          & \textcolor{red}{\scriptsize{\textbf{(-0.90)}}} 
      & \textcolor{red}{\scriptsize{\textbf{(-9.76)}}}          & \textcolor{red}{\scriptsize{\textbf{(-6.08)}}} 
      & \textcolor{red}{\scriptsize{\textbf{(-3.51)}}}          & \textcolor{red}{\scriptsize{\textbf{(-3.51)}}} 
      & \textcolor{red}{\scriptsize{\textbf{(-14.80)}}}         & \textcolor{red}{\scriptsize{\textbf{(-13.59)}}} 
      & \textcolor{red}{\scriptsize{\textbf{(-4.88)}}}          & \textcolor{red}{\scriptsize{\textbf{(-5.06)}}} 
      & \textcolor{ForestGreen}{\scriptsize{\textbf{(+19.51)}}} & \textcolor{ForestGreen}{\scriptsize{\textbf{(+17.39)}}} 
      & \textcolor{ForestGreen}{\scriptsize{\textbf{(+0.17)}}}  & \textcolor{ForestGreen}{\scriptsize{\textbf{(+0.37)}}} 
      & \textcolor{ForestGreen}{\scriptsize{\textbf{(+3.37)}}}  & \textcolor{ForestGreen}{\scriptsize{\textbf{(+4.80)}}} \\
      \hline
     \multicolumn{1}{c|}{\multirow{3}{*}{\textbf{TTMG (Ours)}}} & 79.93 & 72.38 & \textcolor{red}{\textbf{\underline{17.55}}} & \textcolor{red}{\textbf{\underline{11.63}}} & \textcolor{red}{\textbf{\underline{78.79}}} & \textcolor{red}{\textbf{\underline{71.10}}} & \textcolor{red}{\textbf{\underline{52.55}}} & \textcolor{red}{\textbf{\underline{42.40}}} & \textcolor{red}{\textbf{\underline{78.17}}} & \textcolor{blue}{\textbf{\textit{70.84}}} & \textcolor{blue}{\textbf{\textit{38.47}}} & \textcolor{blue}{\textbf{\textit{27.93}}} & \textcolor{red}{\textbf{\underline{78.59}}} & \textcolor{red}{\textbf{\underline{70.61}}} & \textcolor{red}{\textbf{\underline{25.05}}} & \textcolor{red}{\textbf{\underline{17.61}}} \\
      & \textcolor{red}{\scriptsize{\textbf{(-1.05)}}}          & \textcolor{red}{\scriptsize{\textbf{(-1.27)}}} 
      & \textcolor{ForestGreen}{\scriptsize{\textbf{(+3.25)}}}  & \textcolor{ForestGreen}{\scriptsize{\textbf{(+2.51)}}}
      & \textcolor{ForestGreen}{\scriptsize{\textbf{(+2.59)}}}  & \textcolor{ForestGreen}{\scriptsize{\textbf{(+2.43)}}} 
      & \textcolor{ForestGreen}{\scriptsize{\textbf{(+10.12)}}} & \textcolor{ForestGreen}{\scriptsize{\textbf{(+6.16)}}} 
      & \textcolor{ForestGreen}{\scriptsize{\textbf{(+0.22)}}}  & \textcolor{ForestGreen}{\scriptsize{\textbf{(+0.22)}}} 
      & \textcolor{ForestGreen}{\scriptsize{\textbf{(+24.92)}}} & \textcolor{ForestGreen}{\scriptsize{\textbf{(+19.77)}}} 
      & \textcolor{ForestGreen}{\scriptsize{\textbf{(+2.27)}}}  & \textcolor{ForestGreen}{\scriptsize{\textbf{(+2.62)}}} 
      & \textcolor{ForestGreen}{\scriptsize{\textbf{(+5.69)}}}  & \textcolor{ForestGreen}{\scriptsize{\textbf{(+4.83)}}} \\ \cline{2-17}
      & \textbf{-2.31} & \textbf{-2.62} & \textbf{+1.12} & \textbf{+0.70} 
      & \textbf{+2.59} & \textbf{+2.43} & \textbf{+2.44} & \textbf{+2.55} 
      & \textbf{+0.00} & \textbf{-0.02} & \textbf{-5.16} & \textbf{-5.88} 
      & \textbf{+0.72} & \textbf{+1.07} & \textbf{+2.32} & \textbf{+0.03} \\
     \hline
    \end{tabular}
    \caption{Segmentation results on three modality training schemes (C, U, D), (C, U, R), (C, D, R), and (U, D, R).}
    \label{tab:comparison_three_modality}
\end{table*}
\section{Experiment Results}

\begin{figure*}[t]
    \centering
    \includegraphics[width=\textwidth]{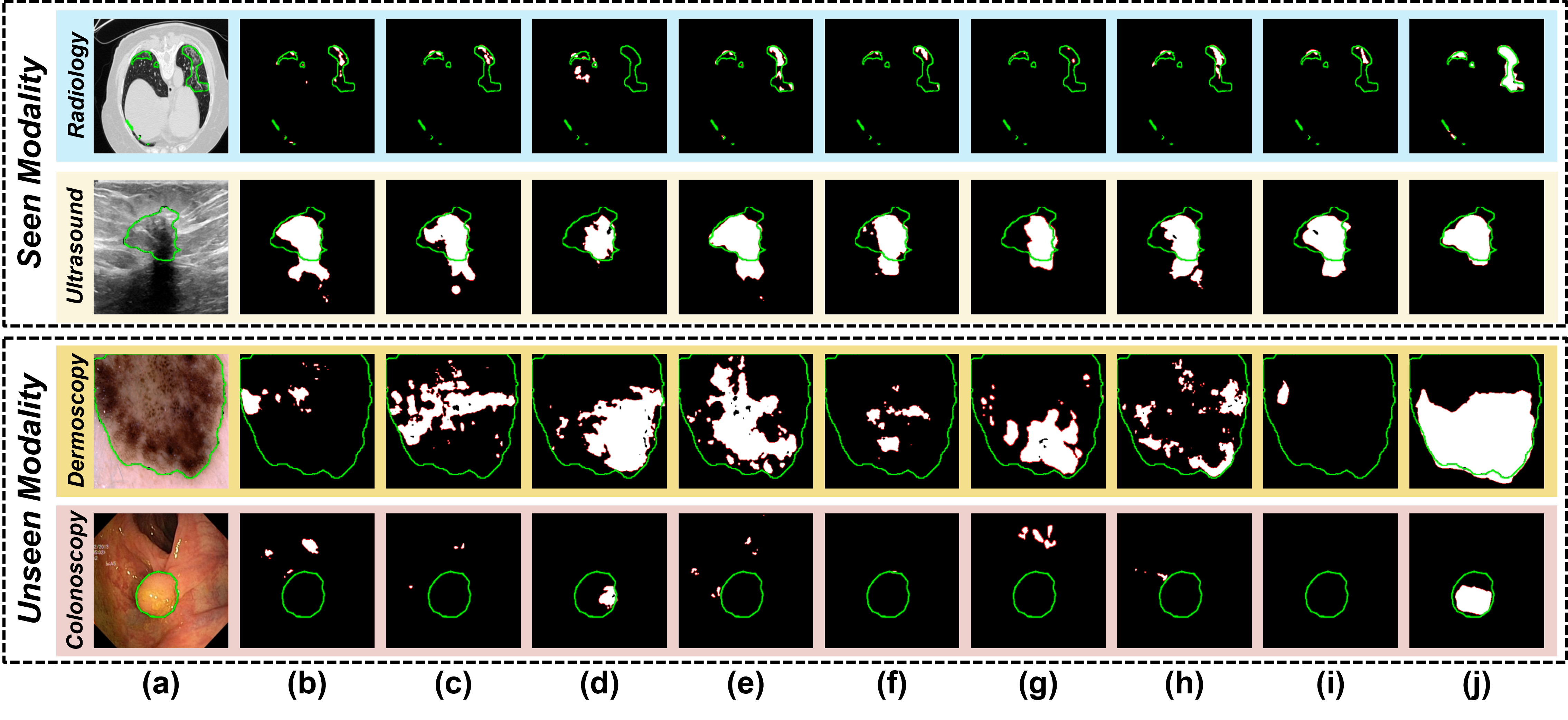}
    \caption{Qualitative comparison of other methods and TTMG on Radiology and Ultrasound modalities.  (a) Input images with ground truth. (b) baseline (DeepLabV3+ with ResNet50, \cite{chen2018encoder}). (c) IN \cite{ulyanov2017improved}. (d) IW \cite{huang2018decorrelated}. (e) IBN \cite{pan2018two}. (f) RobustNet \cite{choi2021robustnet}. (g) SAN-SAW \cite{peng2022semantic}. (h) SPCNet \cite{huang2023style}. (i) BlindNet \cite{ahn2024style}. (j) \textbf{TTMG (Ours)}. In this figure, \textcolor{green}{\textbf{Green}} and \textcolor{red}{\textbf{Red}} lines denote the boundaries of the ground truth and prediction, respectively.}
    \label{fig:QualitativeResults}
\end{figure*}

\subsection{Experiment Settings}
\label{s41_experiment_settings}

In this paper, we use eleven datasets spanning four modalities (colonoscopy (C), ultrasound (U), dermoscopy (D), and radiology (R)), which have been used for training and evaluating various deep learning-based medical image segmentation models \cite{zhou2018unet++, chen2021transunet,  cao2022swin, wang2022uctransnet, zhao2023m, jiang2023vig, wang2024cfatransunet, nam2024modality}. For convenience, we denote the seen and unseen modalities as the test datasets, which are the same and different modalities as source training modality datasets, respectively. Due to the page limit, we present the detailed dataset description in the Appendix \ref{appendix_dataset_descriptions}. Since this paper consider the case of multi-source modality ($M > 1$), we conducted two experiments for $M = 3$ and $M = 2$, which are listed in Tables \ref{tab:comparison_three_modality} and \ref{tab:comparison_two_modality}, respectively. To evaluate the performance of each method, we selected two metrics, the Dice Score Coefficient (DSC) and mean Intersection over Union (mIoU), which are widely used in medical image segmentation. Additionally, the quantitative results with more various metrics \cite{margolin2014evaluate, fan2017structure, fan2018enhanced} and metrics descriptions are also available in the Appendix. 

We compared the proposed \textbf{TTMG} with seven representative domain generalization method, including IN \cite{ulyanov2017improved}, IW \cite{huang2018decorrelated}, IBN \cite{pan2018two}, RobustNet \cite{choi2021robustnet}, SAN-SAW \cite{peng2022semantic}, SPCNet \cite{huang2023style}, and BlindNet \cite{ahn2024style}. For all tables, \textcolor{red}{\textbf{\underline{Red}}} and \textcolor{blue}{\textbf{\textit{Blue}}} are the first and second-best performance results, respectively. And, $( \cdot )$ denotes the performance gap between the baseline and each method. Additionally, the last row indicates the performance gap between \textbf{TTMG} and other best methods.

\subsection{Implementation Details}
\label{s42_implementation_details}

\noindent\textbf{Training Settings.} We implemented TTMG on a single NVIDIA RTX 3090 Ti in Pytorch 1.8 \cite{NEURIPS2019_9015}. Following the other DGSS methods, we choose DeepLabV3+ \cite{chen2018encoder} with ResNet50 \cite{he2016deep} as baseline. Additionally, we also provide the experiment results with other backbones (MobileNetV2 \cite{sandler2018mobilenetv2} and ShuffleNetV2 \cite{ma2018shufflenet}) and TransUNet \cite{chen2021transunet}, which is the most representative Transformer-based model, in the Appendix (Table \ref{tab:backbone_type}) due to the page limit. Additionally, we employ an Adam optimizer \cite{KingmaB14} with an initial learning rate of $10^{-4}$ and reduced the parameters of each method to $10^{-6}$ using a cosine annealing learning rate scheduler \cite{loshchilov2016sgdr}. According to the most representative works for medical image segmentation, the required epochs for each modalities are 50, 100, 100, and 200 epochs. Since we only consider multi-modality ($M > 1$) training scheme in this paper, we choose the largest epoch when different epochs are used. For example, we trained the model using colonoscopy and radiology modalities, and then we optimized the model for 200 epochs since the radiology dataset requires more epochs. During training, we used horizontal/vertical flipping, with a probability of 50\%, and rotation between $-5^{\circ}$ and $5^{\circ}$, which is widely used in medical image segmentation \cite{fan2020pranet, zhao2021automatic, zhao2023m, nam2024modality, nam2025transgunet}. At this stage, because images in each dataset have different resolutions, all images were resized to $352 \times 352$.

\noindent\textbf{Hyperparameters of TTMG.} Key hyperparameters for TTMG on all datasets were set to $K_{m} = 8$ in MASP, and $k = 3$ and $s = 1$ in MSIW following RobustNet \cite{choi2021robustnet}. Additionally, $\beta_{m}$ is randomly initialized to start training. We apply MASP and MSIW in two shallow layers, that is, $L = \{ 1, 2 \}$ (Stage 1 and Stage 2). In the Appendix, we provide the experiment results on various hyperparameter settings. \vspace{-0.15cm}

\begin{table*} [t]
    \centering
    \small
    \setlength\tabcolsep{5.5pt} 
    \begin{tabular}{c|cc|cc|cc|cc|cc|cc} 
    \hline
    \multicolumn{1}{c|}{\multirow{3}{*}{Method}} & \multicolumn{4}{c|}{Training Modalities (C, D)} & \multicolumn{4}{c|}{Training Modalities (C, R)} & \multicolumn{4}{c}{Training Modalities (C, U)} \\\cline{2-13}
     & \multicolumn{2}{c|}{Seen (C, D)} & \multicolumn{2}{c|}{Unseen (U, R)} & \multicolumn{2}{c|}{Seen (C, R)} & \multicolumn{2}{c|}{Unseen (D, U)} & \multicolumn{2}{c|}{Seen (C, U)} & \multicolumn{2}{c}{Unseen (D, R)} \\\cline{2-13}
     & DSC & mIoU & DSC & mIoU & DSC & mIoU & DSC & mIoU & DSC & mIoU & DSC & mIoU \\
    \hline
    baseline \cite{chen2018encoder} & 81.62 & \textcolor{blue}{\textbf{\textit{75.69}}} & 14.81 & 9.00 & 73.51 & 66.57 & 4.82 & 3.40 & 80.62 & \textcolor{blue}{\textbf{\textit{73.45}}} & 31.78 & 25.68 \\
     \hline
     \multicolumn{1}{c|}{\multirow{2}{*}{IN \cite{ulyanov2017improved}}} & 81.19 & 74.55 & 21.86 & 14.59 & \textcolor{blue}{\textbf{\textit{74.82}}} & \textcolor{blue}{\textbf{\textit{67.44}}} & 2.23 & 1.49 & 80.04 & 72.57 & 34.65 & 27.97 \\
      & \textcolor{red}{\scriptsize{\textbf{(-0.43)}}}         & \textcolor{red}{\scriptsize{\textbf{(-1.14)}}} 
      & \textcolor{ForestGreen}{\scriptsize{\textbf{(+7.05)}}} & \textcolor{ForestGreen}{\scriptsize{\textbf{(+5.59)}}} 
      & \textcolor{ForestGreen}{\scriptsize{\textbf{(+1.31)}}} & \textcolor{ForestGreen}{\scriptsize{\textbf{(+0.87)}}} 
      & \textcolor{red}{\scriptsize{\textbf{(-2.59)}}}         & \textcolor{red}{\scriptsize{\textbf{(-1.91)}}} 
      & \textcolor{red}{\scriptsize{\textbf{(-0.58)}}}         & \textcolor{red}{\scriptsize{\textbf{(-0.88)}}} 
      & \textcolor{ForestGreen}{\scriptsize{\textbf{(+2.87)}}} & \textcolor{ForestGreen}{\scriptsize{\textbf{(+2.29)}}} \\
    \hline
     \multicolumn{1}{c|}{\multirow{2}{*}{IW \cite{huang2018decorrelated}}} & 70.82 & 62.99 & 18.41 & 12.37 & 53.71 & 46.37 & 4.52 & 2.67 & 66.70 & 57.98 & 32.87 & 25.58 \\
      & \textcolor{red}{\scriptsize{\textbf{(-10.80)}}}        & \textcolor{red}{\scriptsize{\textbf{(-12.70)}}} 
      & \textcolor{ForestGreen}{\scriptsize{\textbf{(+3.60)}}} & \textcolor{ForestGreen}{\scriptsize{\textbf{(+3.37)}}} 
      & \textcolor{red}{\scriptsize{\textbf{(-19.80)}}}        & \textcolor{red}{\scriptsize{\textbf{(-20.20)}}} 
      & \textcolor{red}{\scriptsize{(\textbf{-0.30)}}}         & \textcolor{red}{\scriptsize{\textbf{(-0.73)}}} 
      & \textcolor{red}{\scriptsize{\textbf{(-13.92)}}}        & \textcolor{red}{\scriptsize{\textbf{(-15.47)}}} 
      & \textcolor{ForestGreen}{\scriptsize{\textbf{(+1.09)}}} & \textcolor{red}{\scriptsize{\textbf{(-0.10)}}} \\
    \hline
     \multicolumn{1}{c|}{\multirow{2}{*}{IBN \cite{pan2018two}}} & \textcolor{red}{\textbf{\underline{82.30}}} & \textcolor{red}{\textbf{\underline{75.72}}} & 21.93 & 15.15 & 73.92 & 66.60 & 2.50 & 1.83 & 80.62 & 73.31 & \textcolor{blue}{\textbf{\textit{37.59}}} & \textcolor{red}{\textbf{\underline{31.41}}} \\
      & \textcolor{ForestGreen}{\scriptsize{\textbf{(+0.68)}}} & \textcolor{ForestGreen}{\scriptsize{\textbf{(+0.03)}}} 
      & \textcolor{ForestGreen}{\scriptsize{\textbf{(+7.12)}}} & \textcolor{ForestGreen}{\scriptsize{\textbf{(+6.15)}}} 
      & \textcolor{ForestGreen}{\scriptsize{\textbf{(+0.41)}}} & \textcolor{ForestGreen}{\scriptsize{\textbf{(+0.03)}}} 
      & \textcolor{red}{\scriptsize{\textbf{(-2.32)}}}         & \textcolor{red}{\scriptsize{\textbf{(-1.57)}}} 
      & \scriptsize{\textbf{(+0.00)}}                          & \textcolor{red}{\scriptsize{\textbf{(-0.14)}}} 
      & \textcolor{ForestGreen}{\scriptsize{\textbf{(+5.81)}}} & \textcolor{ForestGreen}{\scriptsize{\textbf{(+5.73)}}} \\
    \hline
     \multicolumn{1}{c|}{\multirow{2}{*}{RobustNet \cite{choi2021robustnet}}} & 79.69 & 72.54 & \textcolor{blue}{\textbf{\textit{25.49}}} & \textcolor{blue}{\textbf{\textit{19.29}}} & 67.18 & 60.01 & 7.13 & 4.63 & 75.83 & 68.01 & 31.92 & 25.96 \\
      & \textcolor{red}{\scriptsize{\textbf{(-1.93)}}}          & \textcolor{red}{\scriptsize{\textbf{(-3.15)}}} 
      & \textcolor{ForestGreen}{\scriptsize{\textbf{(+10.68)}}} & \textcolor{ForestGreen}{\scriptsize{\textbf{(+10.29)}}} 
      & \textcolor{red}{\scriptsize{\textbf{(-6.33)}}}          & \textcolor{red}{\scriptsize{\textbf{(-6.56)}}} 
      & \textcolor{ForestGreen}{\scriptsize{\textbf{(+2.31)}}}  & \textcolor{ForestGreen}{\scriptsize{\textbf{(+1.23)}}} 
      & \textcolor{red}{\scriptsize{\textbf{(-4.79)}}}          & \textcolor{red}{\scriptsize{\textbf{(-5.44)}}} 
      & \textcolor{ForestGreen}{\scriptsize{\textbf{(+0.14)}}}  & \textcolor{ForestGreen}{\scriptsize{\textbf{(+0.28)}}} \\
    \hline
     \multicolumn{1}{c|}{\multirow{2}{*}{SAN-SAW \cite{peng2022semantic}}} & 75.25 & 68.14 & 17.76 & 12.65 & 68.22 & 61.31 & 4.95 & 3.49 & 70.38 & 62.50 & 28.59 & 22.18 \\
      & \textcolor{red}{\scriptsize{\textbf{(-6.37)}}}         & \textcolor{red}{\scriptsize{\textbf{(-7.55)}}} 
      & \textcolor{ForestGreen}{\scriptsize{\textbf{(+2.95)}}} & \textcolor{ForestGreen}{\scriptsize{\textbf{(+3.65)}}} 
      & \textcolor{red}{\scriptsize{\textbf{(-5.20)}}}         & \textcolor{red}{\scriptsize{\textbf{(-5.26)}}} 
      & \textcolor{ForestGreen}{\scriptsize{\textbf{(+0.13)}}} & \textcolor{ForestGreen}{\scriptsize{\textbf{(+0.09)}}} 
      & \textcolor{red}{\scriptsize{\textbf{(-10.24)}}}        & \textcolor{red}{\scriptsize{\textbf{(-10.95)}}} 
      & \textcolor{red}{\scriptsize{\textbf{(-3.19)}}}         & \textcolor{red}{\scriptsize{\textbf{(-3.50)}}} \\
    \hline
     \multicolumn{1}{c|}{\multirow{2}{*}{SPCNet \cite{huang2023style}}} & \textcolor{blue}{\textbf{\textit{81.95}}} & 75.23 & 13.97 & 8.73 & 72.61 & 65.65 & \textcolor{blue}{\textbf{\textit{8.42}}} & \textcolor{blue}{\textbf{\textit{5.71}}} & \textcolor{blue}{\textbf{\textit{80.63}}} & 73.32 & 35.56 & 28.88 \\
      & \textcolor{ForestGreen}{\scriptsize{\textbf{(+0.33)}}} & \textcolor{red}{\scriptsize{\textbf{(-0.46)}}} 
      & \textcolor{red}{\scriptsize{\textbf{(-0.84)}}}         & \textcolor{red}{\scriptsize{\textbf{(-0.27)}}} 
      & \textcolor{red}{\scriptsize{\textbf{(-0.90)}}}         & \textcolor{red}{\scriptsize{\textbf{(-0.92)}}} 
      & \textcolor{ForestGreen}{\scriptsize{\textbf{(+3.60)}}} & \textcolor{ForestGreen}{\scriptsize{\textbf{(+2.31)}}} 
      & \textcolor{ForestGreen}{\scriptsize{\textbf{(+0.01)}}} & \textcolor{red}{\scriptsize{\textbf{(-0.13)}}} 
      & \textcolor{ForestGreen}{\scriptsize{\textbf{(+3.78)}}} & \textcolor{ForestGreen}{\scriptsize{\textbf{(+3.20)}}} \\
    \hline
     \multicolumn{1}{c|}{\multirow{2}{*}{BlindNet \cite{ahn2024style}}} & 80.58 & 73.76 & 24.51 & 18.32 & 73.02 & 65.75 & 7.06 & 4.81 & 78.07 & 70.57 & 28.96 & 23.27 \\
      & \textcolor{red}{\scriptsize{\textbf{(-1.04)}}}         & \textcolor{red}{\scriptsize{\textbf{(-1.93)}}} 
      & \textcolor{ForestGreen}{\scriptsize{\textbf{(+9.70)}}} & \textcolor{ForestGreen}{\scriptsize{\textbf{(+9.32)}}} 
      & \textcolor{red}{\scriptsize{\textbf{(-0.49)}}}         & \textcolor{red}{\scriptsize{\textbf{(-0.82)}}} 
      & \textcolor{ForestGreen}{\scriptsize{\textbf{(+2.24)}}} & \textcolor{ForestGreen}{\scriptsize{\textbf{(+1.41)}}} 
      & \textcolor{red}{\scriptsize{\textbf{(-2.55)}}}         & \textcolor{red}{\scriptsize{\textbf{(-2.88)}}} 
      & \textcolor{red}{\scriptsize{\textbf{(-2.82)}}}         & \textcolor{red}{\scriptsize{\textbf{(-2.41)}}} \\
     \hline
     \multicolumn{1}{c|}{\multirow{3}{*}{\textbf{TTMG (Ours)}}} & 77.85 & 70.53 & \textcolor{red}{\textbf{\underline{25.75}}} & \textcolor{red}{\textbf{\underline{20.10}}} & \textcolor{red}{\textbf{\underline{76.12}}} & \textcolor{red}{\textbf{\underline{68.77}}} & \textcolor{red}{\textbf{\underline{19.47}}} & \textcolor{red}{\textbf{\underline{13.68}}} & \textcolor{red}{\textbf{\underline{81.19}}} & \textcolor{red}{\textbf{\underline{73.66}}} & \textcolor{red}{\textbf{\underline{39.22}}} & \textcolor{blue}{\textbf{\textit{31.15}}} \\
      & \textcolor{red}{\scriptsize{\textbf{(-3.77)}}}          & \textcolor{red}{\scriptsize{\textbf{(-5.16)}}} 
      & \textcolor{ForestGreen}{\scriptsize{\textbf{(+10.94)}}} & \textcolor{ForestGreen}{\scriptsize{\textbf{(+11.10)}}} 
      & \textcolor{ForestGreen}{\scriptsize{\textbf{(+2.61)}}}  & \textcolor{ForestGreen}{\scriptsize{\textbf{(+2.20)}}} 
      & \textcolor{ForestGreen}{\scriptsize{\textbf{(+14.65)}}} & \textcolor{ForestGreen}{\scriptsize{\textbf{(+10.28)}}} 
      & \textcolor{ForestGreen}{\scriptsize{\textbf{(+0.57)}}}  & \textcolor{ForestGreen}{\scriptsize{\textbf{(+0.21)}}} 
      & \textcolor{ForestGreen}{\scriptsize{\textbf{(+7.44)}}}  & \textcolor{ForestGreen}{\scriptsize{\textbf{(+5.47)}}} \\\cline{2-13}
     & \textbf{-4.45} & \textbf{-5.19} & \textbf{+0.26}  & \textbf{+0.81} 
     & \textbf{+1.30} & \textbf{+1.33} & \textbf{+11.05} & \textbf{+7.97} 
     & \textbf{+0.56} & \textbf{+0.21} & \textbf{+1.63} & \textbf{-0.26} \\
     \hline
     \hline
    \multicolumn{1}{c|}{\multirow{3}{*}{Method}} & \multicolumn{4}{c|}{Training Modalities (D, U)} & \multicolumn{4}{c|}{Training Modalities (D, R)} & \multicolumn{4}{c}{Training Modalities (R, U)} \\\cline{2-13}
     & \multicolumn{2}{c|}{Seen (D, U)} & \multicolumn{2}{c|}{Unseen (C, R)} & \multicolumn{2}{c|}{Seen (D, R)} & \multicolumn{2}{c|}{Unseen (C, U)} & \multicolumn{2}{c|}{Seen (R, U)} & \multicolumn{2}{c}{Unseen (C, D)} \\\cline{2-13}
     & DSC & mIoU & DSC & mIoU & DSC & mIoU & DSC & mIoU & DSC & mIoU & DSC & mIoU \\
    \hline
    baseline \cite{chen2018encoder} & 84.60 & 76.56 & 16.97 & 10.70 & 75.42 & 67.39 & 18.76 & 12.25 & 70.77 & 61.59 & 3.92 & 2.33 \\
     \hline
     \multicolumn{1}{c|}{\multirow{2}{*}{IN \cite{ulyanov2017improved}}} & 85.70 & 77.85 & 21.95 & 14.77 & 57.06 & 49.18 & 20.66 & 13.26 & 70.49 & 60.96 & 19.01 & 15.83 \\
      & \textcolor{ForestGreen}{\scriptsize{\textbf{(+1.10)}}}  & \textcolor{ForestGreen}{\scriptsize{\textbf{(+1.29)}}} 
      & \textcolor{ForestGreen}{\scriptsize{\textbf{(+4.98)}}}  & \textcolor{ForestGreen}{\scriptsize{\textbf{(+4.07)}}} 
      & \textcolor{red}{\scriptsize{\textbf{(-18.36)}}}         & \textcolor{red}{\scriptsize{\textbf{(-18.21)}}} 
      & \textcolor{ForestGreen}{\scriptsize{\textbf{(+1.90)}}}  & \textcolor{ForestGreen}{\scriptsize{\textbf{(+1.01)}}} 
      & \textcolor{red}{\scriptsize{\textbf{(-0.28)}}}          & \textcolor{red}{\scriptsize{\textbf{(-0.63)}}} 
      & \textcolor{ForestGreen}{\scriptsize{\textbf{(+15.09)}}} & \textcolor{ForestGreen}{\scriptsize{\textbf{(+13.50)}}} \\
    \hline
     \multicolumn{1}{c|}{\multirow{2}{*}{IW \cite{huang2018decorrelated}}} & 81.15 & 72.05 & 14.09 & 8.84 & 62.49 & 56.17 & 22.83 & 15.33 & 56.05 & 48.01 & 17.64 & 13.43 \\
      & \textcolor{red}{\scriptsize{\textbf{(-3.45)}}}          & \textcolor{red}{\scriptsize{\textbf{(-4.51)}}} 
      & \textcolor{red}{\scriptsize{\textbf{(-2.88)}}}          & \textcolor{red}{\scriptsize{\textbf{(-1.86)}}} 
      & \textcolor{red}{\scriptsize{\textbf{(-12.93)}}}         &  \textcolor{red}{\scriptsize{\textbf{(-11.22)}}} 
      &  \textcolor{ForestGreen}{\scriptsize{\textbf{(+4.07)}}} &  \textcolor{ForestGreen}{\scriptsize{\textbf{(+3.08)}}} 
      & \textcolor{red}{\scriptsize{\textbf{(-14.72)}}}         & \textcolor{red}{\scriptsize{\textbf{(-13.58)}}} 
      & \textcolor{ForestGreen}{\scriptsize{\textbf{(+13.72)}}} & \textcolor{ForestGreen}{\scriptsize{\textbf{(+11.10)}}} \\
    \hline
     \multicolumn{1}{c|}{\multirow{2}{*}{IBN \cite{pan2018two}}} & \textcolor{red}{\textbf{\underline{86.38}}} & \textcolor{red}{\textbf{\underline{78.42}}} & \textcolor{blue}{\textbf{\textit{22.41}}} & 14.39 & \textcolor{blue}{\textbf{\textit{75.84}}} & \textcolor{blue}{\textbf{\textit{67.81}}} & 22.83 & 15.87 & \textcolor{blue}{\textbf{\textit{71.77}}} & \textcolor{blue}{\textbf{\textit{62.38}}} & 19.76 & 15.81 \\
      & \textcolor{ForestGreen}{\scriptsize{\textbf{(+1.78)}}}  & \textcolor{ForestGreen}{\scriptsize{\textbf{(+1.86)}}} 
      & \textcolor{ForestGreen}{\scriptsize{\textbf{(+5.44)}}}  & \textcolor{ForestGreen}{\scriptsize{\textbf{(+3.69)}}} 
      & \textcolor{ForestGreen}{\scriptsize{\textbf{(+0.42)}}}  & \textcolor{ForestGreen}{\scriptsize{\textbf{(+0.42)}}} 
      & \textcolor{ForestGreen}{\scriptsize{\textbf{(+4.07)}}}  & \textcolor{ForestGreen}{\scriptsize{\textbf{(+3.62)}}} 
      & \textcolor{ForestGreen}{\scriptsize{\textbf{(+1.00)}}}  & \textcolor{ForestGreen}{\scriptsize{\textbf{(+0.79)}}} 
      & \textcolor{ForestGreen}{\scriptsize{\textbf{(+15.84)}}} & \textcolor{ForestGreen}{\scriptsize{\textbf{(+13.48)}}} \\
    \hline
     \multicolumn{1}{c|}{\multirow{2}{*}{RobustNet \cite{choi2021robustnet}}} & 84.21 & 75.95 & 18.86 & 12.87 & 69.28 & 61.64 & 24.19 & 18.15 & 63.79 & 55.14 & 16.47 & 13.42 \\
      & \textcolor{red}{\scriptsize{\textbf{(-0.39)}}}          & \textcolor{red}{\scriptsize{\textbf{(-0.61)}}} 
      & \textcolor{ForestGreen}{\scriptsize{\textbf{(+1.89)}}}  & \textcolor{ForestGreen}{\scriptsize{\textbf{(+2.17)}}} 
      & \textcolor{red}{\scriptsize{\textbf{(-6.14)}}}          & \textcolor{red}{\scriptsize{\textbf{(-5.75)}}} 
      & \textcolor{ForestGreen}{\scriptsize{\textbf{(+5.43)}}}  & \textcolor{ForestGreen}{\scriptsize{\textbf{(+5.90)}}} 
      & \textcolor{red}{\scriptsize{\textbf{(-6.98)}}}          & \textcolor{red}{\scriptsize{\textbf{(-6.45)}}} 
      & \textcolor{ForestGreen}{\scriptsize{\textbf{(+12.55)}}} & \textcolor{ForestGreen}{\scriptsize{\textbf{(+11.09)}}} \\
    \hline
     \multicolumn{1}{c|}{\multirow{2}{*}{SAN-SAW \cite{peng2022semantic}}} & 81.65 & 72.24 & 19.36 & 12.06 & 67.62 & 60.41 & 23.20 & 15.60 & 66.64 & 57.97 & 5.33 & 3.83 \\
      & \textcolor{red}{\scriptsize{\textbf{(-2.95)}}}         & \textcolor{red}{\scriptsize{\textbf{(-4.32)}}} 
      & \textcolor{ForestGreen}{\scriptsize{\textbf{(+2.39)}}} & \textcolor{ForestGreen}{\scriptsize{\textbf{(+1.36)}}} 
      & \textcolor{red}{\scriptsize{\textbf{(-7.80)}}}         & \textcolor{red}{\scriptsize{\textbf{(-6.98)}}} 
      & \textcolor{ForestGreen}{\scriptsize{\textbf{(+4.44)}}} &  \textcolor{ForestGreen}{\scriptsize{\textbf{(+3.35)}}} 
      & \textcolor{red}{\scriptsize{\textbf{(-4.13)}}}         & \textcolor{red}{\scriptsize{\textbf{(-3.62)}}} 
      & \textcolor{ForestGreen}{\scriptsize{\textbf{(+1.41)}}} & \textcolor{ForestGreen}{\scriptsize{\textbf{(+1.50)}}} \\
    \hline
     \multicolumn{1}{c|}{\multirow{2}{*}{SPCNet \cite{huang2023style}}} & 85.99 & 78.23 & 21.67 & 14.71 & 75.75 & 67.59 & 20.34 & 13.66 & 70.72 & 61.27 & 16.69 & 13.15 \\
      & \textcolor{ForestGreen}{\scriptsize{\textbf{(+1.39)}}}  & \textcolor{ForestGreen}{\scriptsize{\textbf{(+1.67)}}} 
      & \textcolor{ForestGreen}{\scriptsize{\textbf{(+4.70)}}}  & \textcolor{ForestGreen}{\scriptsize{\textbf{(+4.01)}}} 
      & \textcolor{ForestGreen}{\scriptsize{\textbf{(+0.33)}}}  &  \textcolor{ForestGreen}{\scriptsize{\textbf{(+0.20)}}} 
      & \textcolor{ForestGreen}{\scriptsize{\textbf{(+1.58)}}}  &  \textcolor{ForestGreen}{\scriptsize{\textbf{(+1.41)}}} 
      & \textcolor{red}{\scriptsize{\textbf{(-0.05)}}}          & \textcolor{red}{\scriptsize{\textbf{(-0.32)}}} 
      & \textcolor{ForestGreen}{\scriptsize{\textbf{(+12.77)}}} & \textcolor{ForestGreen}{\scriptsize{\textbf{(+10.82)}}} \\
    \hline
     \multicolumn{1}{c|}{\multirow{2}{*}{BlindNet \cite{ahn2024style}}} & 84.86 & 77.02 & 21.30 & \textcolor{blue}{\textbf{\textit{15.28}}} & 71.85 & 63.93 & \textcolor{blue}{\textbf{\textit{25.96}}} & \textcolor{blue}{\textbf{\textit{18.73}}} & 66.07 & 57.35 & \textcolor{blue}{\textbf{\textit{20.12}}}  & \textcolor{blue}{\textbf{\textit{16.85}}} \\
      & \textcolor{ForestGreen}{\scriptsize{\textbf{(+0.26)}}}  & \textcolor{ForestGreen}{\scriptsize{\textbf{(+0.46)}}} 
      & \textcolor{ForestGreen}{\scriptsize{\textbf{(+4.33)}}}  & \textcolor{ForestGreen}{\scriptsize{\textbf{(+4.58)}}} 
      & \textcolor{red}{\scriptsize{\textbf{(-3.57)}}}          &  \textcolor{red}{\scriptsize{\textbf{(-3.46)}}} 
      & \textcolor{ForestGreen}{\scriptsize{\textbf{(+7.20)}}}  &  \textcolor{ForestGreen}{\scriptsize{\textbf{(+6.48)}}} 
      & \textcolor{red}{\scriptsize{\textbf{(-4.70)}}}          & \textcolor{red}{\scriptsize{\textbf{(-4.24)}}} 
      & \textcolor{ForestGreen}{\scriptsize{\textbf{(+16.20)}}} & \textcolor{ForestGreen}{\scriptsize{\textbf{(+14.52)}}} \\
     \hline
     \multicolumn{1}{c|}{\multirow{3}{*}{\textbf{TTMG (Ours)}}} & \textcolor{blue}{\textbf{\textit{86.26}}} & \textcolor{blue}{\textbf{\textit{78.39}}} & \textcolor{red}{\textbf{\underline{25.67}}} & \textcolor{red}{\textbf{\underline{17.90}}} & \textcolor{red}{\textbf{\underline{81.29}}} & \textcolor{red}{\textbf{\underline{73.34}}} & \textcolor{red}{\textbf{\underline{27.30}}} & \textcolor{red}{\textbf{\underline{18.98}}} &  \textcolor{red}{\textbf{\underline{73.62}}} &  \textcolor{red}{\textbf{\underline{65.22}}} &  \textcolor{red}{\textbf{\underline{20.96}}} &  \textcolor{red}{\textbf{\underline{16.98}}} \\
      & \textcolor{ForestGreen}{\scriptsize{\textbf{(+1.66)}}}  & \textcolor{ForestGreen}{\scriptsize{\textbf{(+1.83)}}} 
      & \textcolor{ForestGreen}{\scriptsize{\textbf{(+8.70)}}}  & \textcolor{ForestGreen}{\scriptsize{\textbf{(+7.20)}}} 
      & \textcolor{ForestGreen}{\scriptsize{\textbf{(+5.87)}}}  & \textcolor{ForestGreen}{\scriptsize{\textbf{(+5.95)}}} 
      & \textcolor{ForestGreen}{\scriptsize{\textbf{(+8.54)}}}  & \textcolor{ForestGreen}{\scriptsize{\textbf{(+6.73)}}} 
      & \textcolor{ForestGreen}{\scriptsize{\textbf{(+2.85)}}}  & \textcolor{ForestGreen}{\scriptsize{\textbf{(+3.63)}}} 
      & \textcolor{ForestGreen}{\scriptsize{\textbf{(+17.04)}}} & \textcolor{ForestGreen}{\scriptsize{\textbf{(+14.65)}}} \\\cline{2-13}
     & \textbf{-0.12} & \textbf{-0.03} & \textbf{+3.26} & \textbf{+2.62} 
     & \textbf{+5.45} & \textbf{+5.53} & \textbf{+1.34} & \textbf{+0.25} 
     & \textbf{+1.85} & \textbf{+2.84} & \textbf{+0.84} & \textbf{+0.13} \\
     \hline
    \end{tabular}
    \caption{Segmentation results on two modality training schemes (C, D), (C, R), (C, U), (D, U), (D, R), and (R, U).}
    \label{tab:comparison_two_modality}
\end{table*}

\subsection{Result Analysis}
\noindent \textbf{Quantitative Results.} We used the same model as that employed for the ’Seen’ modalities in Table \ref{tab:comparison_three_modality} and Table \ref{tab:comparison_two_modality} to evaluate the domain generalization performance for ’Unseen’ modalities for each table. We first trained each model on three modalities ((C, U, D), (C, U, R), (C, D, R), (U, D, R)) and then evaluated the performance on seen and unseen modality datasets. As shown in Table \ref{tab:comparison_three_modality}, TTMG has significantly improved the unseen modality generalization performance compared to the baseline. When compared to IN, IW, and IBN, TTMG exhibited DSC improvement of 7.63\%, 6.00\%, and 5.61\%, respectively, on average. Additionally, compared to SPCNet, which used the style projection approach based on prototype learning, TTMG demonstrated DSC improvement of 0.86\%, and 8.27\% on seen and unseen modality datasets on average. Furthermore, we trained each model on the more limited training scenario where the number of seen modalities was reduced to two ((C, D), (C, R), (C, U), (D, U), (D, R), (R, U)) and then evaluated them. As shown in Table \ref{tab:comparison_two_modality}, TTMG still outperforms various modalities settings even limited modality scenarios. Most notably, only TTMG and RobustNet improved generalization performance on the unseen modality dataset compared to the baseline in all scenarios. However, RobustNet shows severe performance degradation in all cases on the seen modality dataset. As shown in the Appendix (Table \ref{tab:efficiency_analysis}), TTMG performs without substantial computational cost and parameters. Additionally, TTMG still achieves higher performance in both seen and unseen modalities compared to existing DG methods even on different backbones (MobileNetV2 \cite{sandler2018mobilenetv2} and ShuffleNetV2 \cite{ma2018shufflenet}) and TransUNet \cite{chen2021transunet} in the Appendix (Table \ref{tab:backbone_type}). These results highlight that employing a prototype learning-based style-projection (MASP) with MSIW is versatile and crucial for efficiently generalizing unseen modalities in medical image segmentation.

\noindent \textbf{Qualitative Results.} Figure \ref{fig:QualitativeResults} illustrates qualitative results across different methods for radiology and ultrasound modalities. The baseline provides adequately seen modality datasets while producing highly noisy predictions on unseen modalities.  Although IN, IW, and IBN standardize feature map distributions between seen and unseen modalities, they still produce noisy predictions and demonstrate degraded performance on seen modalities. RobustNet, which removes domain-sensitive information through visual transformation, does not address modality-sensitive information, resulting in suboptimal performance on unseen modalities. Although SPCNet is similar to our approach (prototype learning-based style projection), semantic clustering in the decoder struggles to capture complex anatomical structures in medical images. In contrast, TTMG achieves robust predictions across all modalities, effectively handling noise, varying lesion sizes, and complex anatomical structures by leveraging the dual capabilities of MASP and MSIW. Due to the page limit, we also provide more qualitative results on the other various training scenarios in Appendix.

\begin{figure}[t]
    \centering
    \includegraphics[width=0.48\textwidth]{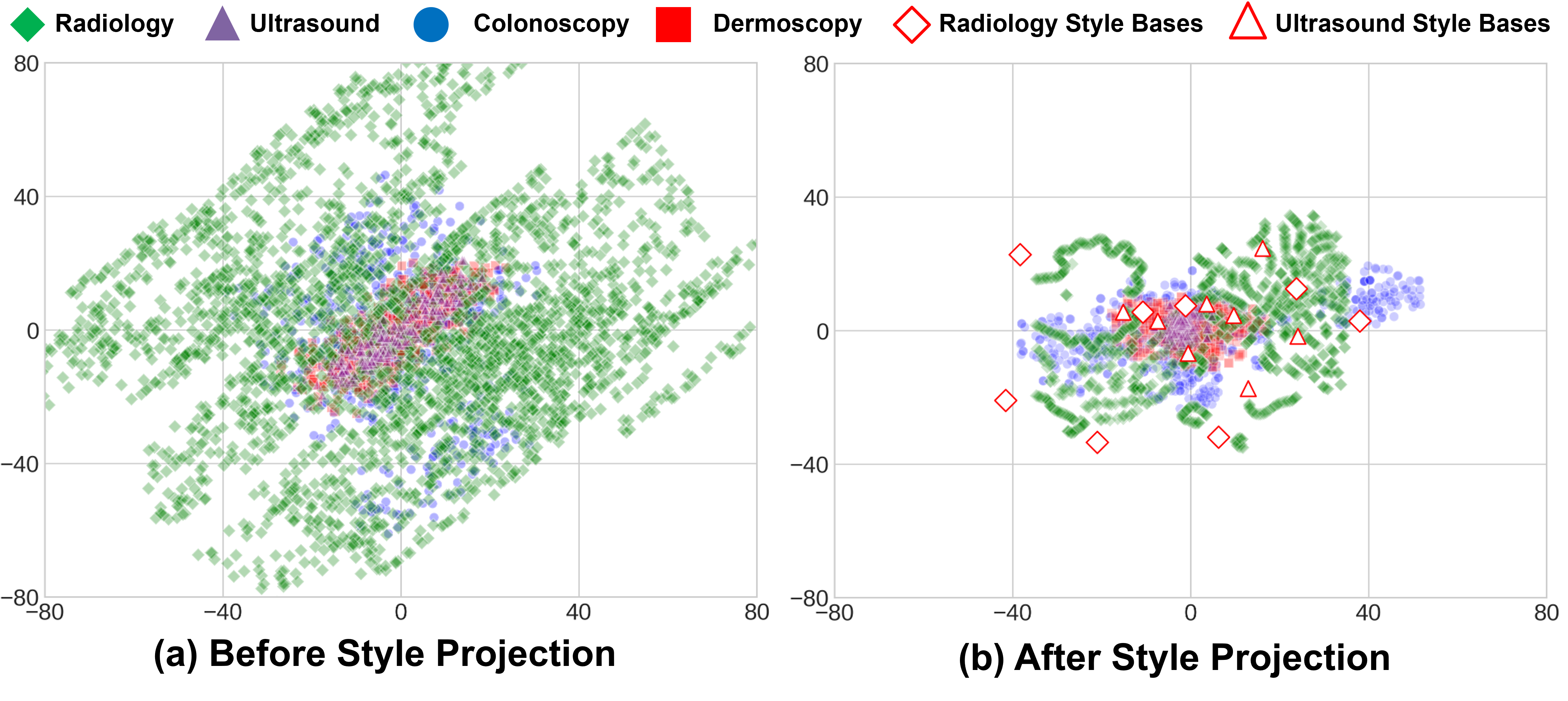}
    \caption{T-SNE visualization \cite{van2008visualizing} of features for different modalities (a) before and (b) after test-time style projection from Stage1.}
    \label{fig:FeatureProjection}
\end{figure}

\begin{figure}[t]
    \centering
    \includegraphics[width=0.48\textwidth]{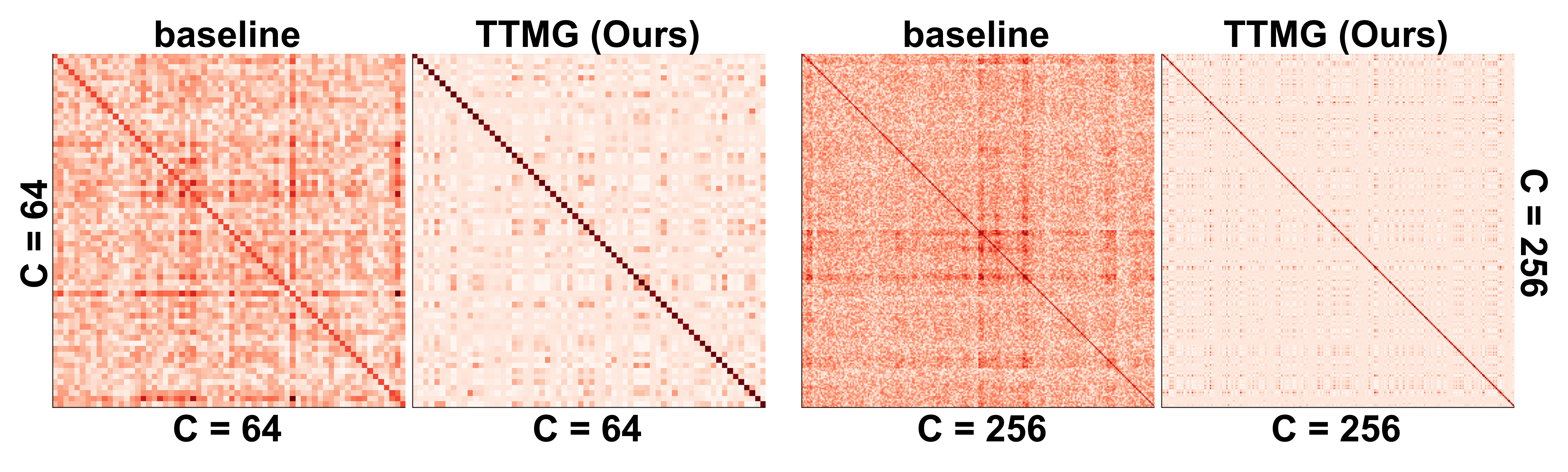}
    \caption{Visualization of covariance matrix extracted from baseline and TTMG.}
    \label{fig:FeatureCovariance}
\end{figure}

\noindent \textbf{Feature Visualization.} We employed T-SNE \cite{van2008visualizing} visualization in the (R, U) to (C, D) training scenario to illustrate how data achieves unseen modality generalization. Figure \ref{fig:FeatureProjection} shows the feature distributions across various modalities, where MASP effectively projects the feature representations around the style bases of seen modalities. This clustering effect compensates for the characteristic differences between modalities, facilitating more robust generalization to unseen modalities. Additionally, we observed that the baseline model exhibits high feature covariance, whereas TTMG selectively maintains lower feature covariance as shown in Figure \ref{fig:FeatureCovariance}. This effect aligns well with the objective of MSIW, effectively enhancing generalization.

\subsection{Ablation Study}

We want to clarify that the same settings that mentioned in Section \ref{s42_implementation_details} were applied across all ablation studies.

\begin{table}[t]
    \centering
    \footnotesize
    \setlength\tabcolsep{5.0pt} 
    \begin{tabular}{c|c|cc|cc}
    \hline
    \multicolumn{1}{c|}{\multirow{3}{*}{Setting}} & \multicolumn{1}{c|}{\multirow{3}{*}{Configuration}} & \multicolumn{4}{c}{Training Modalities (R, U)} \\ \cline{3-6}
     & &  \multicolumn{2}{c|}{Seen (R, U)} & \multicolumn{2}{c}{Unseen (C, D)} \\ \cline{3-6}
     & &  DSC & mIoU & DSC & mIoU \\
     \hline
     S0 & baseline & 70.77 & 61.59 & 3.92 & 2.33 \\
     \hline
     \multicolumn{1}{c|}{\multirow{2}{*}{S1}} &  \multicolumn{1}{c|}{\multirow{2}{*}{+ MASP \scriptsize{(w/o $\mathcal{L}_{\text{con}}$)}}} & 73.17 & 64.53 & 16.31 & 12.38 \\
      & & \textcolor{ForestGreen}{\scriptsize{\textbf{(+2.40)}}} & \textcolor{ForestGreen}{\scriptsize{\textbf{(+2.94)}}} & \textcolor{ForestGreen}{\scriptsize{\textbf{(+12.39)}}} & \textcolor{ForestGreen}{\scriptsize{\textbf{(+10.05)}}} \\
      \hline
     \multicolumn{1}{c|}{\multirow{2}{*}{S2}} &  \multicolumn{1}{c|}{\multirow{2}{*}{+ MASP \scriptsize{(w $\mathcal{L}_{\text{con}}$)}}} & \textcolor{red}{\textbf{\underline{73.87}}} & \textcolor{red}{\textbf{\underline{65.23}}} & 16.64 & 12.90 \\
      & & \textcolor{ForestGreen}{\scriptsize{\textbf{(+3.10)}}} & \textcolor{ForestGreen}{\scriptsize{\textbf{(+3.64)}}} & \textcolor{ForestGreen}{\scriptsize{\textbf{(+12.72)}}} & \textcolor{ForestGreen}{\scriptsize{\textbf{(+10.57)}}} \\
     \hline
     \multicolumn{1}{c|}{\multirow{2}{*}{S3}} &  \multicolumn{1}{c|}{\multirow{2}{*}{+ MSIW}} & 73.50 & 64.87 & \textcolor{blue}{\textbf{\textit{18.94}}} & \textcolor{blue}{\textbf{\textit{15.18}}} \\
      & & \textcolor{ForestGreen}{\scriptsize{\textbf{(+2.73)}}} & \textcolor{ForestGreen}{\scriptsize{\textbf{(+3.28)}}} & \textcolor{ForestGreen}{\scriptsize{\textbf{(+15.02)}}} & \textcolor{ForestGreen}{\scriptsize{\textbf{(+12.85)}}} \\
     \hline
     \multicolumn{1}{c|}{\multirow{3}{*}{S4}} & \multicolumn{1}{c|}{\multirow{3}{*}{\textbf{TTMG \scriptsize{(Ours)}}}} & \textcolor{blue}{\textbf{\textit{73.62}}} &  \textcolor{blue}{\textbf{\textit{65.22}}} &  \textcolor{red}{\textbf{\underline{20.96}}} &  \textcolor{red}{\textbf{\underline{16.98}}} \\
      & & \textcolor{ForestGreen}{\scriptsize{\textbf{(+2.85)}}} & \textcolor{ForestGreen}{\scriptsize{\textbf{(+3.63)}}} & \textcolor{ForestGreen}{\scriptsize{\textbf{(+17.04)}}} & \textcolor{ForestGreen}{\scriptsize{\textbf{(+14.66)}}} \\ \cline{3-6}
      & & \textbf{-0.25} & \textbf{-0.01} & \textbf{+2.02} & \textbf{+1.80} \\
     \hline
    \end{tabular}
    \caption{Ablation study of TTMG components (MASP and MSIW) on seen (R, U) and unseen (C, D) modalities. } 
    \label{tab:effectiveness_masp_msiw}
\end{table}

\noindent \textbf{Effectiveness of MASP and MSIW.} In this section, we conducted ablation studies on the (R, U) to (C, D) training scenario to demonstrate the effectiveness of MASP and MSIW, which are core components of the TTMG framework. As listed in Table \ref{tab:effectiveness_masp_msiw}, our approach (S4) exhibited the best performance on unseen modalities. The most notable result is that the application of MASP regardless of $\mathcal{L}_{\text{con}}$ provides the significant DSC improvement of 12.39\% and 12.72\% compared to baseline (S0). Compared to S1 and S2, applying $\mathcal{L}_{\text{con}}$ achieves performance improvement in both seen and unseen modalities because the content of the feature can be maintained after the projection. Additionally, applying MSIW alone (S3) resulted in a significant DSC and mIoU improvement of 15.02\% and 12.85\%, respectively, for unseen modalities, underscoring MSIW's role in reducing modality-sensitive information. Consequently, TTMG (S2 + S3), which employs the dual utilization of MASP and MSIW with $\mathcal{L}_{\text{con}}$, performs significantly better generalization ability on unseen modality datasets.

\begin{table}[t]
    \centering
    \footnotesize
    \setlength\tabcolsep{7.0pt} 
    \begin{tabular}{c|cc|cc}
    \hline
    \multicolumn{1}{c|}{\multirow{3}{*}{Number of Style Bases $K_{m}$}} & \multicolumn{4}{c}{Training Modalities (R, U)} \\ \cline{2-5}
     & \multicolumn{2}{c|}{Seen (R, U)} & \multicolumn{2}{c}{Unseen (C, D)} \\ \cline{2-5}
     & DSC & mIoU & DSC & mIoU \\
     \hline
     $K_{m} = 4$ & 73.21 & 64.46 & \textcolor{blue}{\textbf{\textit{18.93}}} & \textcolor{blue}{\textbf{\textit{15.17}}} \\
     \textbf{$K_{m} = 8$ \scriptsize{(Ours)}} & \textcolor{blue}{\textbf{\textit{73.62}}} & \textcolor{blue}{\textbf{\textit{65.22}}} & \textcolor{red}{\textbf{\underline{20.96}}} & \textcolor{red}{\textbf{\underline{16.98}}} \\
     $K_{m} = 16$ & \textcolor{red}{\textbf{\underline{75.02}}} & \textcolor{red}{\textbf{\underline{66.34}}} & 18.04 & 14.27 \\
     $K_{m} = 32$ & 73.13 & 64.57 & 17.65 & 12.76 \\
     $K_{m} = 64$ & 72.21 & 63.43 & 14.77 & 70.72 \\
     \hline
    \end{tabular}
    \caption{Effect of the number of style bases \( K_{m} \) on TTMG’s generalization performance across seen (R, U) and unseen (C, D) modalities.}
    \label{tab:number_of_style_bases}
\end{table}

\noindent \textbf{Number of Style Bases.} In this section, we investigate the impact of the number of style bases $K_{m}$ on the generalization ability of TTMG in the (R, U) to (C, D) training scenario. As shown in Table \ref{tab:number_of_style_bases}, TTMG achieves the highest generalization performance on unseen modality datasets with $K_{m} = 8$, while also maintaining strong performance on seen modality datasets. With a lower number of style bases, such as $K_{m} = 4$, the limited bases fail to capture the complex style distribution of medical image datasets, resulting in lower performance. Conversely, as $K_{m}$ increases to 16, 32, and 64, performance on seen modalities improves; however, generalization on unseen modalities decreases, likely due to overfitting to the seen modality style bases. Based on these results, we set $K_{m} = 8$ for all experiments. \vspace{-0.25cm}
\section{Conclusion}

Based on extensive experiments across various training scenarios and ablation studies, we summarize the effectiveness of TTMG as follows: 1) Developing a modality-generalizable medical image segmentation model requires addressing the distributional differences in features across modalities. 2) MASP with content consistency loss is essential for aligning feature distributions across different modality datasets while maintaining performance on seen modalities. 3) MSIW selectively whitens modality-sensitive features, effectively mitigating overfitting to specific modalities and preserving critical modality-invariant information. In conclusion, we propose a novel framework, called \textbf{Test-Time Modality Generalization (TTMG)}, to tackle the challenge of generalizing to unseen modalities in medical image segmentation. TTMG integrates two core components—MASP and MSIW—to enhance generalization to unseen modalities \textbf{without requiring additional retraining}. Our rigorous experiments demonstrate that TTMG significantly improves performance on unseen modalities, providing a scalable and efficient solution for medical image segmentation in real-world applications. This work lays a solid foundation for future research in modality generalizable medical image segmentation, and we plan to extend TTMG to support broader training scenarios involving multiple modality sources including single modality training scheme.
{
    \small
    \bibliographystyle{ieeenat_fullname}
    \bibliography{main}

\begin{thebibliography}{84}
\providecommand{\natexlab}[1]{#1}
\providecommand{\url}[1]{\texttt{#1}}
\expandafter\ifx\csname urlstyle\endcsname\relax
  \providecommand{\doi}[1]{doi: #1}\else
  \providecommand{\doi}{doi: \begingroup \urlstyle{rm}\Url}\fi

\bibitem[COV()]{COVID19_2}
Covid19 dataset.
\newblock \url{https://www.kaggle.com/datasets/piyushsamant11/pidata-new-names}.

\bibitem[Ahn et~al.(2024)Ahn, Yang, Choi, and Lim]{ahn2024style}
Woo-Jin Ahn, Geun-Yeong Yang, Hyun-Duck Choi, and Myo-Taeg Lim.
\newblock Style blind domain generalized semantic segmentation via covariance alignment and semantic consistence contrastive learning.
\newblock In \emph{Proceedings of the IEEE/CVF Conference on Computer Vision and Pattern Recognition}, pages 3616--3626, 2024.

\bibitem[Al-Dhabyani et~al.(2020)Al-Dhabyani, Gomaa, Khaled, and Fahmy]{al2020dataset}
Walid Al-Dhabyani, Mohammed Gomaa, Hussien Khaled, and Aly Fahmy.
\newblock Dataset of breast ultrasound images.
\newblock \emph{Data in brief}, 28:\penalty0 104863, 2020.

\bibitem[Bastico et~al.(2023)Bastico, Ryckelynck, Cort{\'e}, Tillier, and Decenci{\`e}re]{bastico2023simple}
Matteo Bastico, David Ryckelynck, Laurent Cort{\'e}, Yannick Tillier, and Etienne Decenci{\`e}re.
\newblock A simple and robust framework for cross-modality medical image segmentation applied to vision transformers.
\newblock In \emph{Proceedings of the IEEE/CVF International Conference on Computer Vision}, pages 4128--4138, 2023.

\bibitem[Bayram and Ahmed(2023)]{bayram2023domain}
Firas Bayram and Bestoun~S Ahmed.
\newblock A domain-region based evaluation of ml performance robustness to covariate shift.
\newblock \emph{Neural Computing and Applications}, 35\penalty0 (24):\penalty0 17555--17577, 2023.

\bibitem[Bernal et~al.(2015)Bernal, S{\'a}nchez, Fern{\'a}ndez-Esparrach, Gil, Rodr{\'\i}guez, and Vilari{\~n}o]{bernal2015wm}
Jorge Bernal, F~Javier S{\'a}nchez, Gloria Fern{\'a}ndez-Esparrach, Debora Gil, Cristina Rodr{\'\i}guez, and Fernando Vilari{\~n}o.
\newblock Wm-dova maps for accurate polyp highlighting in colonoscopy: Validation vs. saliency maps from physicians.
\newblock \emph{Computerized medical imaging and graphics}, 43:\penalty0 99--111, 2015.

\bibitem[Cao et~al.(2022)Cao, Wang, Chen, Jiang, Zhang, Tian, and Wang]{cao2022swin}
Hu Cao, Yueyue Wang, Joy Chen, Dongsheng Jiang, Xiaopeng Zhang, Qi Tian, and Manning Wang.
\newblock Swin-unet: Unet-like pure transformer for medical image segmentation.
\newblock In \emph{European conference on computer vision}, pages 205--218. Springer, 2022.

\bibitem[Chen et~al.(2021)Chen, Lu, Yu, Luo, Adeli, Wang, Lu, Yuille, and Zhou]{chen2021transunet}
Jieneng Chen, Yongyi Lu, Qihang Yu, Xiangde Luo, Ehsan Adeli, Yan Wang, Le Lu, Alan~L Yuille, and Yuyin Zhou.
\newblock Transunet: Transformers make strong encoders for medical image segmentation.
\newblock \emph{arXiv preprint arXiv:2102.04306}, 2021.

\bibitem[Chen et~al.(2018)Chen, Zhu, Papandreou, Schroff, and Adam]{chen2018encoder}
Liang-Chieh Chen, Yukun Zhu, George Papandreou, Florian Schroff, and Hartwig Adam.
\newblock Encoder-decoder with atrous separable convolution for semantic image segmentation.
\newblock In \emph{Proceedings of the European conference on computer vision (ECCV)}, pages 801--818, 2018.

\bibitem[Chen et~al.(2024)Chen, Zheng, Li, Ma, Ma, Li, and Fan]{chen2024versatile}
Xiaoyang Chen, Hao Zheng, Yuemeng Li, Yuncong Ma, Liang Ma, Hongming Li, and Yong Fan.
\newblock Versatile medical image segmentation learned from multi-source datasets via model self-disambiguation.
\newblock In \emph{Proceedings of the IEEE/CVF Conference on Computer Vision and Pattern Recognition}, pages 11747--11756, 2024.

\bibitem[Choi et~al.(2021)Choi, Jung, Yun, Kim, Kim, and Choo]{choi2021robustnet}
Sungha Choi, Sanghun Jung, Huiwon Yun, Joanne~T Kim, Seungryong Kim, and Jaegul Choo.
\newblock Robustnet: Improving domain generalization in urban-scene segmentation via instance selective whitening.
\newblock In \emph{Proceedings of the IEEE/CVF conference on computer vision and pattern recognition}, pages 11580--11590, 2021.

\bibitem[Coates et~al.(2015)Coates, Winer, Goldhirsch, Gelber, Gnant, Piccart-Gebhart, Th{\"u}rlimann, Senn, Members, Andr{\'e}, et~al.]{coates2015tailoring}
Alan~S Coates, Eric~P Winer, Aron Goldhirsch, Richard~D Gelber, Michael Gnant, M Piccart-Gebhart, Beat Th{\"u}rlimann, H-J Senn, Panel Members, Fabrice Andr{\'e}, et~al.
\newblock Tailoring therapies—improving the management of early breast cancer: St gallen international expert consensus on the primary therapy of early breast cancer 2015.
\newblock \emph{Annals of oncology}, 26\penalty0 (8):\penalty0 1533--1546, 2015.

\bibitem[Darzi and Bocklitz(2024)]{darzi2024review}
Fatemehzahra Darzi and Thomas Bocklitz.
\newblock A review of medical image registration for different modalities.
\newblock \emph{Bioengineering}, 11\penalty0 (8):\penalty0 786, 2024.

\bibitem[Din et~al.(2024)Din, Mourad, and Serpedin]{din2024lscs}
Sadia Din, Omar Mourad, and Erchin Serpedin.
\newblock Lscs-net: A lightweight skin cancer segmentation network with densely connected multi-rate atrous convolution.
\newblock \emph{Computers in Biology and Medicine}, 173:\penalty0 108303, 2024.

\bibitem[Dosovitskiy et~al.(2021)Dosovitskiy, Beyer, Kolesnikov, Weissenborn, Zhai, Unterthiner, Dehghani, Minderer, Heigold, Gelly, Uszkoreit, and Houlsby]{dosovitskiy2021an}
Alexey Dosovitskiy, Lucas Beyer, Alexander Kolesnikov, Dirk Weissenborn, Xiaohua Zhai, Thomas Unterthiner, Mostafa Dehghani, Matthias Minderer, Georg Heigold, Sylvain Gelly, Jakob Uszkoreit, and Neil Houlsby.
\newblock An image is worth 16x16 words: Transformers for image recognition at scale.
\newblock In \emph{International Conference on Learning Representations}, 2021.

\bibitem[Fan et~al.(2017)Fan, Cheng, Liu, Li, and Borji]{fan2017structure}
Deng-Ping Fan, Ming-Ming Cheng, Yun Liu, Tao Li, and Ali Borji.
\newblock Structure-measure: A new way to evaluate foreground maps.
\newblock In \emph{Proceedings of the IEEE international conference on computer vision}, pages 4548--4557, 2017.

\bibitem[Fan et~al.(2018)Fan, Gong, Cao, Ren, Cheng, and Borji]{fan2018enhanced}
Deng-Ping Fan, Cheng Gong, Yang Cao, Bo Ren, Ming-Ming Cheng, and Ali Borji.
\newblock Enhanced-alignment measure for binary foreground map evaluation.
\newblock \emph{arXiv preprint arXiv:1805.10421}, 2018.

\bibitem[Fan et~al.(2020{\natexlab{a}})Fan, Ji, Zhou, Chen, Fu, Shen, and Shao]{fan2020pranet}
Deng-Ping Fan, Ge-Peng Ji, Tao Zhou, Geng Chen, Huazhu Fu, Jianbing Shen, and Ling Shao.
\newblock Pranet: Parallel reverse attention network for polyp segmentation.
\newblock In \emph{International conference on medical image computing and computer-assisted intervention}, pages 263--273. Springer, 2020{\natexlab{a}}.

\bibitem[Fan et~al.(2020{\natexlab{b}})Fan, Zhou, Ji, Zhou, Chen, Fu, Shen, and Shao]{fan2020inf}
Deng-Ping Fan, Tao Zhou, Ge-Peng Ji, Yi Zhou, Geng Chen, Huazhu Fu, Jianbing Shen, and Ling Shao.
\newblock Inf-net: Automatic covid-19 lung infection segmentation from ct images.
\newblock \emph{IEEE transactions on medical imaging}, 39\penalty0 (8):\penalty0 2626--2637, 2020{\natexlab{b}}.

\bibitem[Gao et~al.(2022)Gao, Zhang, Hong, Zhang, Fan, and Yan]{gao2022towards}
Yangcheng Gao, Zhao Zhang, Richang Hong, Haijun Zhang, Jicong Fan, and Shuicheng Yan.
\newblock Towards feature distribution alignment and diversity enhancement for data-free quantization.
\newblock In \emph{2022 IEEE International Conference on Data Mining (ICDM)}, pages 141--150. IEEE, 2022.

\bibitem[Gaube et~al.(2021)Gaube, Suresh, Raue, Merritt, Berkowitz, Lermer, Coughlin, Guttag, Colak, and Ghassemi]{gaube2021ai}
Susanne Gaube, Harini Suresh, Martina Raue, Alexander Merritt, Seth~J Berkowitz, Eva Lermer, Joseph~F Coughlin, John~V Guttag, Errol Colak, and Marzyeh Ghassemi.
\newblock Do as ai say: susceptibility in deployment of clinical decision-aids.
\newblock \emph{NPJ digital medicine}, 4\penalty0 (1):\penalty0 31, 2021.

\bibitem[Gu et~al.(2023)]{gu2024train}
Shi Gu et~al.
\newblock Train once, deploy anywhere: Edge-guided single-source domain generalization for medical image segmentation.
\newblock In \emph{Medical Imaging with Deep Learning}, 2023.

\bibitem[Guo et~al.(2019)Guo, Li, Huang, Guo, and Li]{guo2019deep}
Zhe Guo, Xiang Li, Heng Huang, Ning Guo, and Quanzheng Li.
\newblock Deep learning-based image segmentation on multimodal medical imaging.
\newblock \emph{IEEE Transactions on Radiation and Plasma Medical Sciences}, 3\penalty0 (2):\penalty0 162--169, 2019.

\bibitem[Gutman et~al.(2016)Gutman, Codella, Celebi, Helba, Marchetti, Mishra, and Halpern]{gutman2016skin}
David Gutman, Noel~CF Codella, Emre Celebi, Brian Helba, Michael Marchetti, Nabin Mishra, and Allan Halpern.
\newblock Skin lesion analysis toward melanoma detection: A challenge at the international symposium on biomedical imaging (isbi) 2016, hosted by the international skin imaging collaboration (isic).
\newblock \emph{arXiv preprint arXiv:1605.01397}, 2016.

\bibitem[Han et~al.(2022)Han, Wang, Guo, Tang, and Wu]{han2022vision}
Kai Han, Yunhe Wang, Jianyuan Guo, Yehui Tang, and Enhua Wu.
\newblock Vision gnn: An image is worth graph of nodes.
\newblock \emph{Advances in neural information processing systems}, 35:\penalty0 8291--8303, 2022.

\bibitem[Haralick et~al.(1987)Haralick, Sternberg, and Zhuang]{haralick1987image}
Robert~M Haralick, Stanley~R Sternberg, and Xinhua Zhuang.
\newblock Image analysis using mathematical morphology.
\newblock \emph{IEEE transactions on pattern analysis and machine intelligence}, \penalty0 (4):\penalty0 532--550, 1987.

\bibitem[He et~al.(2016)He, Zhang, Ren, and Sun]{he2016deep}
Kaiming He, Xiangyu Zhang, Shaoqing Ren, and Jian Sun.
\newblock Deep residual learning for image recognition.
\newblock In \emph{Proceedings of the IEEE conference on computer vision and pattern recognition}, pages 770--778, 2016.

\bibitem[He et~al.(2024)He, Yang, Su, and Wang]{he2024multi}
Qiqi He, Qiuju Yang, Hang Su, and Yixuan Wang.
\newblock Multi-task learning for segmentation and classification of breast tumors from ultrasound images.
\newblock \emph{Computers in Biology and Medicine}, 173:\penalty0 108319, 2024.

\bibitem[He et~al.(2023)He, Wang, Zhao, and Chen]{he2023co}
Xiaoyu He, Yong Wang, Shuang Zhao, and Xiang Chen.
\newblock Co-attention fusion network for multimodal skin cancer diagnosis.
\newblock \emph{Pattern Recognition}, 133:\penalty0 108990, 2023.

\bibitem[Hu et~al.(2023)Hu, Chen, Sun, Hu, Zhou, and Zheng]{hu2023ppnet}
Keli Hu, Wenping Chen, YuanZe Sun, Xiaozhao Hu, Qianwei Zhou, and Zirui Zheng.
\newblock Ppnet: Pyramid pooling based network for polyp segmentation.
\newblock \emph{Computers in Biology and Medicine}, 160:\penalty0 107028, 2023.

\bibitem[Huang et~al.(2018)Huang, Yang, Lang, and Deng]{huang2018decorrelated}
Lei Huang, Dawei Yang, Bo Lang, and Jia Deng.
\newblock Decorrelated batch normalization.
\newblock In \emph{Proceedings of the IEEE Conference on Computer Vision and Pattern Recognition}, pages 791--800, 2018.

\bibitem[Huang et~al.(2023)Huang, Chen, Li, Li, Li, Song, Yan, and Xiong]{huang2023style}
Wei Huang, Chang Chen, Yong Li, Jiacheng Li, Cheng Li, Fenglong Song, Youliang Yan, and Zhiwei Xiong.
\newblock Style projected clustering for domain generalized semantic segmentation.
\newblock In \emph{Proceedings of the IEEE/CVF conference on computer vision and pattern recognition}, pages 3061--3071, 2023.

\bibitem[Jha et~al.(2020)Jha, Smedsrud, Riegler, Halvorsen, de~Lange, Johansen, and Johansen]{jha2020kvasir}
Debesh Jha, Pia~H Smedsrud, Michael~A Riegler, P{\aa}l Halvorsen, Thomas de Lange, Dag Johansen, and H{\aa}vard~D Johansen.
\newblock Kvasir-seg: A segmented polyp dataset.
\newblock In \emph{MultiMedia Modeling: 26th International Conference, MMM 2020, Daejeon, South Korea, January 5--8, 2020, Proceedings, Part II 26}, pages 451--462. Springer, 2020.

\bibitem[Jiang et~al.(2023)Jiang, Chen, Tian, and Liu]{jiang2023vig}
Juntao Jiang, Xiyu Chen, Guanzhong Tian, and Yong Liu.
\newblock Vig-unet: vision graph neural networks for medical image segmentation.
\newblock In \emph{2023 IEEE 20th International Symposium on Biomedical Imaging (ISBI)}, pages 1--5. IEEE, 2023.

\bibitem[Jun et~al.(2020)Jun, Cheng, Yixin, Xingle, Jiantao, Ziqi, Minqing, Xin, Xueyuan, Shucheng, Hao, Sen, Xiaoyu, Ziwei, Chen, Lu, Yuntao, Qiongjie, Guoqiang, and Jian]{ma_jun_2020_3757476}
Ma Jun, Ge Cheng, Wang Yixin, An Xingle, Gao Jiantao, Yu Ziqi, Zhang Minqing, Liu Xin, Deng Xueyuan, Cao Shucheng, Wei Hao, Mei Sen, Yang Xiaoyu, Nie Ziwei, Li Chen, Tian Lu, Zhu Yuntao, Zhu Qiongjie, Dong Guoqiang, and He Jian.
\newblock {COVID-19 CT Lung and Infection Segmentation Dataset}.
\newblock 2020.

\bibitem[Jung et~al.(2023)Jung, Nizam, and Lee]{jung2023local}
Yerim Jung, Nur Suriza Syazwany Binti~Ahmad Nizam, and Sang-Chul Lee.
\newblock Local feature extraction from salient regions by feature map transformation.
\newblock \emph{arXiv preprint arXiv:2301.10413}, 2023.

\bibitem[Kass et~al.(1988)Kass, Witkin, and Terzopoulos]{kass1988snakes}
Michael Kass, Andrew Witkin, and Demetri Terzopoulos.
\newblock Snakes: Active contour models.
\newblock \emph{International journal of computer vision}, 1\penalty0 (4):\penalty0 321--331, 1988.

\bibitem[Kingma and Ba(2015)]{KingmaB14}
Diederik~P. Kingma and Jimmy Ba.
\newblock Adam: {A} method for stochastic optimization.
\newblock In \emph{3rd International Conference on Learning Representations, {ICLR} 2015, San Diego, CA, USA, May 7-9, 2015, Conference Track Proceedings}, 2015.

\bibitem[Krizhevsky et~al.(2012)Krizhevsky, Sutskever, and Hinton]{krizhevsky2012imagenet}
Alex Krizhevsky, Ilya Sutskever, and Geoffrey~E Hinton.
\newblock Imagenet classification with deep convolutional neural networks.
\newblock \emph{Advances in neural information processing systems}, 25, 2012.

\bibitem[Lee et~al.(2022)Lee, Seong, Lee, and Kim]{lee2022wildnet}
Suhyeon Lee, Hongje Seong, Seongwon Lee, and Euntai Kim.
\newblock Wildnet: Learning domain generalized semantic segmentation from the wild.
\newblock In \emph{Proceedings of the IEEE/CVF conference on computer vision and pattern recognition}, pages 9936--9946, 2022.

\bibitem[Li et~al.(2021)Li, Li, Li, Gong, Fu, and Hospedales]{li2021simple}
Pan Li, Da Li, Wei Li, Shaogang Gong, Yanwei Fu, and Timothy~M Hospedales.
\newblock A simple feature augmentation for domain generalization.
\newblock In \emph{Proceedings of the IEEE/CVF International Conference on Computer Vision}, pages 8886--8895, 2021.

\bibitem[Liu et~al.(2024{\natexlab{a}})Liu, Yao, Liu, Chang, Chen, Wang, and Wei]{liu2024cafe}
Guoqi Liu, Sheng Yao, Dong Liu, Baofang Chang, Zongyu Chen, Jiajia Wang, and Jiangqi Wei.
\newblock Cafe-net: Cross-attention and feature exploration network for polyp segmentation.
\newblock \emph{Expert Systems with Applications}, 238:\penalty0 121754, 2024{\natexlab{a}}.

\bibitem[Liu et~al.(2024{\natexlab{b}})Liu, Zhao, Cen, Wu, Wu, Zhang, and Zhou]{liu2024automated}
Jinping Liu, Junqi Zhao, Lihui Cen, Xiaoqiang Wu, Juanjuan Wu, Kun Zhang, and Ying Zhou.
\newblock Automated segmentation of covid-19 lung infections from chest ct images via 2.5 d dual-path encoder with multiscale dynamic weight deep supervision mechanism.
\newblock \emph{IEEE Transactions on Instrumentation and Measurement}, 2024{\natexlab{b}}.

\bibitem[Lloyd(1982)]{lloyd1982least}
Stuart Lloyd.
\newblock Least squares quantization in pcm.
\newblock \emph{IEEE transactions on information theory}, 28\penalty0 (2):\penalty0 129--137, 1982.

\bibitem[Loshchilov and Hutter(2016)]{loshchilov2016sgdr}
Ilya Loshchilov and Frank Hutter.
\newblock Sgdr: Stochastic gradient descent with warm restarts.
\newblock \emph{arXiv preprint arXiv:1608.03983}, 2016.

\bibitem[Luo et~al.(2018)Luo, Ren, Peng, Zhang, and Li]{luo2018differentiable}
Ping Luo, Jiamin Ren, Zhanglin Peng, Ruimao Zhang, and Jingyu Li.
\newblock Differentiable learning-to-normalize via switchable normalization.
\newblock In \emph{International Conference on Learning Representations}, 2018.

\bibitem[Ma et~al.(2024)Ma, Kim, Li, Baharoon, Asakereh, Lyu, and Wang]{ma2024segment}
Jun Ma, Sumin Kim, Feifei Li, Mohammed Baharoon, Reza Asakereh, Hongwei Lyu, and Bo Wang.
\newblock Segment anything in medical images and videos: Benchmark and deployment.
\newblock \emph{arXiv preprint arXiv:2408.03322}, 2024.

\bibitem[Ma et~al.(2018)Ma, Zhang, Zheng, and Sun]{ma2018shufflenet}
Ningning Ma, Xiangyu Zhang, Hai-Tao Zheng, and Jian Sun.
\newblock Shufflenet v2: Practical guidelines for efficient cnn architecture design.
\newblock In \emph{Proceedings of the European conference on computer vision (ECCV)}, pages 116--131, 2018.

\bibitem[Margolin et~al.(2014)Margolin, Zelnik-Manor, and Tal]{margolin2014evaluate}
Ran Margolin, Lihi Zelnik-Manor, and Ayellet Tal.
\newblock How to evaluate foreground maps?
\newblock In \emph{Proceedings of the IEEE conference on computer vision and pattern recognition}, pages 248--255, 2014.

\bibitem[Mendon{\c{c}}a et~al.(2013)Mendon{\c{c}}a, Ferreira, Marques, Marcal, and Rozeira]{mendoncca2013ph}
Teresa Mendon{\c{c}}a, Pedro~M Ferreira, Jorge~S Marques, Andr{\'e}~RS Marcal, and Jorge Rozeira.
\newblock Ph 2-a dermoscopic image database for research and benchmarking.
\newblock In \emph{2013 35th annual international conference of the IEEE engineering in medicine and biology society (EMBC)}, pages 5437--5440. IEEE, 2013.

\bibitem[Milletari et~al.(2016)Milletari, Navab, and Ahmadi]{milletari2016v}
Fausto Milletari, Nassir Navab, and Seyed-Ahmad Ahmadi.
\newblock V-net: Fully convolutional neural networks for volumetric medical image segmentation.
\newblock In \emph{2016 fourth international conference on 3D vision (3DV)}, pages 565--571. Ieee, 2016.

\bibitem[Nam et~al.(2024)Nam, Syazwany, Kim, and Lee]{nam2024modality}
Ju-Hyeon Nam, Nur~Suriza Syazwany, Su~Jung Kim, and Sang-Chul Lee.
\newblock Modality-agnostic domain generalizable medical image segmentation by multi-frequency in multi-scale attention.
\newblock In \emph{Proceedings of the IEEE/CVF Conference on Computer Vision and Pattern Recognition}, pages 11480--11491, 2024.

\bibitem[Nam et~al.(2025)Nam, Syazwany, and Lee]{nam2025transgunet}
Ju-Hyeon Nam, Nur~Suriza Syazwany, and Sang-Chul Lee.
\newblock Transgunet: Transformer meets graph-based skip connection for medical image segmentation.
\newblock \emph{arXiv preprint arXiv:2502.09931}, 2025.

\bibitem[Otsu(1979)]{otsu1979threshold}
Nobuyuki Otsu.
\newblock A threshold selection method from gray-level histograms.
\newblock \emph{IEEE transactions on systems, man, and cybernetics}, 9\penalty0 (1):\penalty0 62--66, 1979.

\bibitem[Pan et~al.(2018)Pan, Luo, Shi, and Tang]{pan2018two}
Xingang Pan, Ping Luo, Jianping Shi, and Xiaoou Tang.
\newblock Two at once: Enhancing learning and generalization capacities via ibn-net.
\newblock In \emph{Proceedings of the european conference on computer vision (ECCV)}, pages 464--479, 2018.

\bibitem[Pan et~al.(2019)Pan, Zhan, Shi, Tang, and Luo]{pan2019switchable}
Xingang Pan, Xiaohang Zhan, Jianping Shi, Xiaoou Tang, and Ping Luo.
\newblock Switchable whitening for deep representation learning.
\newblock In \emph{Proceedings of the IEEE/CVF international conference on computer vision}, pages 1863--1871, 2019.

\bibitem[Park et~al.(2019)Park, Garcia-Palacios, Cohen, and Varga]{park2019treatment}
Sophie Park, Javier Garcia-Palacios, Andrew Cohen, and Zsuzsanna Varga.
\newblock From treatment to prevention: The evolution of digital healthcare.
\newblock \emph{Nature}, 573\penalty0 (7775), 2019.

\bibitem[Paszke et~al.(2019)Paszke, Gross, Massa, Lerer, Bradbury, Chanan, Killeen, Lin, Gimelshein, Antiga, Desmaison, Kopf, Yang, DeVito, Raison, Tejani, Chilamkurthy, Steiner, Fang, Bai, and Chintala]{NEURIPS2019_9015}
Adam Paszke, Sam Gross, Francisco Massa, Adam Lerer, James Bradbury, Gregory Chanan, Trevor Killeen, Zeming Lin, Natalia Gimelshein, Luca Antiga, Alban Desmaison, Andreas Kopf, Edward Yang, Zachary DeVito, Martin Raison, Alykhan Tejani, Sasank Chilamkurthy, Benoit Steiner, Lu Fang, Junjie Bai, and Soumith Chintala.
\newblock Pytorch: An imperative style, high-performance deep learning library.
\newblock In \emph{Advances in Neural Information Processing Systems 32}, pages 8024--8035. Curran Associates, Inc., 2019.

\bibitem[Peng et~al.(2022)Peng, Lei, Hayat, Guo, and Li]{peng2022semantic}
Duo Peng, Yinjie Lei, Munawar Hayat, Yulan Guo, and Wen Li.
\newblock Semantic-aware domain generalized segmentation.
\newblock In \emph{Proceedings of the IEEE/CVF conference on computer vision and pattern recognition}, pages 2594--2605, 2022.

\bibitem[Qi et~al.(2023)Qi, Wu, and Chan]{qi2023mdf}
Wenbo Qi, Ho-Chun Wu, and Shing-Chow Chan.
\newblock Mdf-net: A multi-scale dynamic fusion network for breast tumor segmentation of ultrasound images.
\newblock \emph{IEEE Transactions on Image Processing}, 2023.

\bibitem[Rahman and Marculescu(2023)]{Rahman_2023_WACV}
Md~Mostafijur Rahman and Radu Marculescu.
\newblock Medical image segmentation via cascaded attention decoding.
\newblock In \emph{Proceedings of the IEEE/CVF Winter Conference on Applications of Computer Vision (WACV)}, pages 6222--6231, 2023.

\bibitem[Rahman and Marculescu(2024)]{rahman2024g}
Md~Mostafijur Rahman and Radu Marculescu.
\newblock G-cascade: Efficient cascaded graph convolutional decoding for 2d medical image segmentation.
\newblock In \emph{Proceedings of the IEEE/CVF Winter Conference on Applications of Computer Vision}, pages 7728--7737, 2024.

\bibitem[Riccio et~al.(2018)Riccio, Brancati, Frucci, and Gragnaniello]{riccio2018new}
Daniel Riccio, Nadia Brancati, Maria Frucci, and Diego Gragnaniello.
\newblock A new unsupervised approach for segmenting and counting cells in high-throughput microscopy image sets.
\newblock \emph{IEEE journal of biomedical and health informatics}, 23\penalty0 (1):\penalty0 437--448, 2018.

\bibitem[Ronneberger et~al.(2015)Ronneberger, Fischer, and Brox]{ronneberger2015u}
Olaf Ronneberger, Philipp Fischer, and Thomas Brox.
\newblock U-net: Convolutional networks for biomedical image segmentation.
\newblock In \emph{Medical Image Computing and Computer-Assisted Intervention--MICCAI 2015: 18th International Conference, Munich, Germany, October 5-9, 2015, Proceedings, Part III 18}, pages 234--241. Springer, 2015.

\bibitem[Sandler et~al.(2018)Sandler, Howard, Zhu, Zhmoginov, and Chen]{sandler2018mobilenetv2}
Mark Sandler, Andrew Howard, Menglong Zhu, Andrey Zhmoginov, and Liang-Chieh Chen.
\newblock Mobilenetv2: Inverted residuals and linear bottlenecks.
\newblock In \emph{Proceedings of the IEEE conference on computer vision and pattern recognition}, pages 4510--4520, 2018.

\bibitem[Sarker et~al.(2018)Sarker, Rashwan, Akram, Banu, Saleh, Singh, Chowdhury, Abdulwahab, Romani, Radeva, et~al.]{sarker2018slsdeep}
Md~Mostafa~Kamal Sarker, Hatem~A Rashwan, Farhan Akram, Syeda~Furruka Banu, Adel Saleh, Vivek~Kumar Singh, Forhad~UH Chowdhury, Saddam Abdulwahab, Santiago Romani, Petia Radeva, et~al.
\newblock Slsdeep: Skin lesion segmentation based on dilated residual and pyramid pooling networks.
\newblock In \emph{Medical Image Computing and Computer Assisted Intervention--MICCAI 2018: 21st International Conference, Granada, Spain, September 16-20, 2018, Proceedings, Part II 11}, pages 21--29. Springer, 2018.

\bibitem[Silva et~al.(2014)Silva, Histace, Romain, Dray, and Granado]{silva2014toward}
Juan Silva, Aymeric Histace, Olivier Romain, Xavier Dray, and Bertrand Granado.
\newblock Toward embedded detection of polyps in wce images for early diagnosis of colorectal cancer.
\newblock \emph{International journal of computer assisted radiology and surgery}, 9:\penalty0 283--293, 2014.

\bibitem[Tajbakhsh et~al.(2015)Tajbakhsh, Gurudu, and Liang]{tajbakhsh2015automated}
Nima Tajbakhsh, Suryakanth~R Gurudu, and Jianming Liang.
\newblock Automated polyp detection in colonoscopy videos using shape and context information.
\newblock \emph{IEEE transactions on medical imaging}, 35\penalty0 (2):\penalty0 630--644, 2015.

\bibitem[Tizhoosh(2005)]{tizhoosh2005image}
Hamid~R Tizhoosh.
\newblock Image thresholding using type ii fuzzy sets.
\newblock \emph{Pattern recognition}, 38\penalty0 (12):\penalty0 2363--2372, 2005.

\bibitem[Ulyanov et~al.(2017)Ulyanov, Vedaldi, and Lempitsky]{ulyanov2017improved}
Dmitry Ulyanov, Andrea Vedaldi, and Victor Lempitsky.
\newblock Improved texture networks: Maximizing quality and diversity in feed-forward stylization and texture synthesis.
\newblock In \emph{Proceedings of the IEEE conference on computer vision and pattern recognition}, pages 6924--6932, 2017.

\bibitem[Van~der Maaten and Hinton(2008)]{van2008visualizing}
Laurens Van~der Maaten and Geoffrey Hinton.
\newblock Visualizing data using t-sne.
\newblock \emph{Journal of machine learning research}, 9\penalty0 (11), 2008.

\bibitem[V{\'a}zquez et~al.(2017)V{\'a}zquez, Bernal, S{\'a}nchez, Fern{\'a}ndez-Esparrach, L{\'o}pez, Romero, Drozdzal, Courville, et~al.]{vazquez2017benchmark}
David V{\'a}zquez, Jorge Bernal, F~Javier S{\'a}nchez, Gloria Fern{\'a}ndez-Esparrach, Antonio~M L{\'o}pez, Adriana Romero, Michal Drozdzal, Aaron Courville, et~al.
\newblock A benchmark for endoluminal scene segmentation of colonoscopy images.
\newblock \emph{Journal of healthcare engineering}, 2017, 2017.

\bibitem[Wang et~al.(2024)Wang, Wang, Wang, Wei, Feng, Wu, Yao, and Zhang]{wang2024cfatransunet}
Cheng Wang, Le Wang, Nuoqi Wang, Xiaoling Wei, Ting Feng, Minfeng Wu, Qi Yao, and Rongjun Zhang.
\newblock Cfatransunet: Channel-wise cross fusion attention and transformer for 2d medical image segmentation.
\newblock \emph{Computers in Biology and Medicine}, 168:\penalty0 107803, 2024.

\bibitem[Wang et~al.(2022{\natexlab{a}})Wang, Cao, Wang, and Zaiane]{wang2022uctransnet}
Haonan Wang, Peng Cao, Jiaqi Wang, and Osmar~R Zaiane.
\newblock Uctransnet: rethinking the skip connections in u-net from a channel-wise perspective with transformer.
\newblock In \emph{Proceedings of the AAAI conference on artificial intelligence}, pages 2441--2449, 2022{\natexlab{a}}.

\bibitem[Wang et~al.(2022{\natexlab{b}})Wang, Xie, Li, Fan, Song, Liang, Lu, Luo, and Shao]{wang2022pvt}
Wenhai Wang, Enze Xie, Xiang Li, Deng-Ping Fan, Kaitao Song, Ding Liang, Tong Lu, Ping Luo, and Ling Shao.
\newblock Pvt v2: Improved baselines with pyramid vision transformer.
\newblock \emph{Computational Visual Media}, 8\penalty0 (3):\penalty0 415--424, 2022{\natexlab{b}}.

\bibitem[Zhang et~al.(2022)Zhang, Lin, Chen, Tian, Yang, Tang, and Cheng]{zhang2022understanding}
Dong Zhang, Yi Lin, Hao Chen, Zhuotao Tian, Xin Yang, Jinhui Tang, and Kwang~Ting Cheng.
\newblock Understanding the tricks of deep learning in medical image segmentation: Challenges and future directions.
\newblock \emph{arXiv preprint arXiv:2209.10307}, 2022.

\bibitem[Zhang et~al.(2024)Zhang, Wang, Jha, Demir, and Bagci]{zhang2024domain}
Zheyuan Zhang, Bin Wang, Debesh Jha, Ugur Demir, and Ulas Bagci.
\newblock Domain generalization with correlated style uncertainty.
\newblock In \emph{Proceedings of the IEEE/CVF Winter Conference on Applications of Computer Vision}, pages 2000--2009, 2024.

\bibitem[Zhao et~al.(2021)Zhao, Zhang, and Lu]{zhao2021automatic}
Xiaoqi Zhao, Lihe Zhang, and Huchuan Lu.
\newblock Automatic polyp segmentation via multi-scale subtraction network.
\newblock In \emph{Medical Image Computing and Computer Assisted Intervention--MICCAI 2021: 24th International Conference, Strasbourg, France, September 27--October 1, 2021, Proceedings, Part I 24}, pages 120--130. Springer, 2021.

\bibitem[Zhao et~al.(2023)Zhao, Jia, Pang, Lv, Tian, Zhang, Sun, and Lu]{zhao2023m}
Xiaoqi Zhao, Hongpeng Jia, Youwei Pang, Long Lv, Feng Tian, Lihe Zhang, Weibing Sun, and Huchuan Lu.
\newblock $\text{M}^{2}\text{SNet}$: Multi-scale in multi-scale subtraction network for medical image segmentation.
\newblock \emph{arXiv preprint arXiv:2303.10894}, 2023.

\bibitem[Zhou et~al.(2023)Zhou, Zhang, Yao, Lu, Yi, Ding, and Ma]{zhou2023instance}
Qianyu Zhou, Ke-Yue Zhang, Taiping Yao, Xuequan Lu, Ran Yi, Shouhong Ding, and Lizhuang Ma.
\newblock Instance-aware domain generalization for face anti-spoofing.
\newblock In \emph{Proceedings of the IEEE/CVF Conference on Computer Vision and Pattern Recognition}, pages 20453--20463, 2023.

\bibitem[Zhou et~al.(2024)Zhou, Zhang, Yao, Lu, Ding, and Ma]{zhou2024test}
Qianyu Zhou, Ke-Yue Zhang, Taiping Yao, Xuequan Lu, Shouhong Ding, and Lizhuang Ma.
\newblock Test-time domain generalization for face anti-spoofing.
\newblock In \emph{Proceedings of the IEEE/CVF Conference on Computer Vision and Pattern Recognition}, pages 175--187, 2024.

\bibitem[Zhou et~al.(2018)Zhou, Rahman~Siddiquee, Tajbakhsh, and Liang]{zhou2018unet++}
Zongwei Zhou, Md~Mahfuzur Rahman~Siddiquee, Nima Tajbakhsh, and Jianming Liang.
\newblock Unet++: A nested u-net architecture for medical image segmentation.
\newblock In \emph{Deep Learning in Medical Image Analysis and Multimodal Learning for Clinical Decision Support: 4th International Workshop, DLMIA 2018, and 8th International Workshop, ML-CDS 2018, Held in Conjunction with MICCAI 2018, Granada, Spain, September 20, 2018, Proceedings 4}, pages 3--11. Springer, 2018.

\bibitem[Zhou et~al.(2022)Zhou, Qi, Yang, Ni, and Shi]{zhou2022generalizable}
Ziqi Zhou, Lei Qi, Xin Yang, Dong Ni, and Yinghuan Shi.
\newblock Generalizable cross-modality medical image segmentation via style augmentation and dual normalization.
\newblock In \emph{Proceedings of the IEEE/CVF conference on computer vision and pattern recognition}, pages 20856--20865, 2022.

\bibitem[Zhuang et~al.(2019)Zhuang, Li, Joseph~Raj, Mahesh, and Qiu]{zhuang2019rdau}
Zhemin Zhuang, Nan Li, Alex~Noel Joseph~Raj, Vijayalakshmi~GV Mahesh, and Shunmin Qiu.
\newblock An rdau-net model for lesion segmentation in breast ultrasound images.
\newblock \emph{PloS one}, 14\penalty0 (8):\penalty0 e0221535, 2019.

\end{thebibliography}
}
\clearpage
\setcounter{page}{1}
\maketitlesupplementary

\section{Dataset Descriptions}
\label{appendix_dataset_descriptions}

\begin{table}[h]
    \centering
    \scriptsize
    \setlength\tabcolsep{2pt} 
    \begin{tabular}{c|cccccc}
    \hline
    Dataset & Modality & Images & Resolutions & Train & Valid & Test \\
     \hline
     ISIC2018 \cite{gutman2016skin}                   & Dermoscopy   & 2594 & Variable           & 1868 & 465 & 261 \\
     COVID19-1 \cite{ma_jun_2020_3757476}             & Radiology    & 1277 & 512 $\times$ 512   & 643  & 251 & 383 \\
     BUSI \cite{al2020dataset}                        & Ultrasound   & 645  & Variable           & 324  & 160 & 161 \\
     CVC-ClinicDB \cite{bernal2015wm}                 & Colonoscopy  & 612  & 384 $\times$ 288   & 490  & 60  & 62  \\
     Kvasir-SEG \cite{jha2020kvasir}                  & Colonoscopy  & 1000 & Variable           & 800  & 100 & 100 \\
    \hline
    \end{tabular}
    \caption{Details of the medical segmentation \textit{seen} clinical settings used in our experiments.}
    \label{tab:seen_clilical_dataset}
\end{table}

\begin{table}[h]
    \centering
    \scriptsize
    \begin{tabular}{c|cccc}
    \hline
    Dataset & Modality & Images & Resolutions & Test \\
     \hline
     PH2 \cite{mendoncca2013ph}                   & Dermoscopy   & 200  & 767 $\times$ 576  & 200  \\
     COVID19-2 \cite{COVID19_2}                   & Radiology    & 2535 & 512 $\times$ 512  & 2535 \\
     STU \cite{zhuang2019rdau}                    & Ultrasound   & 42   & Variable          & 42	  \\
     CVC-300 \cite{vazquez2017benchmark}          & Colonoscopy  & 60   & 574 $\times$ 500  & 60   \\
     CVC-ColonDB \cite{tajbakhsh2015automated}    & Colonoscopy  & 380  & 574 $\times$ 500  & 380  \\
     ETIS \cite{silva2014toward}                  & Colonoscopy  & 196  & 1255 $\times$ 966 & 196  \\
    \hline
    \end{tabular}
    \caption{Details of the medical segmentation \textit{unseen} clinical settings used in our experiments.}
    \label{tab:unseen_clilical_dataset}
\end{table}

\begin{itemize}
    \item \textit{Breast Ultrasound Segmentation:} The BUSI \cite{al2020dataset} comprises 780 images from 600 female patients, including 133 normal cases, 437 benign cases, and 210 malignant tumors. On the other hand, the STU \cite{zhuang2019rdau} includes only 42 breast ultrasound images collected by Shantou University. Due to the limited number of images in the STU, it is used only to evaluate the generalizability of each model across different datasets. \\

    \item \textit{Skin Lesion Segmentation:} The ISIC 2018 \cite{gutman2016skin} comprises 2,594 images with various sizes. We randomly selected train, validation, and test images with 1,868, 465, and 261, respectively. And, we used PH2 \cite{mendoncca2013ph} to evaluate the domain generalizability of each model. Note that ISIC2018 \cite{gutman2016skin} and PH2 \cite{mendoncca2013ph} are \textit{seen}, and \textit{unseen} clinical settings, respectively. \\
    
    \item \textit{COVID19 Lung Infection Segmentation:} COVID19-1 \cite{ma_jun_2020_3757476} comprises 1,277 high-quality CT images. We randomly selected train, validation, and test images with 643, 251, and 383, respectively. And, we used COVID19-2 \cite{COVID19_2} to evaluate the domain generalizability of each model. Note that COVID19-2 \cite{COVID19_2} is used for only testing. \\
     
    \item \textit{Polyp Segmentation:} Colorectal cancer is the third most prevalent cancer globally and the second most common cause of death. It typically originates as small, non-cancerous (benign) clusters of cells known as polyps, which develop inside the colon. To evaluate the proposed model, we have used five benchmark datasets, namely CVC-ColonDB \cite{tajbakhsh2015automated}, ETIS \cite{silva2014toward}, Kvasir \cite{jha2020kvasir}, CVC-300 \cite{vazquez2017benchmark}, and CVC-ClinicDB \cite{bernal2015wm}. The same training set as the latest image polyp segmentation method has been adopted, consisting of 900 samples from Kvasir and 550 samples from CVC-ClinicDB for training. The remaining images and the other three datasets are used for only testing.  \\
\end{itemize}

\begin{table}[t]
    \centering
    \scriptsize
    \setlength\tabcolsep{3.5pt} 
\begin{tabular}{c|cccc}
    \hline
    \multicolumn{1}{c|}{\multirow{2}{*}{Method}} & Parameters & FLOPs & Training Time & Inference Time \\
     & (MB) & (GB) & (sec/epoch) & (ms/image) \\
     \hline
     baseline \cite{ronneberger2015u}   & 38.48 & 64.68 & 11 & 7.98  \\
     \hline
     IN \cite{ulyanov2017improved}      & 38.48 \tiny{(0.00\%)} & 64.68 \tiny{(0.00\%)}
                                        & 11 \tiny{(0.00\%)}    & 8.35  \tiny{(4.64\%)} \\
     IW \cite{huang2018decorrelated}    & 38.48 \tiny{(0.00\%)} & 64.68 \tiny{(0.00\%)}
                                        & 12 \tiny{(9.09\%)}    & 8.00  \tiny{(0.25\%)} \\
     IBN \cite{pan2018two}              & 38.48 \tiny{(0.00\%)} & 64.70 \tiny{(0.03\%)}
                                        & 12 \tiny{(9.09\%)}    & 8.06  \tiny{(1.00\%)} \\
     RobustNet \cite{choi2021robustnet} & 38.48 \tiny{(0.00\%)} & 64.68 \tiny{(0.00\%)}
                                        & 21 \tiny{(90.91\%)}   & 11.23 \tiny{(0.38\%)} \\ 
     SAN-SAW \cite{peng2022semantic}    & 38.48 \tiny{(0.00\%)} & 74.80 \tiny{(15.65\%)}
                                        & 92 \tiny{(736.36\%)}  & 14.23 \tiny{(78.32\%)} \\
     SPCNet \cite{huang2023style}       & 38.48 \tiny{(0.00\%)} & 64.68 \tiny{(0.00\%)}
                                        & 13 \tiny{(18.18\%)}   & 8.41  \tiny{(5.39\%)} \\
     BlindNet \cite{ahn2024style}       & 38.48 \tiny{(0.00\%)} & 64.68 \tiny{(0.00\%)}
                                        & 30 \tiny{(172.73\%)}  & 8.14  \tiny{(2.01\%)} \\
     \hline
     \textbf{TTMG (Ours)}               & 38.48 \tiny{(0.00\%)} & 64.69 \tiny{(0.02\%)}
                                        & 15 \tiny{(36.36\%)}   & 8.70 \tiny{(9.02\%)} \\
    \hline
    \end{tabular}
    \caption{The number of parameters (MB), FLOPs (GB), Training Time (sec/epoch) and Inference Speed (ms/image) for each different method.}
    \label{tab:efficiency_analysis}
\end{table}

\begin{table*}[t]
    \centering
    \footnotesize
    \setlength\tabcolsep{2.0pt} 
    \begin{tabular}{c|cc|cc|c||cc|cc|c||cc|cc|c}
    \hline
    \multicolumn{1}{c|}{\multirow{4}{*}{Method}} & \multicolumn{15}{c}{Training Modalities (R, U)} \\ \cline{2-16}
     & \multicolumn{5}{c||}{DeepLabV3+ with MobileNetV2 \cite{sandler2018mobilenetv2}}
     & \multicolumn{5}{c||}{DeepLabV3+ with ShuffleNetV2 \cite{ma2018shufflenet}}
     & \multicolumn{5}{c}{TransUNet \cite{chen2021transunet}} \\ \cline{2-16}
     & \multicolumn{2}{c|}{Seen (R, U)} & \multicolumn{2}{c|}{Unseen (C, D)} & \multicolumn{1}{c||}{\multirow{2}{*}{$P$-value}}
     & \multicolumn{2}{c|}{Seen (R, U)} & \multicolumn{2}{c|}{Unseen (C, D)} & \multicolumn{1}{c||}{\multirow{2}{*}{$P$-value}}
     & \multicolumn{2}{c|}{Seen (R, U)} & \multicolumn{2}{c|}{Unseen (C, D)} & \multicolumn{1}{c}{\multirow{2}{*}{$P$-value}} \\ \cline{2-5}\cline{7-10}\cline{12-15}
     & DSC & mIoU & DSC & mIoU &  & DSC & mIoU & DSC & mIoU &  & DSC & mIoU & DSC & mIoU & \\
     \hline
     baseline                                                                & 69.99 & 60.76 & 0.15  & 0.09 
                                                                             & 1.5E-02
                                                                             & 64.32 & 55.11 & 0.76  & 0.45 
                                                                             & 1.1E-02
                                                                             & 58.59 & 48.37 & 12.22 & 9.14 
                                                                             & 2.3E-04 \\
\hline
    \end{tabular}
    \caption{Quantitative results on seen (R, U) and unseen (C, D) modalities in DeepLabV3+ with MobileNetV2 \cite{sandler2018mobilenetv2}, DeepLabV3+ with ShuffleNetV2 \cite{ma2018shufflenet}, and TransUNet \cite{chen2021transunet}. We also provide one tailed Student \textit{t}-Test results (\textit{P}-value) compared to our method (TTMG) and other methods.}
    \label{tab:backbone_type}
\end{table*}

\section{Technical Novelty of TTMG}
\label{appendix_technical_novelty_of_ttmg}

\noindent \textbf{RobustNet (CVPR2021) vs TTMG.} Compared to RobustNet, TTMG’s technical novelty lies in its ability to adapt dynamically at test time with Modality-Aware Style Projection (MASP), which projects unseen modality instances onto seen modality style distributions. Additionally, Modality-Sensitive Instance Whitening (MSIW) selectively whitens only modality-sensitive features, preserving invariant information essential for cross-modality segmentation. This dual approach enables TTMG to achieve robust generalization across highly diverse medical imaging modalities, beyond what RobustNet’s fixed training-time ISW can offer.

\noindent \textbf{SAN-SAW (CVPR2022) vs TTMG.} TTMG’s technical novelty over SAN-SAW lies in its test-time adaptability via Modality-Aware Style Projection (MASP), which dynamically projects unseen modality instances to seen modality styles, unlike SAN-SAW’s fixed semantic-aware normalization. Additionally, Modality-Sensitive Instance Whitening (MSIW) selectively removes only modality-sensitive noise, ensuring robust cross-modality generalization with lower computational cost than SAN-SAW’s intensive regional normalization approach.

\noindent \textbf{SPCNet (CVPR2023) vs TTMG.} TTMG’s technical novelty over SPCNet lies in its test-time adaptability through Modality-Aware Style Projection (MASP), which dynamically aligns unseen modality features to the closest seen modality style, unlike SPCNet’s fixed prototypes. Additionally, Modality-Sensitive Instance Whitening (MSIW) selectively removes only modality-sensitive noise, enhancing cross-modality generalization without sacrificing essential content features—a capability SPCNet’s prototype-based approach lacks.

\noindent \textbf{BlindNet (CVPR2024) vs TTMG.} TTMG’s technical novelty over BlindNet lies in its modality-specific adaptability at test time through Modality-Aware Style Projection (MASP), which aligns unseen modality instances to seen modality styles rather than applying random noise. Additionally, Modality-Sensitive Instance Whitening (MSIW) selectively filters out modality-sensitive features, preserving invariant information crucial for accurate segmentation across diverse modalities—a precision BlindNet’s generalized noise-based approach does not offer.

\section{Broader Impact in Artificial Intelligence}
\label{appendix_broader_impact_in_artificial_intelligence}

The development of \textbf{Test-Time Modality Generalization (TTMG)} has broader implications for the field of Artificial Intelligence, particularly in enhancing the adaptability and robustness of AI models across diverse domains. By focusing on unseen modality generalization, TTMG addresses a core challenge in AI: building models that can perform reliably in environments or contexts not encountered during training. In healthcare, TTMG’s ability to generalize across different imaging modalities without additional training has the potential to reduce the need for extensive data collection and model retraining, accelerating deployment in clinical settings. This capability could lead to more accessible and scalable AI solutions, making it feasible to use a single model across different hospitals, regions, and patient demographics. Furthermore, TTMG’s dual approach of \textit{Modality-Aware Style Projection (MASP)} and \textit{Modality-Sensitive Instance Whitening (MSIW)} could inspire new techniques for addressing data variability in other critical applications where data consistency is challenging.

\section{More Detailed Ablation Study on TTMG}
\label{appendix_more_detailed_ablation_study_on_TTMG}

\subsection{Ablation Study on Backbone Types}

In this section, we evaluate the performance of DeepLabV3+ with different backbone networks (MobileNetV2 \cite{sandler2018mobilenetv2} and ShuffleNetV2 \cite{ma2018shufflenet}) and TransUNet \cite{chen2021transunet} when integrating \textbf{TTMG (Ours)} and other methods. Notably, only the backbone network was altered, while all other architectural settings were kept consistent with those used in the main experiment as mentioned in Section \ref{s42_implementation_details}. The \textit{mean} performance for each \textit{seen} and \textit{unseen} modalities is reported in Table \ref{tab:backbone_type}. The datasets for the \textit{seen} and \textit{unseen} modalities are identical to those used in Tables \ref{tab:comparison_three_modality} and \ref{tab:comparison_two_modality}, respectively.

As shown in Table \ref{tab:backbone_type}, TTMG consistently achieves high performance not only with DeepLabV3+ but also with TransUNet, which is the most representative Transformer-based approach. We also present one-tailed Student’s \textit{t}-test results (\textit{P}-value) comparing our method (TTMG) with other approaches, confirming that the performance improvement achieved by TTMG is statistically significant. These results demonstrate the versatility of TTMG, confirming its applicability as a plug-and-play module across a broader range of models.

\subsection{Ablation Study on Applied Stage for TTMG} 

\begin{table}[h]
    \centering
    \footnotesize
    \setlength\tabcolsep{3.0pt} 
    \begin{tabular}{c|cc|cc|c}
    \hline 
    \multicolumn{1}{c|}{\multirow{3}{*}{TTMG Applied Stage $L$}} & \multicolumn{4}{c|}{Training Modalities (R, U)} & \multicolumn{1}{c}{\multirow{3}{*}{$P$-value}} \\ \cline{2-5}
     & \multicolumn{2}{c|}{Seen (R, U)} & \multicolumn{2}{c|}{Unseen (C, D)} &  \\ \cline{2-5}
     & DSC & mIoU & DSC & mIoU &  \\
     \hline
     $L = \{ \}$ (baseline) & 70.77 & 61.59 & 3.92 & 2.33 & 7.8E-02 \\
     \hline
     \multicolumn{1}{c|}{\multirow{2}{*}{$L = \{ 1 \}$}}  & 76.24 & 64.74 & 14.17 & 10.41 & \multicolumn{1}{c}{\multirow{2}{*}{3.2E-02}} \\
                                                          & \textcolor{ForestGreen}{\scriptsize{\textbf{(+5.47)}}}
                                                          & \textcolor{ForestGreen}{\scriptsize{\textbf{(+3.15)}}}
                                                          & \textcolor{ForestGreen}{\scriptsize{\textbf{(+9.73)}}}
                                                          & \textcolor{ForestGreen}{\scriptsize{\textbf{(+7.68)}}} 
                                                          & \\ \hline
     \multicolumn{1}{c|}{\multirow{2}{*}{$L = \{ 2 \}$}}  & 75.41 & 66.66 & 16.85 & 13.18 & \multicolumn{1}{c}{\multirow{2}{*}{5.3E-02}} \\
                                                          & \textcolor{ForestGreen}{\scriptsize{\textbf{(+4.65)}}}
                                                          & \textcolor{ForestGreen}{\scriptsize{\textbf{(+5.07)}}}
                                                          & \textcolor{ForestGreen}{\scriptsize{\textbf{(+12.41)}}}
                                                          & \textcolor{ForestGreen}{\scriptsize{\textbf{(+10.44)}}} 
                                                          & \\ \hline
     \multicolumn{1}{c|}{\multirow{2}{*}{$L = \{ 3 \}$}}  & 73.86 & 65.40 & 12.89 & 10.36 & \multicolumn{1}{c}{\multirow{2}{*}{2.1E-02}} \\
                                                          & \textcolor{ForestGreen}{\scriptsize{\textbf{(+3.09)}}}
                                                          & \textcolor{ForestGreen}{\scriptsize{\textbf{(+3.81)}}}
                                                          & \textcolor{ForestGreen}{\scriptsize{\textbf{(+8.45)}}}
                                                          & \textcolor{ForestGreen}{\scriptsize{\textbf{(+7.63)}}}
                                                          & \\ \hline
     \multicolumn{1}{c|}{\multirow{2}{*}{$L = \{ 4 \}$}}  & 73.98 & 65.53 & 10.18 & 6.61 & \multicolumn{1}{c}{\multirow{2}{*}{2.0E-02}} \\
                                                          & \textcolor{ForestGreen}{\scriptsize{\textbf{(+3.21)}}}
                                                          & \textcolor{ForestGreen}{\scriptsize{\textbf{(+3.94)}}}
                                                          & \textcolor{ForestGreen}{\scriptsize{\textbf{(+5.74)}}}
                                                          & \textcolor{ForestGreen}{\scriptsize{\textbf{(+3.87)}}}
                                                          & \\ \hline
     \multicolumn{1}{c|}{\multirow{2}{*}{\textbf{$L = \{ 1, 2 \}$ (Ours)}}} & 73.62 & 65.22 & 20.96 & 16.98 & \multicolumn{1}{c}{\multirow{2}{*}{-}} \\
                                                                            & \textcolor{ForestGreen}{\scriptsize{\textbf{(+2.85)}}} 
                                                                            & \textcolor{ForestGreen}{\scriptsize{\textbf{(+3.63)}}} 
                                                                            & \textcolor{ForestGreen}{\scriptsize{\textbf{(+17.04)}}} 
                                                                            & \textcolor{ForestGreen}{\scriptsize{\textbf{(+14.65)}}} 
                                                                            & \\ \hline
     \multicolumn{1}{c|}{\multirow{2}{*}{$L = \{ 3, 4 \}$}}                 & 72.14 & 63.61 & 5.59 & 3.90 & \multicolumn{1}{c}{\multirow{2}{*}{1.2E-02}} \\
                                                                            & \textcolor{ForestGreen}{\scriptsize{\textbf{(+1.37)}}}
                                                                            & \textcolor{ForestGreen}{\scriptsize{\textbf{(+2.02)}}}
                                                                            & \textcolor{ForestGreen}{\scriptsize{\textbf{(+1.16)}}}
                                                                            & \textcolor{ForestGreen}{\scriptsize{\textbf{(+1.16)}}}
                                                                            & \\ \hline
     \multicolumn{1}{c|}{\multirow{2}{*}{$L = \{ 1, 2, 3, 4 \}$}}           & 70.15 & 61.30 & 5.16 & 3.59 & \multicolumn{1}{c}{\multirow{2}{*}{6.4E-03}} \\
                                                                            & \textcolor{red}{\scriptsize{\textbf{(-0.62)}}}
                                                                            & \textcolor{red}{\scriptsize{\textbf{(-0.30)}}}	
                                                                            & \textcolor{ForestGreen}{\scriptsize{\textbf{(+0.73)}}}
                                                                            & \textcolor{ForestGreen}{\scriptsize{\textbf{(+0.85)}}}
                                                                            & \\ \hline
    \end{tabular}
    \caption{Quantitative results for each \textit{Seen} and \textit{Unseen} modalities according to various applied stage $L$ for TTMG. We also provide one tailed Student \textit{t}-Test results (\textit{P}-value) compared to our method (TTMG) and other configurations.}
    \label{tab:applied_stage}
\end{table}

In this section, we conducted an ablation study to evaluate the performance impact of applying the TTMG framework (MASP and MSIW) at different stages for $L = \{ 1 \}, \{ 2 \}, \{ 3 \}, \{ 4 \}, \{ 1, 2 \}, \{ 3, 4 \}, \{ 1, 2, 3, 4 \}$. In the main manuscript, we used $L = \{ 1, 2 \}$ as the TTMG applied stage. We want to clarify that the same settings that were mentioned in Section \ref{s42_implementation_details} were applied in the ablation study. The \textit{mean} performance for each \textit{seen} and \textit{unseen} modalities is reported in Table \ref{tab:applied_stage}. The datasets for the \textit{seen} and \textit{unseen} modalities are identical to those used in Tables \ref{tab:comparison_three_modality} and \ref{tab:comparison_two_modality}, respectively.

As listed in Table \ref{tab:applied_stage}, the experimental results indicate that the applied stage $L = \{ 1, 2 \}$ achieved the best unseen modality generalization performance. Additionally, we observed that applying TTMG at deeper stages $(L \ge 3)$ leads to relatively smaller performance improvements compared to shallow layers $(L \le 2)$. This results finding suggests that the input image style is primarily captured in the shallow layers, consistent with observations from previous studies.

\subsection{Ablation Study on Weighting Strategy for TTMG}

\begin{table}[h]
    \centering
    \footnotesize
    \setlength\tabcolsep{3.0pt} 
    \begin{tabular}{c|cc|cc|c}
    \hline 
    \multicolumn{1}{c|}{\multirow{3}{*}{Weighting Strategy}} & \multicolumn{4}{c|}{Training Modalities (R, U)} & \multicolumn{1}{c}{\multirow{3}{*}{$P$-value}} \\ \cline{2-5}
     & \multicolumn{2}{c|}{Seen (R, U)} & \multicolumn{2}{c|}{Unseen (C, D)} &  \\ \cline{2-5}
     & DSC & mIoU & DSC & mIoU &  \\
     \hline
     baseline & 70.77 & 61.59 & 3.92 & 2.33 & 7.8E-02 \\
     \hline
     \multicolumn{1}{c|}{\multirow{2}{*}{Hard Weighting}}  & 73.21 & 64.46 & 18.93 & 15.17 & \multicolumn{1}{c}{\multirow{2}{*}{5.1E-02}} \\
                                                          & \textcolor{ForestGreen}{\scriptsize{\textbf{(+2.44)}}}
                                                          & \textcolor{ForestGreen}{\scriptsize{\textbf{(+2.87)}}}
                                                          & \textcolor{ForestGreen}{\scriptsize{\textbf{(+14.50)}}}
                                                          & \textcolor{ForestGreen}{\scriptsize{\textbf{(+12.43)}}} 
                                                          & \\ \hline
     \multicolumn{1}{c|}{\multirow{2}{*}{\textbf{Soft Weighting (Ours)}}}   & 73.62 & 65.22 & 20.96 & 16.98 & \multicolumn{1}{c}{\multirow{2}{*}{-}} \\
                                                                            & \textcolor{ForestGreen}{\scriptsize{\textbf{(+2.85)}}} 
                                                                            & \textcolor{ForestGreen}{\scriptsize{\textbf{(+3.63)}}} 
                                                                            & \textcolor{ForestGreen}{\scriptsize{\textbf{(+17.04)}}} 
                                                                            & \textcolor{ForestGreen}{\scriptsize{\textbf{(+14.65)}}} 
                                                                            & \\ \hline
    \end{tabular}
    \caption{Quantitative results for each \textit{Seen} and \textit{Unseen} modalities according to weighting strategy (Hard vs Soft weighting) for TTMG. We also provide one tailed Student \textit{t}-Test results (\textit{P}-value) compared to our method (TTMG) and other configurations.}
    \label{tab:weighting_strategy}
\end{table}

In this section, we conducted an ablation study to evaluate the performance impact of weight strategy (Hard vs Soft weighting) in the MASP stage. In the main manuscript, we used the Soft weighting strategy to project the feature map. The hard weighting strategy is defined as follows:
\begin{equation}
\label{eq:hard_projection}
    F_{\text{MASP}} = \text{argmax}_{m} (p_{m}) \cdot F_{m} \in \mathbb{R}^{H_{l} \times W_{l} \times C_{l}}
\end{equation}
As defined in Eq. \ref{eq:hard_projection}, the Hard Strategy projects features into the modality style distribution with the highest probability, disregarding probabilistic information across the seen modalities. We want to clarify that the same settings that were mentioned in Section \ref{s42_implementation_details} were applied in the ablation study. The \textit{mean} performance for each \textit{seen} and \textit{unseen} modalities is reported in Table \ref{tab:weighting_strategy}. The datasets for the \textit{seen} and \textit{unseen} modalities are identical to those used in Tables \ref{tab:comparison_three_modality} and \ref{tab:comparison_two_modality}, respectively.

As listed in Table \ref{tab:weighting_strategy}, the experimental results underscore the significance of leveraging probabilistic information (Soft Weighting Strategy).  While both strategies consistently outperform the baseline, the generalization performance on unseen modalities is notably lower with hard weighting compared to soft weighting.

\subsection{Ablation Study on Feature Map Combination for MSIW}

\begin{table}[h]
    \centering
    \footnotesize
    \setlength\tabcolsep{0.25pt} 
    \begin{tabular}{c|cc|cc|c}
    \hline 
    \multicolumn{1}{c|}{\multirow{3}{*}{Feature Map Combination $\mathbf{F}_{m}$}} & \multicolumn{4}{c|}{Training Modalities (R, U)} & \multicolumn{1}{c}{\multirow{3}{*}{$P$-value}} \\ \cline{2-5}
     & \multicolumn{2}{c|}{Seen (R, U)} & \multicolumn{2}{c|}{Unseen (C, D)} &  \\ \cline{2-5}
     & DSC & mIoU & DSC & mIoU &  \\
     \hline
     baseline & 70.77 & 61.59 & 3.92 & 2.33 & 7.8E-02 \\
     \hline
     \multicolumn{1}{c|}{\multirow{2}{*}{$\mathbf{F}_{m} = \{ F_{1}, F_{2}, \dots, F_{M} \}$}}  & 73.88 & 65.24 & 18.10 & 13.91 & \multicolumn{1}{c}{\multirow{2}{*}{2.2E-02}} \\
                                                          & \textcolor{ForestGreen}{\scriptsize{\textbf{(+3.11)}}}
                                                          & \textcolor{ForestGreen}{\scriptsize{\textbf{(+3.65)}}}
                                                          & \textcolor{ForestGreen}{\scriptsize{\textbf{(+13.66)}}}
                                                          & \textcolor{ForestGreen}{\scriptsize{\textbf{(+11.18)}}}
                                                          & \\ \hline
    \multicolumn{1}{c|}{\multirow{2}{*}{$\mathbf{F}_{m} = \{ F_{0}, F_{M + 1} \}$}}  & 73.17 & 64.66 & 17.46 & 13.70 & \multicolumn{1}{c}{\multirow{2}{*}{1.0E-03}} \\
                                                          & \textcolor{ForestGreen}{\scriptsize{\textbf{(+2.40)}}}
                                                          & \textcolor{ForestGreen}{\scriptsize{\textbf{(+3.07)}}}
                                                          & \textcolor{ForestGreen}{\scriptsize{\textbf{(+13.03)}}}
                                                          & \textcolor{ForestGreen}{\scriptsize{\textbf{(+10.96)}}}
                                                          & \\ \hline
     $\mathbf{F}_{m} = \{ F_{0}, F_{1}, \dots, F_{M}, F_{M + 1} \}$   & 73.62 & 65.22 & 20.96 & 16.98 & \multicolumn{1}{c}{\multirow{2}{*}{-}} \\
     \textbf{(Ours)}                                                  & \textcolor{ForestGreen}{\scriptsize{\textbf{(+2.85)}}} 
                                                                            & \textcolor{ForestGreen}{\scriptsize{\textbf{(+3.63)}}} 
                                                                            & \textcolor{ForestGreen}{\scriptsize{\textbf{(+17.04)}}} 
                                                                            & \textcolor{ForestGreen}{\scriptsize{\textbf{(+14.65)}}} 
                                                                            & \\ \hline
    \end{tabular}
    \caption{Quantitative results for each \textit{Seen} and \textit{Unseen} modalities according to feature map combination in MSIW. We also provide one tailed Student \textit{t}-Test results (\textit{P}-value) compared to our method (TTMG) and other configurations.}
    \label{tab:feature_map_combination}
\end{table}

In this section, we conducted an ablation study to evaluate the performance impact of feature map combination at MSIW stages for $\mathbf{F}_{m} = \{ F_{1}, F_{2}, \dots, F_{M} \}, \{ F_{0}, F_{M + 1} \}, \{ F_{0}, F_{1}, \dots, F_{M}, F_{M + 1} \}$. In the main manuscript, we used $\mathbf{F}_{m} = \{ F_{0}, F_{1}, \dots, F_{M}, F_{M + 1} \}$ as feature map combination at MSIW stages to calculate MSIW loss $\mathcal{L}_{\text{MSIW}}$. We want to clarify that the same settings that were mentioned in Section \ref{s42_implementation_details} were applied in the ablation study. The \textit{mean} performance for each \textit{seen} and \textit{unseen} modalities is reported in Table \ref{tab:feature_map_combination}. The datasets for the \textit{seen} and \textit{unseen} modalities are identical to those used in Tables \ref{tab:comparison_three_modality} and \ref{tab:comparison_two_modality}, respectively.

As shown in Table \ref{tab:feature_map_combination}, the experimental results indicate that incorporating the MSIW loss $\mathcal{L}_{\text{MSIW}}$ with all feature maps $\mathbf{F}{m} = \{ F_{0}, F_{1}, \dots, F_{M}, F_{M+1} \}$ achieves the highest generalization performance on unseen modalities. This improvement is attributed to the inclusion of $F_{m}$ for $m = 1, 2, \dots, M$, which captures more modality-specific styles within $F_{\text{MASP}}$, enabling more precise selection of modality-sensitive information.

\begin{table*}[t]
    \centering
    \tiny
    \setlength\tabcolsep{1.00pt} 
    \renewcommand{\arraystretch}{1.35} 
\begin{tabular}{c|ccccc||ccccc||ccccc||ccccc||ccccc}
    \hline
    \multicolumn{1}{c|}{\multirow{2}{*}{Method}} & \multicolumn{5}{c||}{Average on Seen Modality (C, U, D)}  & \multicolumn{5}{c||}{Average on Seen Modality (C, U, R)} & \multicolumn{5}{c||}{Average on Seen Modality (C, D, R)} & \multicolumn{5}{c||}{Average on Seen Modality (U, D, R)} & \multicolumn{5}{c}{Mean} \\ \cline{2-26}
     & DSC & mIoU & $F_{\beta}^{w}$ & $S_{\alpha}$ & $E_{\phi}^{m}$ & DSC & mIoU & $F_{\beta}^{w}$ & $S_{\alpha}$ & $E_{\phi}^{m}$ & DSC & mIoU & $F_{\beta}^{w}$ & $S_{\alpha}$ & $E_{\phi}^{m}$ & DSC & mIoU & $F_{\beta}^{w}$ & $S_{\alpha}$ & $E_{\phi}^{m}$ & DSC & mIoU & $F_{\beta}^{w}$ & $S_{\alpha}$ & $E_{\phi}^{m}$ \\
     \hline
     Baseline \cite{chen2018encoder}                                     & 80.98 & 73.65 & 79.26 & 81.89 & 90.14 & 76.20 & 68.67 & 73.93 & 82.14 & 86.40 & 77.95 & 70.62 & 75.78 & 80.83 & 86.55 & 76.32 & 67.99 & 72.99 & 78.12 & 86.45 & 77.86 & 70.23 & 75.49 & 80.75 & 87.39 \\
     \hline
     \multicolumn{1}{c|}{\multirow{2}{*}{IN \cite{ulyanov2017improved}}} & 82.24 & 75.00 & 80.72 & 82.61 & 90.85 & 74.23 & 66.48 & 71.75 & 81.04 & 85.48 & 78.17 & 70.84 & 75.80 & 81.10 & 87.58 & 74.49 & 65.32 & 71.53 & 76.95 & 86.74 & 77.28 & 69.41 & 74.95 & 80.42 & 87.66 \\
\hline
    \end{tabular}
    \caption{Average segmentation results with training scheme of three modality (C, U, D), (C, U, R), (C, D, R), and (U, D, R). Full segmentation results of (C, U, D), (C, U, R), (C, D, R), and (U, D, R) is listed in Tables \ref{tab:comparison_CUD_other_metrics}, \ref{tab:comparison_CUR_other_metrics}, \ref{tab:comparison_CDR_other_metrics}, and \ref{tab:comparison_UDR_other_metrics}, respectively.}
    \label{tab:comparison_other_three_metrics_average}
\end{table*}

\begin{table*}[t]
    \centering
    \fontsize{3.95}{4}\selectfont
    \setlength\tabcolsep{0.1pt} 
    \renewcommand{\arraystretch}{1.35} 
\begin{tabular}{c|ccccc||ccccc||ccccc||ccccc||ccccc||ccccc||ccccc}
    \hline
    \multicolumn{1}{c|}{\multirow{2}{*}{Method}} & \multicolumn{5}{c||}{Average on Seen Modality (C, D)} & \multicolumn{5}{c||}{Average on Seen Modality (C, R)} & \multicolumn{5}{c||}{Average on Seen Modality (C, U)} & \multicolumn{5}{c||}{Average on Seen Modality (D, U)} & \multicolumn{5}{c||}{Average on Seen Modality (D, R)} & \multicolumn{5}{c||}{Average on Seen Modality (R, U)} & \multicolumn{5}{c}{Mean} \\ \cline{2-36}
     & DSC & mIoU & $F_{\beta}^{w}$ & $S_{\alpha}$ & $E_{\phi}^{m}$ 
     & DSC & mIoU & $F_{\beta}^{w}$ & $S_{\alpha}$ & $E_{\phi}^{m}$ 
     & DSC & mIoU & $F_{\beta}^{w}$ & $S_{\alpha}$ & $E_{\phi}^{m}$
     & DSC & mIoU & $F_{\beta}^{w}$ & $S_{\alpha}$ & $E_{\phi}^{m}$ 
     & DSC & mIoU & $F_{\beta}^{w}$ & $S_{\alpha}$ & $E_{\phi}^{m}$ 
     & DSC & mIoU & $F_{\beta}^{w}$ & $S_{\alpha}$ & $E_{\phi}^{m}$
     & DSC & mIoU & $F_{\beta}^{w}$ & $S_{\alpha}$ & $E_{\phi}^{m}$ \\
     \hline
     Baseline \cite{chen2018encoder}    & 81.62 & 75.69 & 80.38 & 82.07 & 90.03 & 73.51 & 66.57 & 70.83 & 81.31 & 84.80 & 80.62 & 73.45 & 79.01 & 84.19 & 90.16 & 84.60 & 76.56 & 82.66 & 79.80 & 91.52 & 75.42 & 67.39 & 71.98 & 76.35 & 84.12 & 70.77 & 61.59 & 66.71 & 78.57 & 84.30 & 77.76 & 70.21 & 75.26 & 80.38 & 87.49 \\
     \hline
     \multicolumn{1}{c|}{\multirow{2}{*}{IN \cite{ulyanov2017improved}}} & 81.19 & 74.55 & 80.01 & 81.94 & 89.46 & 74.82 & 67.44 & 71.98 & 81.92 & 85.69 & 80.04 & 72.57 & 78.54 & 83.84 & 90.35 & 85.70 & 77.85 & 83.49 & 80.58 & 92.28 & 57.06 & 49.18 & 52.12 & 70.61 & 71.56 & 70.49 & 60.96 & 66.77 & 78.17 & 84.22 & 74.88 & 67.09 & 72.15 & 79.51 & 85.59 \\
\hline
    \end{tabular}
    \caption{Average segmentation results with training scheme of two modality (C, D), (C, R), (C, U), (D, U), (D, R), and (R, U). Full segmentation results of (C, D), (C, R), (C, U), (D, U), (D, R), and (R, U) is listed in Tables \ref{tab:comparison_CD_other_metrics}, \ref{tab:comparison_CR_other_metrics}, \ref{tab:comparison_CU_other_metrics}, \ref{tab:comparison_DU_other_metrics}, \ref{tab:comparison_DR_other_metrics}, and \ref{tab:comparison_RU_other_metrics}.}
    \label{tab:comparison_other_two_metrics_average}
\end{table*}

\section{Metrics Descriptions}
\label{appendix_metric_descriptions}

In this section, we describe the metrics used in this paper. For convenience, we denote $TP, FP$, and $FN$ as the number of samples of true positive, false positive, and false negative between two binary masks $A$ and $B$.

\begin{itemize}
    \item The \textit{Mean Dice Similarity Coefficient (DSC)} \cite{milletari2016v} measures the similarity between two samples and is widely used to assess the performance of segmentation tasks, such as image segmentation or object detection. \textbf{\underline{Higher is better}}. For given two binary masks $A$ and $B$, DSC is defined as follows:
    \begin{equation}
        \textbf{DSC}(A, B) = \frac{2 \times | A \cap B |}{| A \cup B |} = \frac{2 \times TP}{2 \times TP + FP + FN}
    \end{equation}

    \item The \textit{Mean Intersection over Union (mIoU)} measures the ratio of the intersection area to the union area between predicted and ground truth masks in segmentation tasks. \textbf{\underline{Higher is better}}. For given two binary masks $A$ and $B$, mIoU is defined as follows:
    \begin{equation}
        \textbf{mIoU}(A, B) = \frac{A \cap B}{A \cup B} = \frac{TP}{TP + FP + FN}
    \end{equation}

    \item The \textit{Mean Weighted F-Measure} $F_{\beta}^{\omega}$ \cite{margolin2014evaluate} is a metric that combines weighted precision $Precision^{\omega}$ (Measure of exactness) and weighted recall $Recall^{\omega}$ (Measure of completeness) into a single value by calculating the harmonic mean. $\beta$ signifies the effectiveness of detection with respect to a user who attaches $\beta$ times as much importance to $Recall^{\omega}$ as to $Precision^{\omega}$. \textbf{\underline{Higher is better}}. $F_{\beta}^{\omega}$ is defined as follows:
    \begin{equation}
        F_{\beta}^{\omega} = (1 + \beta^{2}) \cdot \frac{Precision^{\omega} \cdot Recall^{\omega}}{\beta^{2} \cdot Precision^{\omega} + Recall^{\omega}}
    \end{equation}

    \item The \textit{Mean S-Measure} $S_{\alpha}$ \cite{fan2017structure} is used to evaluate the quality of image segmentation, specifically focusing on the structural similarity between the region-aware $S_{o}$ and object-aware $S_{r}$. \textbf{\underline{Higher is better}}. $S_{\alpha}$ is defined as follows:
    \begin{equation}
        S_{\alpha} = \gamma S_{o} + (1 - \gamma) S_{r}
    \end{equation}
    
    \item \textit{Mean E-Measure} $E_{\phi}^{max}$ \cite{fan2018enhanced} assesses the edge accuracy in edge detection or segmentation tasks. It evaluates how well the predicted edges align with the ground truth edges using foreground map $FM$. \textbf{\underline{Higher is better}}. $E_{\phi}^{max}$ is defined as follows: \\
    \begin{equation}
        E_{\phi}^{max} = \frac{1}{H \times W} \sum_{h = 1}^{H} \sum_{w = 1}^{W} \phi FM(h, w)
    \end{equation}
\end{itemize}

\section{More Qualitative and Quantitative Results}
\label{appendix_additional_experiment_results_with_various_metrics}

\begin{figure*}[t]
    \centering
    \includegraphics[width=\textwidth]{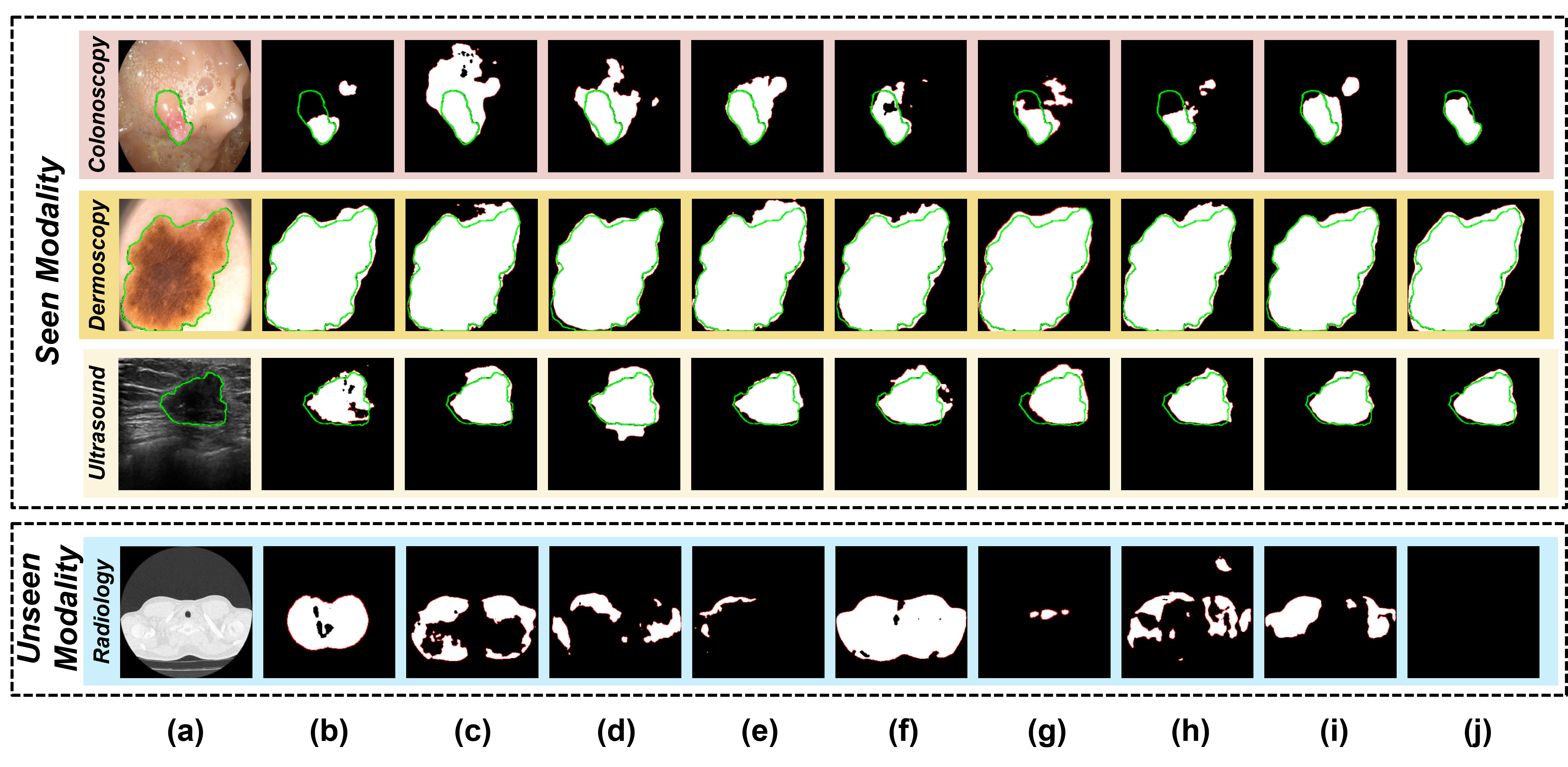}
    \caption{Qualitative comparison of other methods and TTMG on Colonoscopy, Dermoscopy, and Ultrasound modalities.  (a) Input images with ground truth. (b) baseline (DeepLabV3+ with ResNet50, \cite{chen2018encoder}). (c) IN \cite{ulyanov2017improved}. (d) IW \cite{huang2018decorrelated}. (e) IBN \cite{pan2018two}. (f) RobustNet \cite{choi2021robustnet}. (g) SAN-SAW \cite{peng2022semantic}. (h) SPCNet \cite{huang2023style}. (i) BlindNet \cite{ahn2024style}. (j) \textbf{TTMG (Ours)}. In this figure, \textcolor{green}{\textbf{Green}} and \textcolor{red}{\textbf{Red}} lines denote the boundaries of the ground truth and prediction, respectively.}
    \label{fig:Sup_CDU_R}
\end{figure*}

\begin{figure*}[t]
    \centering
    \includegraphics[width=\textwidth]{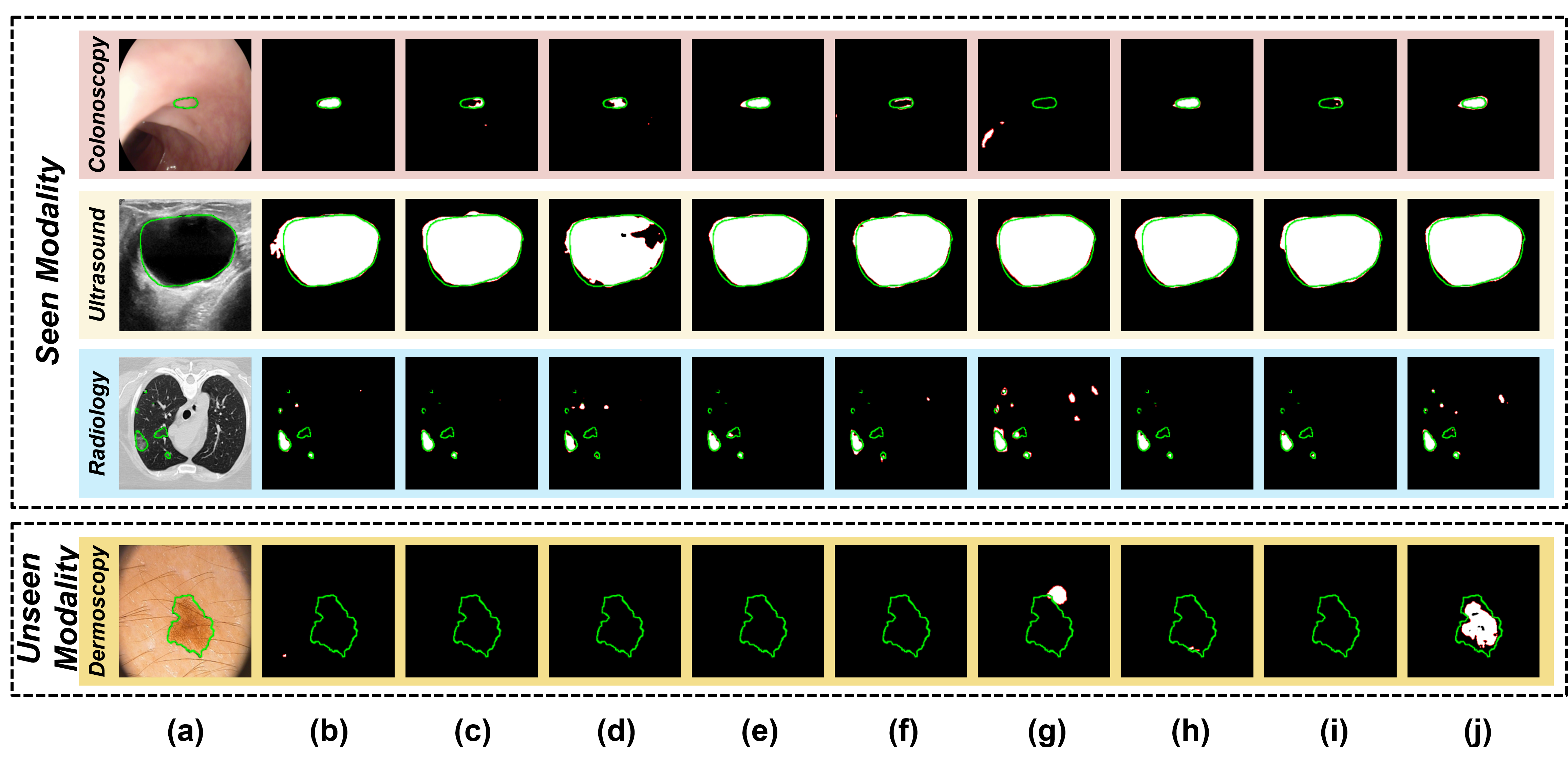}
    \caption{Qualitative comparison of other methods and TTMG on Colonoscopy, Ultrasound, and Radiology modalities.  (a) Input images with ground truth. (b) baseline (DeepLabV3+ with ResNet50, \cite{chen2018encoder}). (c) IN \cite{ulyanov2017improved}. (d) IW \cite{huang2018decorrelated}. (e) IBN \cite{pan2018two}. (f) RobustNet \cite{choi2021robustnet}. (g) SAN-SAW \cite{peng2022semantic}. (h) SPCNet \cite{huang2023style}. (i) BlindNet \cite{ahn2024style}. (j) \textbf{TTMG (Ours)}. In this figure, \textcolor{green}{\textbf{Green}} and \textcolor{red}{\textbf{Red}} lines denote the boundaries of the ground truth and prediction, respectively.}
    \label{fig:Sup_CUR_D}
\end{figure*}

\begin{figure*}[t]
    \centering
    \includegraphics[width=\textwidth]{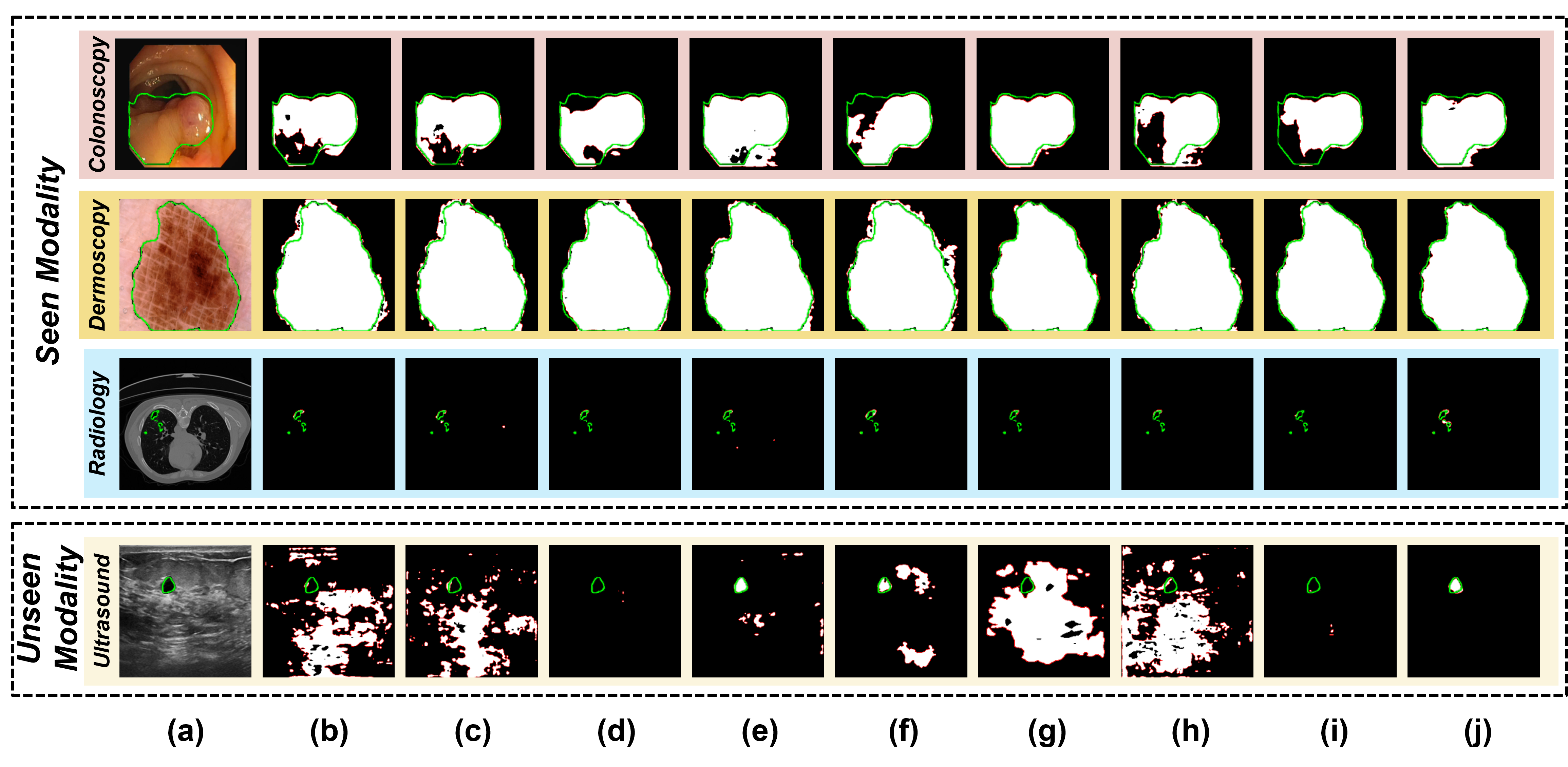}
    \caption{Qualitative comparison of other methods and TTMG on Colonoscopy, Dermoscopy, and Radiology modalities.  (a) Input images with ground truth. (b) baseline (DeepLabV3+ with ResNet50, \cite{chen2018encoder}). (c) IN \cite{ulyanov2017improved}. (d) IW \cite{huang2018decorrelated}. (e) IBN \cite{pan2018two}. (f) RobustNet \cite{choi2021robustnet}. (g) SAN-SAW \cite{peng2022semantic}. (h) SPCNet \cite{huang2023style}. (i) BlindNet \cite{ahn2024style}. (j) \textbf{TTMG (Ours)}. In this figure, \textcolor{green}{\textbf{Green}} and \textcolor{red}{\textbf{Red}} lines denote the boundaries of the ground truth and prediction, respectively.}
    \label{fig:Sup_CDR_U}
\end{figure*}

\begin{figure*}[t]
    \centering
    \includegraphics[width=\textwidth]{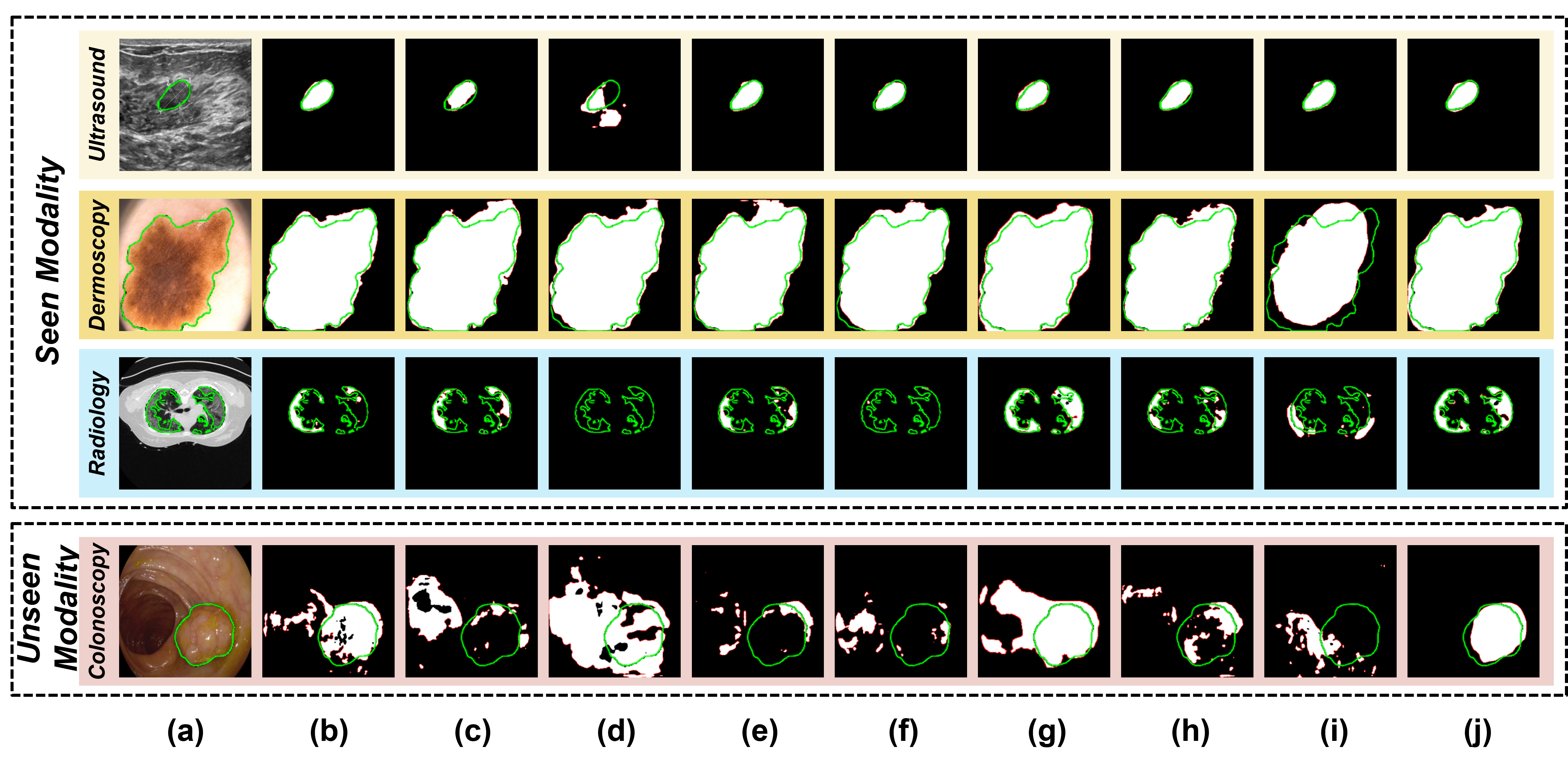}
    \caption{Qualitative comparison of other methods and TTMG on Ultrasound, Dermoscopy, and Radiology modalities.  (a) Input images with ground truth. (b) baseline (DeepLabV3+ with ResNet50, \cite{chen2018encoder}). (c) IN \cite{ulyanov2017improved}. (d) IW \cite{huang2018decorrelated}. (e) IBN \cite{pan2018two}. (f) RobustNet \cite{choi2021robustnet}. (g) SAN-SAW \cite{peng2022semantic}. (h) SPCNet \cite{huang2023style}. (i) BlindNet \cite{ahn2024style}. (j) \textbf{TTMG (Ours)}. In this figure, \textcolor{green}{\textbf{Green}} and \textcolor{red}{\textbf{Red}} lines denote the boundaries of the ground truth and prediction, respectively.}
    \label{fig:Sup_UDR_C}
\end{figure*}

\begin{figure*}[t]
    \centering
    \includegraphics[width=\textwidth]{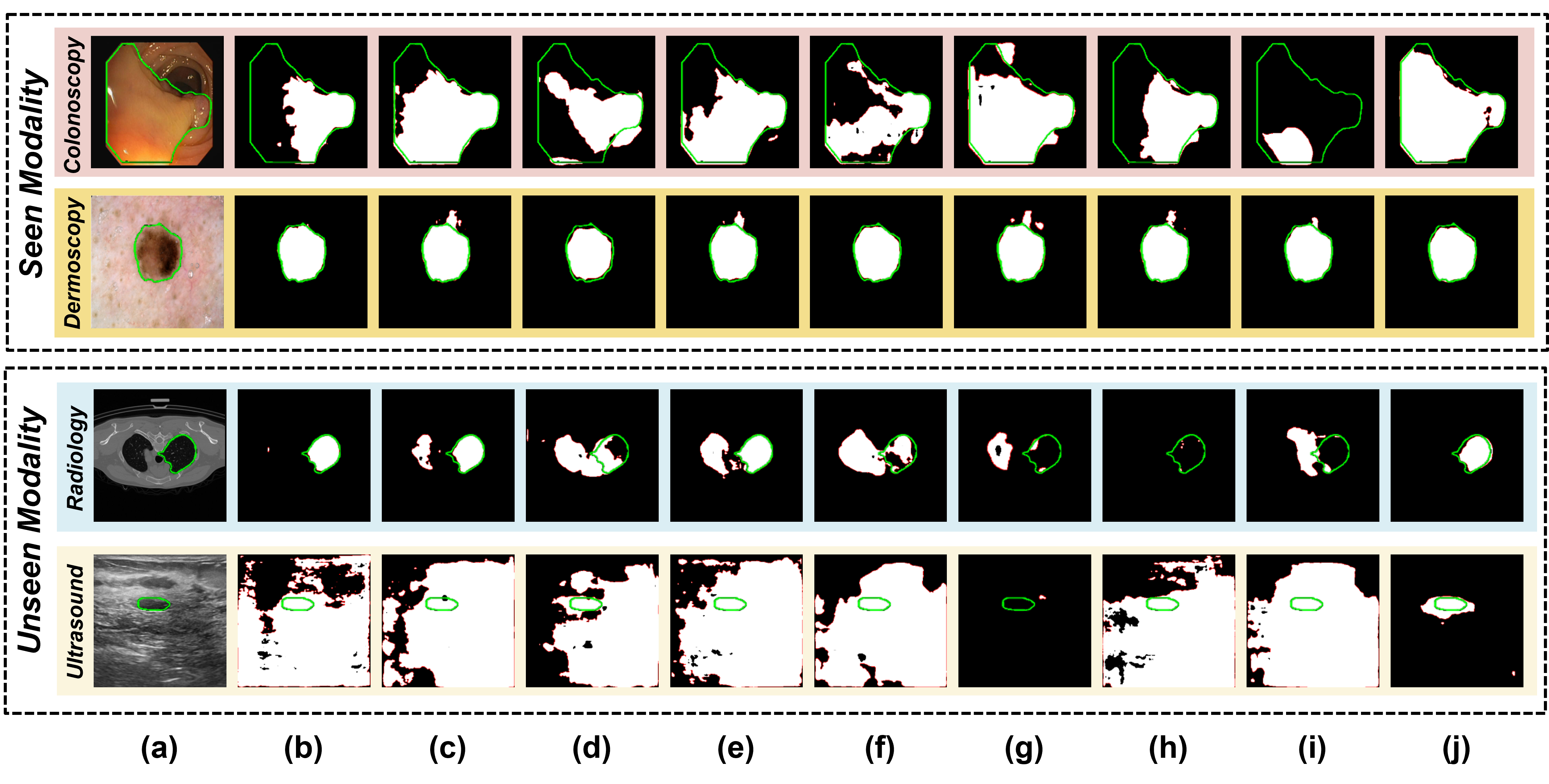}
    \caption{Qualitative comparison of other methods and TTMG on Colonoscopy and Dermoscopy modalities.  (a) Input images with ground truth. (b) baseline (DeepLabV3+ with ResNet50, \cite{chen2018encoder}). (c) IN \cite{ulyanov2017improved}. (d) IW \cite{huang2018decorrelated}. (e) IBN \cite{pan2018two}. (f) RobustNet \cite{choi2021robustnet}. (g) SAN-SAW \cite{peng2022semantic}. (h) SPCNet \cite{huang2023style}. (i) BlindNet \cite{ahn2024style}. (j) \textbf{TTMG (Ours)}. In this figure, \textcolor{green}{\textbf{Green}} and \textcolor{red}{\textbf{Red}} lines denote the boundaries of the ground truth and prediction, respectively.}
    \label{fig:Sup_CD_RU}
\end{figure*}

\begin{figure*}[t]
    \centering
    \includegraphics[width=\textwidth]{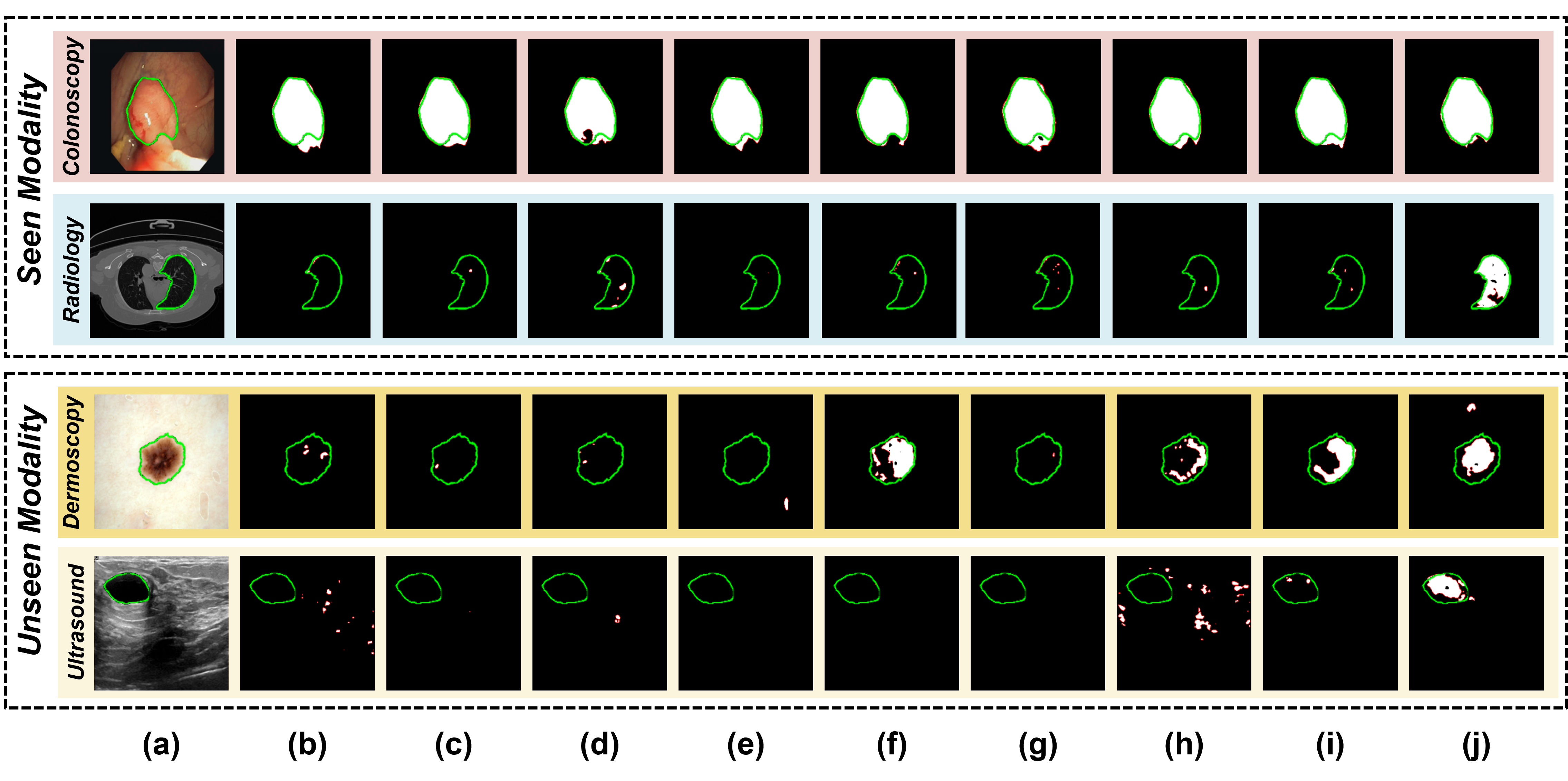}
    \caption{Qualitative comparison of other methods and TTMG on Colonoscopy and Radiology modalities.  (a) Input images with ground truth. (b) baseline (DeepLabV3+ with ResNet50, \cite{chen2018encoder}). (c) IN \cite{ulyanov2017improved}. (d) IW \cite{huang2018decorrelated}. (e) IBN \cite{pan2018two}. (f) RobustNet \cite{choi2021robustnet}. (g) SAN-SAW \cite{peng2022semantic}. (h) SPCNet \cite{huang2023style}. (i) BlindNet \cite{ahn2024style}. (j) \textbf{TTMG (Ours)}. In this figure, \textcolor{green}{\textbf{Green}} and \textcolor{red}{\textbf{Red}} lines denote the boundaries of the ground truth and prediction, respectively.}
    \label{fig:Sup_CR_DU}
\end{figure*}

\begin{figure*}[t]
    \centering
    \includegraphics[width=\textwidth]{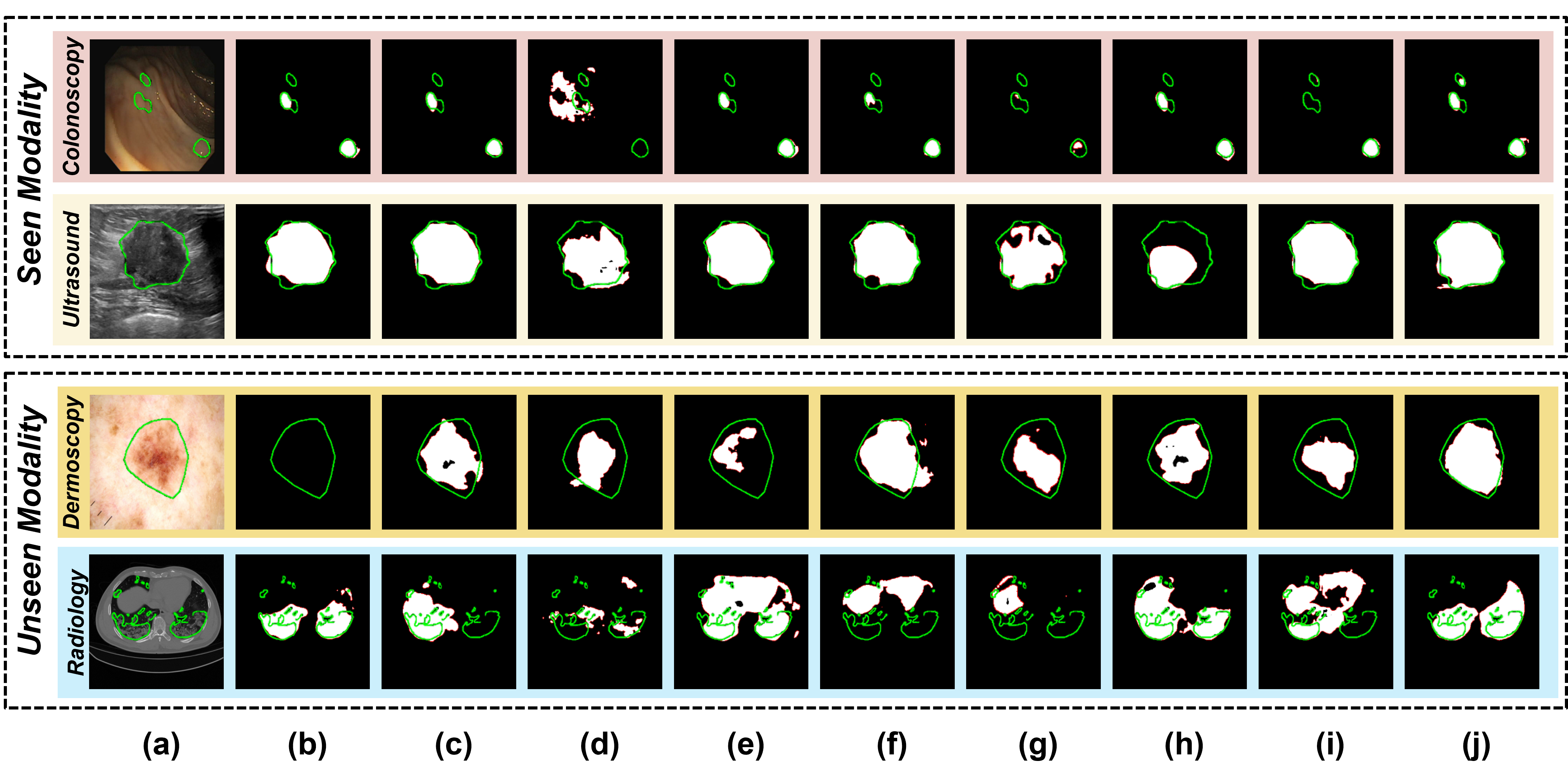}
    \caption{Qualitative comparison of other methods and TTMG on Colonoscopy and Ultrasound modalities.  (a) Input images with ground truth. (b) baseline (DeepLabV3+ with ResNet50, \cite{chen2018encoder}). (c) IN \cite{ulyanov2017improved}. (d) IW \cite{huang2018decorrelated}. (e) IBN \cite{pan2018two}. (f) RobustNet \cite{choi2021robustnet}. (g) SAN-SAW \cite{peng2022semantic}. (h) SPCNet \cite{huang2023style}. (i) BlindNet \cite{ahn2024style}. (j) \textbf{TTMG (Ours)}. In this figure, \textcolor{green}{\textbf{Green}} and \textcolor{red}{\textbf{Red}} lines denote the boundaries of the ground truth and prediction, respectively.}
    \label{fig:Sup_CU_DR}
\end{figure*}

\begin{figure*}[t]
    \centering
    \includegraphics[width=\textwidth]{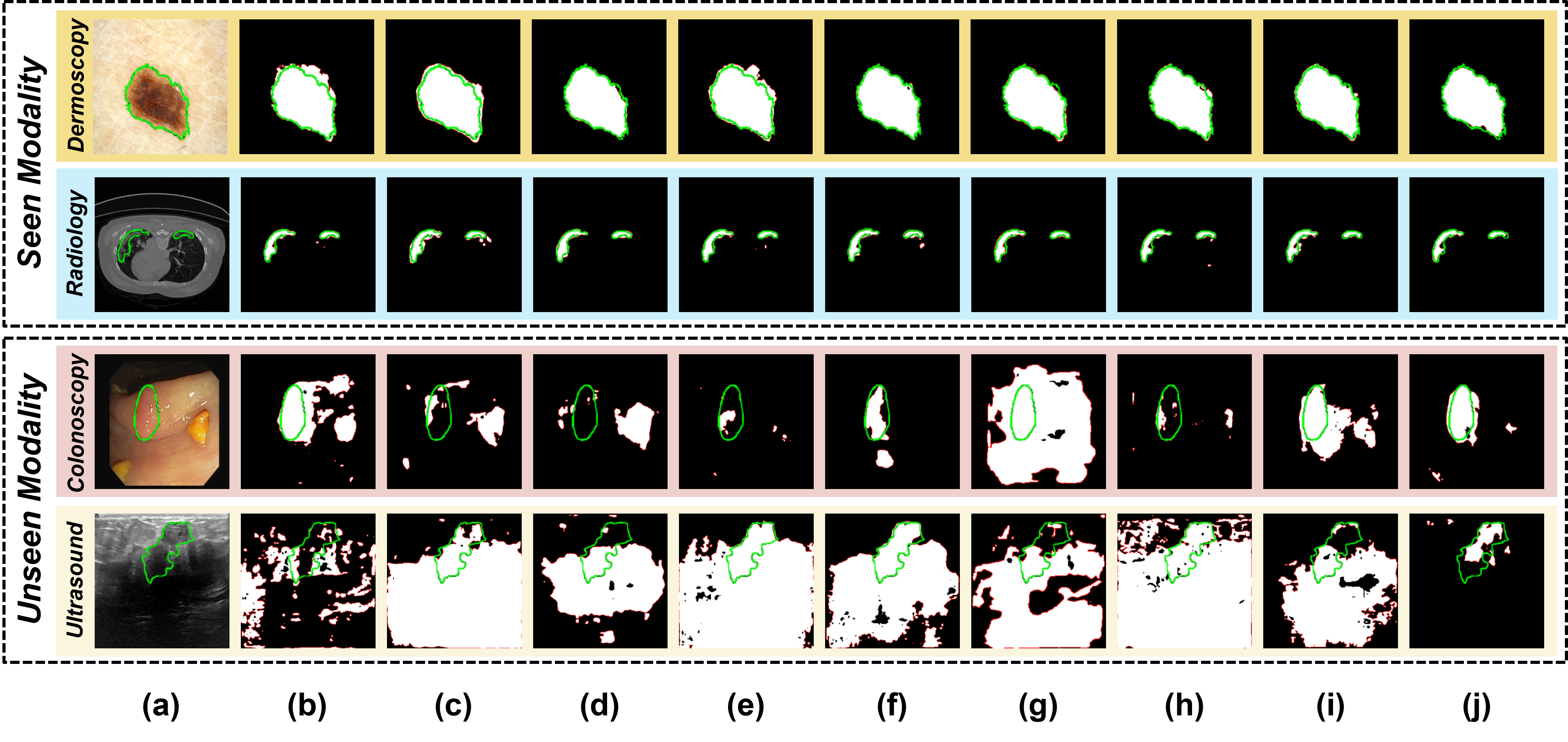}
    \caption{Qualitative comparison of other methods and TTMG on Dermoscopy and Radiology modalities.  (a) Input images with ground truth. (b) baseline (DeepLabV3+ with ResNet50, \cite{chen2018encoder}). (c) IN \cite{ulyanov2017improved}. (d) IW \cite{huang2018decorrelated}. (e) IBN \cite{pan2018two}. (f) RobustNet \cite{choi2021robustnet}. (g) SAN-SAW \cite{peng2022semantic}. (h) SPCNet \cite{huang2023style}. (i) BlindNet \cite{ahn2024style}. (j) \textbf{TTMG (Ours)}. In this figure, \textcolor{green}{\textbf{Green}} and \textcolor{red}{\textbf{Red}} lines denote the boundaries of the ground truth and prediction, respectively.}
    \label{fig:Sup_DR_CU}
\end{figure*}

\begin{figure*}[t]
    \centering
    \includegraphics[width=\textwidth]{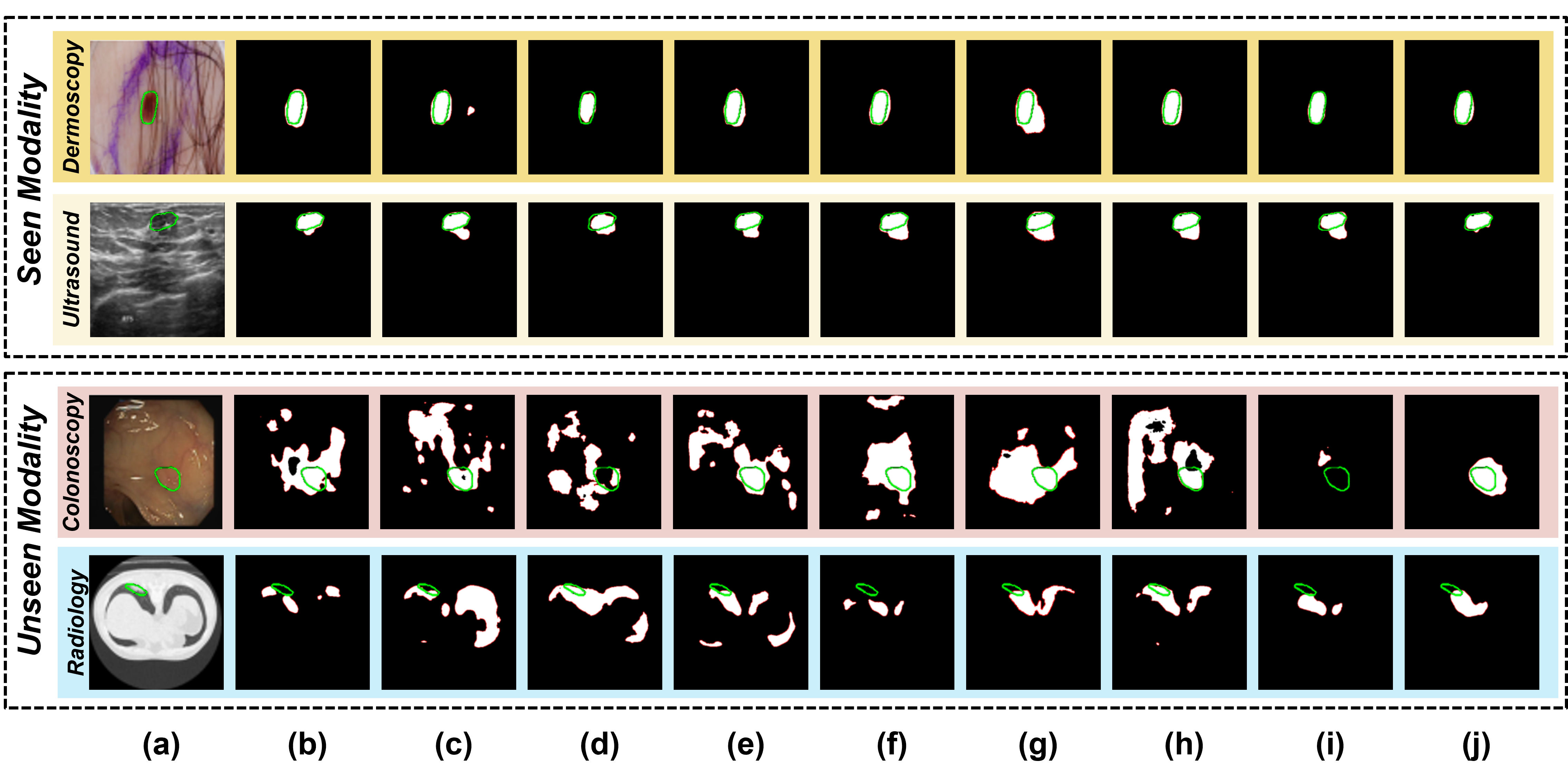}
    \caption{Qualitative comparison of other methods and TTMG on Dermoscopy and Ultrasound modalities.  (a) Input images with ground truth. (b) baseline (DeepLabV3+ with ResNet50, \cite{chen2018encoder}). (c) IN \cite{ulyanov2017improved}. (d) IW \cite{huang2018decorrelated}. (e) IBN \cite{pan2018two}. (f) RobustNet \cite{choi2021robustnet}. (g) SAN-SAW \cite{peng2022semantic}. (h) SPCNet \cite{huang2023style}. (i) BlindNet \cite{ahn2024style}. (j) \textbf{TTMG (Ours)}. In this figure, \textcolor{green}{\textbf{Green}} and \textcolor{red}{\textbf{Red}} lines denote the boundaries of the ground truth and prediction, respectively.}
    \label{fig:Sup_DU_CR}
\end{figure*}

\begin{figure*}[t]
    \centering
    \includegraphics[width=\textwidth]{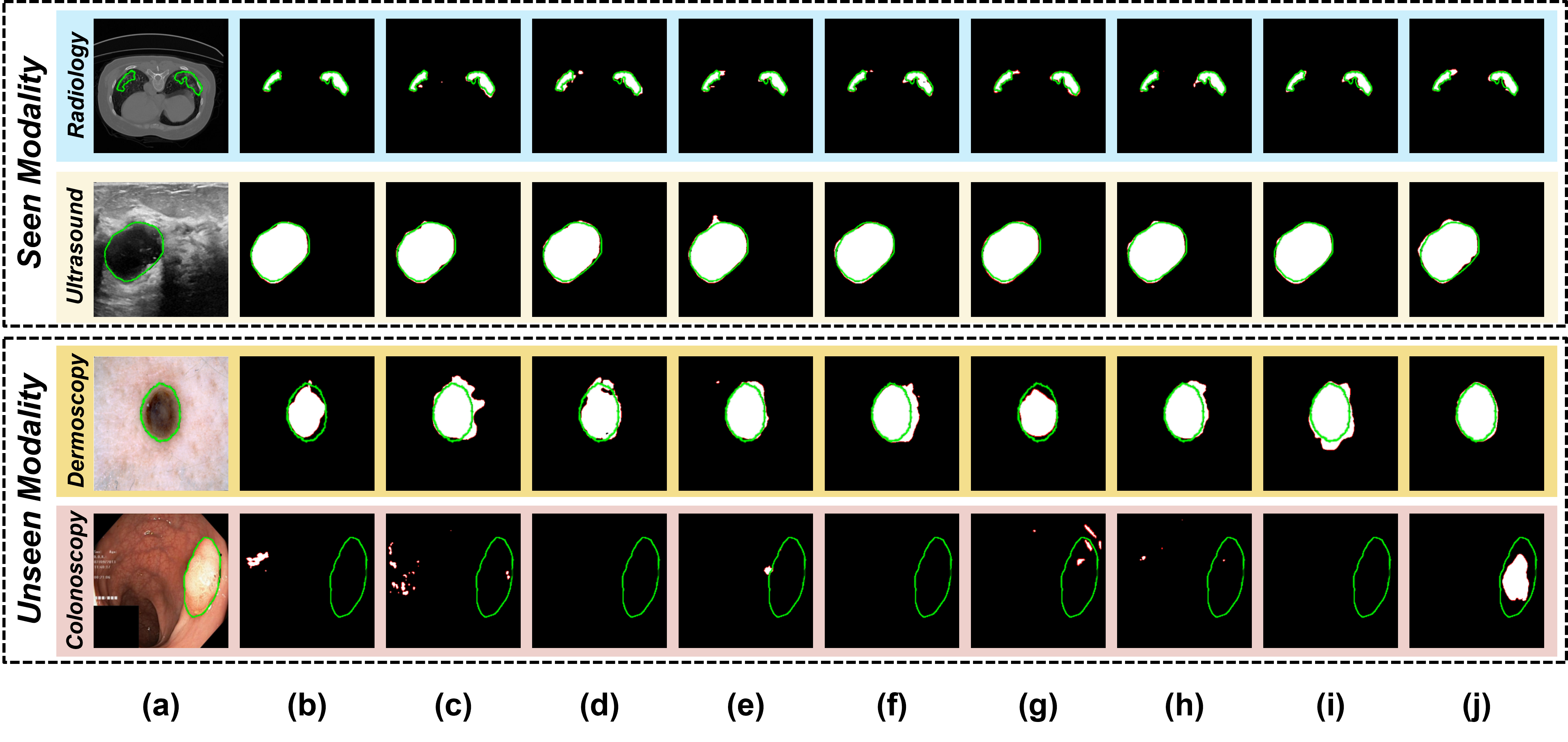}
    \caption{Qualitative comparison of other methods and TTMG on Radiology and Ultrasound modalities.  (a) Input images with ground truth. (b) baseline (DeepLabV3+ with ResNet50, \cite{chen2018encoder}). (c) IN \cite{ulyanov2017improved}. (d) IW \cite{huang2018decorrelated}. (e) IBN \cite{pan2018two}. (f) RobustNet \cite{choi2021robustnet}. (g) SAN-SAW \cite{peng2022semantic}. (h) SPCNet \cite{huang2023style}. (i) BlindNet \cite{ahn2024style}. (j) \textbf{TTMG (Ours)}. In this figure, \textcolor{green}{\textbf{Green}} and \textcolor{red}{\textbf{Red}} lines denote the boundaries of the ground truth and prediction, respectively.}
    \label{fig:Sup_RU_CD}
\end{figure*}

In this section, we provide the quantitative results with various metrics, which are listed in the Appendix \ref{appendix_metric_descriptions}, in Tables \ref{tab:comparison_other_three_metrics_average} and \ref{tab:comparison_other_two_metrics_average}. And, $( \cdot )$ denotes the performance gap between the baseline and each method. These experimental results demonstrate that TTMG consistently achieves significant performance improvements across various structural metrics, extending beyond DSC and mIoU. Additionally, we present experimental results for both the three-modality (Tables \ref{tab:comparison_CUD_other_metrics}, \ref{tab:comparison_CDR_other_metrics}, \ref{tab:comparison_CUR_other_metrics}, and \ref{tab:comparison_UDR_other_metrics}) and two-modality (Tables \ref{tab:comparison_CD_other_metrics}, \ref{tab:comparison_CR_other_metrics}, \ref{tab:comparison_CU_other_metrics}, \ref{tab:comparison_DU_other_metrics}, \ref{tab:comparison_DR_other_metrics}, and \ref{tab:comparison_RU_other_metrics}) training schemes. We also present more various qualitative results for both the three-modality (Figures \ref{fig:Sup_CDU_R}, \ref{fig:Sup_CUR_D} \ref{fig:Sup_CDR_U}, and \ref{fig:Sup_UDR_C}) and two-modality (Figures \ref{fig:Sup_CD_RU}, \ref{fig:Sup_CR_DU}, \ref{fig:Sup_CU_DR}, \ref{fig:Sup_DR_CU}, \ref{fig:Sup_DU_CR}, and \ref{fig:Sup_RU_CD}) training schemes. 

\begin{table*}[h]
    \centering
    \tiny
    \setlength\tabcolsep{0.1pt} 
    \renewcommand{\arraystretch}{1.15} 
\begin{tabular}{c|ccccc|ccccc|ccccc|ccccc|ccccc|ccccc}
    \hline
    \multicolumn{1}{c|}{\multirow{2}{*}{Method}} & \multicolumn{5}{c|}{BUSI \cite{al2020dataset}} & \multicolumn{5}{c|}{STU \cite{zhuang2019rdau}} & \multicolumn{5}{c|}{ISIC2018 \cite{gutman2016skin}} & \multicolumn{5}{c|}{PH2 \cite{mendoncca2013ph}} & \multicolumn{5}{c|}{COVID19-1 \cite{ma_jun_2020_3757476}}  & \multicolumn{5}{c}{COVID19-2 \cite{COVID19_2}}\\ \cline{2-31}
     & DSC & mIoU & $F_{\beta}^{w}$ & $S_{\alpha}$ & $E_{\phi}^{m}$ & DSC & mIoU & $F_{\beta}^{w}$ & $S_{\alpha}$ & $E_{\phi}^{m}$ & DSC & mIoU & $F_{\beta}^{w}$ & $S_{\alpha}$ & $E_{\phi}^{m}$ & DSC & mIoU & $F_{\beta}^{w}$ & $S_{\alpha}$ & $E_{\phi}^{m}$ & DSC & mIoU & $F_{\beta}^{w}$ & $S_{\alpha}$ & $E_{\phi}^{m}$ & DSC & mIoU & $F_{\beta}^{w}$ & $S_{\alpha}$ & $E_{\phi}^{m}$ \\
     \hline
     Baseline \cite{chen2018encoder}                                           & 78.76 & 70.05 & 76.65 & 82.13 & 90.70 & 82.68 & 73.30 & 80.28 & 82.63 & 92.18 & 88.91 & 81.87 & 87.96 & 80.86 & 92.76 & 90.82 & 83.82 & 87.94 & 75.43 & 92.20 & 15.90 & 10.65 & 11.18 & 60.66 &	56.63 &	12.70 &	7.58 & 8.77 & 54.24 & 40.77 \\
     \hline
     \multicolumn{1}{c|}{\multirow{2}{*}{IN \cite{ulyanov2017improved}}}       & 77.95 & 69.03 & 75.80 & 81.49 & 90.49 & 86.38 & 78.05 & 84.33 & 85.05 & 95.40 & 89.51 & 82.54 & 88.64 & 81.43 & 93.42 & 91.28 & 84.52 & 88.47 & 75.73 & 92.58 & 16.50 & 10.83 & 11.85 & 60.36 & 52.55 & 11.34 & 6.74 & 7.64 & 54.00 & 36.21 \\
\end{tabular}
    \caption{Segmentation results with training scheme of (C, U, D). We train each model on C (CVC-ClinicDB \cite{bernal2015wm} + Kvasir-SEG \cite{jha2020kvasir}) + U (BUSI \cite{al2020dataset}) + D (ISIC2018 \cite{gutman2016skin}) train dataset and evaluate on test datasets of seen modalities (C \cite{bernal2015wm, jha2020kvasir, vazquez2017benchmark, tajbakhsh2015automated, silva2014toward} + U \cite{al2020dataset, zhuang2019rdau} + D \cite{gutman2016skin, mendoncca2013ph}) and unseen modality (R \cite{ma_jun_2020_3757476, COVID19_2}).}
    \label{tab:comparison_CUD_other_metrics}
\end{table*}

\begin{table*}[h]
    \centering
    \tiny
    \setlength\tabcolsep{0.1pt} 
    \renewcommand{\arraystretch}{1.15} 
\begin{tabular}{c|ccccc|ccccc|ccccc|ccccc|ccccc|ccccc}
    \hline
    \multicolumn{1}{c|}{\multirow{2}{*}{Method}} & \multicolumn{5}{c|}{BUSI \cite{al2020dataset}} & \multicolumn{5}{c|}{STU \cite{zhuang2019rdau}} & \multicolumn{5}{c|}{ISIC2018 \cite{gutman2016skin}} & \multicolumn{5}{c|}{PH2 \cite{mendoncca2013ph}} & \multicolumn{5}{c|}{COVID19-1 \cite{ma_jun_2020_3757476}}  & \multicolumn{5}{c}{COVID19-2 \cite{COVID19_2}}\\ \cline{2-31}
     & DSC & mIoU & $F_{\beta}^{w}$ & $S_{\alpha}$ & $E_{\phi}^{m}$ & DSC & mIoU & $F_{\beta}^{w}$ & $S_{\alpha}$ & $E_{\phi}^{m}$ & DSC & mIoU & $F_{\beta}^{w}$ & $S_{\alpha}$ & $E_{\phi}^{m}$ & DSC & mIoU & $F_{\beta}^{w}$ & $S_{\alpha}$ & $E_{\phi}^{m}$ & DSC & mIoU & $F_{\beta}^{w}$ & $S_{\alpha}$ & $E_{\phi}^{m}$ & DSC & mIoU & $F_{\beta}^{w}$ & $S_{\alpha}$ & $E_{\phi}^{m}$ \\
     \hline
     Baseline \cite{chen2018encoder}    & 78.71 & 70.27 & 76.88 & 82.20 & 90.24 & 85.38 & 77.09 & 83.77 & 84.58 & 93.55 & 47.63 & 40.44 & 48.64 & 59.81 & 62.58 & 37.23 & 32.03 & 38.09 & 49.39 & 53.22 & 75.75 & 68.37 & 64.30 & 82.70 & 77.48 & 43.73 & 32.80 & 45.26 & 65.31 & 70.03 \\
     \hline
     \multicolumn{1}{c|}{\multirow{2}{*}{IN \cite{ulyanov2017improved}}}      & 75.36 & 67.30 & 74.10 & 80.52 & 88.14 & 80.24 & 70.19 & 78.92 & 81.09 & 90.41 & 48.86 & 40.28 & 49.48 & 59.79 & 63.37 & 47.44 & 39.42 & 48.09 & 52.96 & 60.82 & 77.96 & 70.51 & 66.03 & 83.43 & 79.69 & 44.03 & 33.06 & 43.32 & 65.75 & 73.45 \\
\end{tabular}
    \caption{Segmentation results with training scheme of (C, U, R). We train each model on C (CVC-ClinicDB \cite{bernal2015wm} + Kvasir-SEG \cite{jha2020kvasir}) + U (BUSI \cite{al2020dataset}) + R (COVID19-1 \cite{ma_jun_2020_3757476}) train dataset and evaluate on test datasets of seen modalities (C \cite{bernal2015wm, jha2020kvasir, vazquez2017benchmark, tajbakhsh2015automated, silva2014toward} + U \cite{al2020dataset, zhuang2019rdau} + R \cite{ma_jun_2020_3757476, COVID19_2}) and unseen modality (D \cite{gutman2016skin, mendoncca2013ph}).}
    \label{tab:comparison_CUR_other_metrics}
\end{table*}

\begin{table*}[h]
    \centering
    \tiny
    \setlength\tabcolsep{0.1pt} 
    \renewcommand{\arraystretch}{1.15} 
\begin{tabular}{c|ccccc|ccccc|ccccc|ccccc|ccccc|ccccc}
    \hline
    \multicolumn{1}{c|}{\multirow{2}{*}{Method}} & \multicolumn{5}{c|}{BUSI \cite{al2020dataset}} & \multicolumn{5}{c|}{STU \cite{zhuang2019rdau}} & \multicolumn{5}{c|}{ISIC2018 \cite{gutman2016skin}} & \multicolumn{5}{c|}{PH2 \cite{mendoncca2013ph}} & \multicolumn{5}{c|}{COVID19-1 \cite{ma_jun_2020_3757476}}  & \multicolumn{5}{c}{COVID19-2 \cite{COVID19_2}}\\ \cline{2-31}
     & DSC & mIoU & $F_{\beta}^{w}$ & $S_{\alpha}$ & $E_{\phi}^{m}$ & DSC & mIoU & $F_{\beta}^{w}$ & $S_{\alpha}$ & $E_{\phi}^{m}$ & DSC & mIoU & $F_{\beta}^{w}$ & $S_{\alpha}$ & $E_{\phi}^{m}$ & DSC & mIoU & $F_{\beta}^{w}$ & $S_{\alpha}$ & $E_{\phi}^{m}$ & DSC & mIoU & $F_{\beta}^{w}$ & $S_{\alpha}$ & $E_{\phi}^{m}$ & DSC & mIoU & $F_{\beta}^{w}$ & $S_{\alpha}$ & $E_{\phi}^{m}$ \\
     \hline
     Baseline \cite{chen2018encoder}    & 14.16 & 8.70 & 11.61 & 47.95 & 52.27 & 12.94 & 7.61 & 9.74 & 39.30 & 50.90 & 88.68 & 81.56 & 87.75 & 80.83 & 92.98 & 90.90 & 83.95 & 88.15 & 75.80 & 92.06 & 78.51 & 71.06 & 66.73 & 83.66 & 79.09 & 46.68 & 35.29 & 48.14 & 66.20 & 72.81 \\
     \hline
     \multicolumn{1}{c|}{\multirow{2}{*}{IN \cite{ulyanov2017improved}}}      & 21.85 & 14.17 & 15.97 & 52.97 & 39.02 & 31.79 & 21.71 & 24.09 & 54.08 & 49.79 & 89.41 & 82.45 & 88.66 & 81.38 & 93.25 & 91.24 & 84.58 & 88.53 & 75.87 & 92.73 & 77.54 & 70.43 & 65.66 & 83.92 & 79.17 & 45.91 & 34.49 & 45.58 & 66.46 & 74.53 \\
\end{tabular}
    \caption{Segmentation results with training scheme of (C, D, R). We train each model on C (CVC-ClinicDB \cite{bernal2015wm} + Kvasir-SEG \cite{jha2020kvasir}) + D (ISIC2018 \cite{gutman2016skin}) + R (COVID19-1 \cite{ma_jun_2020_3757476}) train dataset and evaluate on test datasets of seen modalities (C \cite{bernal2015wm, jha2020kvasir, vazquez2017benchmark, tajbakhsh2015automated, silva2014toward} + D \cite{gutman2016skin, mendoncca2013ph} + R \cite{ma_jun_2020_3757476, COVID19_2}) and unseen modality (U \cite{al2020dataset, zhuang2019rdau}).}
    \label{tab:comparison_CDR_other_metrics}
\end{table*}

\begin{table*}[h]
    \centering
    \tiny
    \setlength\tabcolsep{0.1pt} 
    \renewcommand{\arraystretch}{1.15} 
\begin{tabular}{c|ccccc|ccccc|ccccc|ccccc|ccccc|ccccc}
    \hline
    \multicolumn{1}{c|}{\multirow{2}{*}{Method}} & \multicolumn{5}{c|}{BUSI \cite{al2020dataset}} & \multicolumn{5}{c|}{STU \cite{zhuang2019rdau}} & \multicolumn{5}{c|}{ISIC2018 \cite{gutman2016skin}} & \multicolumn{5}{c|}{PH2 \cite{mendoncca2013ph}} & \multicolumn{5}{c|}{COVID19-1 \cite{ma_jun_2020_3757476}}  & \multicolumn{5}{c}{COVID19-2 \cite{COVID19_2}}\\ \cline{2-31}
     & DSC & mIoU & $F_{\beta}^{w}$ & $S_{\alpha}$ & $E_{\phi}^{m}$ & DSC & mIoU & $F_{\beta}^{w}$ & $S_{\alpha}$ & $E_{\phi}^{m}$ & DSC & mIoU & $F_{\beta}^{w}$ & $S_{\alpha}$ & $E_{\phi}^{m}$ & DSC & mIoU & $F_{\beta}^{w}$ & $S_{\alpha}$ & $E_{\phi}^{m}$ & DSC & mIoU & $F_{\beta}^{w}$ & $S_{\alpha}$ & $E_{\phi}^{m}$ & DSC & mIoU & $F_{\beta}^{w}$ & $S_{\alpha}$ & $E_{\phi}^{m}$ \\
     \hline
     Baseline \cite{chen2018encoder}    & 76.74 & 68.41 & 75.20 & 81.19 & 89.54 & 80.55 & 70.91 & 78.50 & 81.99 & 91.23 & 88.27 & 81.09 & 87.38 & 80.54 & 92.24 & 91.52 & 84.96 & 89.08 & 75.86 & 92.52 & 78.28 & 71.09 & 66.12 & 83.93 & 79.59 & 42.58 & 31.46 & 41.67 & 65.22 & 73.60 \\
     \hline
     \multicolumn{1}{c|}{\multirow{2}{*}{IN \cite{ulyanov2017improved}}}      & 72.09 & 62.24 & 70.41 & 78.08 & 88.50 & 81.91 & 71.61 & 79.65 & 81.55 & 92.44 & 88.54 & 81.19 & 88.12 & 80.67 & 92.70 & 90.74 & 84.08 & 88.42 & 75.46 & 91.89 & 68.89 & 60.11 & 59.89 & 80.32 & 78.45 & 44.76 & 32.71 & 42.71 & 65.59 & 76.44 \\
\end{tabular}
    \caption{Segmentation results with training scheme of (U, D, R). We train each model on U (BUSI \cite{al2020dataset}) + D (ISIC2018 \cite{gutman2016skin}) + R (COVID19-1 \cite{ma_jun_2020_3757476}) train dataset and evaluate on test datasets of seen modalities (U \cite{al2020dataset, zhuang2019rdau} + D \cite{gutman2016skin, mendoncca2013ph} + R \cite{ma_jun_2020_3757476, COVID19_2}) and unseen modality (C \cite{bernal2015wm, jha2020kvasir, vazquez2017benchmark, tajbakhsh2015automated, silva2014toward}).}
    \label{tab:comparison_UDR_other_metrics}
\end{table*}

\begin{table*}[h]
    \centering
    \tiny
    \setlength\tabcolsep{0.1pt} 
    \renewcommand{\arraystretch}{1.15} 
\begin{tabular}{c|ccccc|ccccc|ccccc|ccccc|ccccc|ccccc}
    \hline
    \multicolumn{1}{c|}{\multirow{2}{*}{Method}} & \multicolumn{5}{c|}{BUSI \cite{al2020dataset}} & \multicolumn{5}{c|}{STU \cite{zhuang2019rdau}} & \multicolumn{5}{c|}{ISIC2018 \cite{gutman2016skin}} & \multicolumn{5}{c|}{PH2 \cite{mendoncca2013ph}} & \multicolumn{5}{c|}{COVID19-1 \cite{ma_jun_2020_3757476}}  & \multicolumn{5}{c}{COVID19-2 \cite{COVID19_2}}\\ \cline{2-31}
     & DSC & mIoU & $F_{\beta}^{w}$ & $S_{\alpha}$ & $E_{\phi}^{m}$ & DSC & mIoU & $F_{\beta}^{w}$ & $S_{\alpha}$ & $E_{\phi}^{m}$ & DSC & mIoU & $F_{\beta}^{w}$ & $S_{\alpha}$ & $E_{\phi}^{m}$ & DSC & mIoU & $F_{\beta}^{w}$ & $S_{\alpha}$ & $E_{\phi}^{m}$ & DSC & mIoU & $F_{\beta}^{w}$ & $S_{\alpha}$ & $E_{\phi}^{m}$ & DSC & mIoU & $F_{\beta}^{w}$ & $S_{\alpha}$ & $E_{\phi}^{m}$ \\
     \hline
     Baseline \cite{chen2018encoder}    & 18.51 & 11.26 & 11.67 & 46.80 & 18.34 & 22.93 & 13.59 & 13.90 & 26.97 & 14.76 & 89.49 & 82.45 & 88.58 & 81.35 & 93.35 & 91.58 & 85.03 & 89.07 & 76.26 & 92.61 & 9.23 & 6.09 & 6.42 & 58.14 & 53.92 & 8.57 & 5.04 & 6.48 & 52.74 & 52.25 \\
     \hline
     \multicolumn{1}{c|}{\multirow{2}{*}{IN \cite{ulyanov2017improved}}} & 24.69 & 16.61 & 17.49 & 51.89 & 31.10 
                                                                         & 37.87 & 25.89 & 26.97 & 54.22 & 40.00 
                                                                         & 89.00 & 82.02 & 88.03 & 81.17 & 92.93 
                                                                         & 91.23 & 84.45 & 88.34 & 75.86 & 92.54 
                                                                         & 14.30 & 9.52  & 11.06 & 57.56 & 47.23 
                                                                         & 10.56 & 6.33  & 7.25  & 54.16 & 40.24 \\
\end{tabular}
    \caption{Segmentation results with training scheme of (C, D). We train each model on C (CVC-ClinicDB \cite{bernal2015wm} + Kvasir-SEG \cite{jha2020kvasir}) + D (ISIC2018 \cite{gutman2016skin}) train dataset and evaluate on test datasets of seen modalities (C \cite{bernal2015wm, jha2020kvasir, vazquez2017benchmark, tajbakhsh2015automated, silva2014toward} + D \cite{gutman2016skin, mendoncca2013ph}) and unseen modality (R \cite{ma_jun_2020_3757476, COVID19_2} + U \cite{al2020dataset, zhuang2019rdau}).}
    \label{tab:comparison_CD_other_metrics}
\end{table*}

\begin{table*}[h]
    \centering
    \tiny
    \setlength\tabcolsep{0.1pt} 
    \renewcommand{\arraystretch}{1.15} 
\begin{tabular}{c|ccccc|ccccc|ccccc|ccccc|ccccc|ccccc}
    \hline
    \multicolumn{1}{c|}{\multirow{2}{*}{Method}} & \multicolumn{5}{c|}{BUSI \cite{al2020dataset}} & \multicolumn{5}{c|}{STU \cite{zhuang2019rdau}} & \multicolumn{5}{c|}{ISIC2018 \cite{gutman2016skin}} & \multicolumn{5}{c|}{PH2 \cite{mendoncca2013ph}} & \multicolumn{5}{c|}{COVID19-1 \cite{ma_jun_2020_3757476}}  & \multicolumn{5}{c}{COVID19-2 \cite{COVID19_2}}\\ \cline{2-31}
     & DSC & mIoU & $F_{\beta}^{w}$ & $S_{\alpha}$ & $E_{\phi}^{m}$ & DSC & mIoU & $F_{\beta}^{w}$ & $S_{\alpha}$ & $E_{\phi}^{m}$ & DSC & mIoU & $F_{\beta}^{w}$ & $S_{\alpha}$ & $E_{\phi}^{m}$ & DSC & mIoU & $F_{\beta}^{w}$ & $S_{\alpha}$ & $E_{\phi}^{m}$ & DSC & mIoU & $F_{\beta}^{w}$ & $S_{\alpha}$ & $E_{\phi}^{m}$ & DSC & mIoU & $F_{\beta}^{w}$ & $S_{\alpha}$ & $E_{\phi}^{m}$ \\
     \hline
     Baseline \cite{chen2018encoder}    & 1.41 & 0.73 & 1.86 & 45.47 & 32.78 & 0.69 & 0.35 & 0.90 & 43.96 & 26.37 & 10.25 & 7.50 & 10.99 & 42.44 & 33.51 & 6.93 & 5.03 & 7.62 & 35.30 & 29.58 & 77.10 & 69.54 & 65.24 & 83.10 & 77.82 & 41.60 & 30.96 & 42.30 & 64.69 & 71.25 \\
     \hline
     \multicolumn{1}{c|}{\multirow{2}{*}{IN \cite{ulyanov2017improved}}} & 0.57  & 0.29  & 0.71  & 45.40 & 27.99 
                                                                         & 0.23  & 0.11  & 0.29  & 43.94 & 25.61 
                                                                         & 3.34  & 2.07  & 3.71  & 39.89 & 28.78 
                                                                         & 4.77  & 3.47  & 5.34  & 34.54 & 28.25 
                                                                         & 77.64 & 70.29 & 65.96 & 83.55 & 79.59
                                                                         & 46.26 & 35.08 & 46.45 & 66.34 & 74.84 \\
\end{tabular}
    \caption{Segmentation results with training scheme of (C, R). We train each model on C (CVC-ClinicDB \cite{bernal2015wm} + Kvasir-SEG \cite{jha2020kvasir}) + R (COVID19-1 \cite{ma_jun_2020_3757476}) train dataset and evaluate on test datasets of seen modalities (C \cite{bernal2015wm, jha2020kvasir, vazquez2017benchmark, tajbakhsh2015automated, silva2014toward} + R \cite{ma_jun_2020_3757476, COVID19_2}) and unseen modality (D \cite{gutman2016skin, mendoncca2013ph} + U \cite{al2020dataset, zhuang2019rdau}).}
    \label{tab:comparison_CR_other_metrics}
\end{table*}

\begin{table*}[h]
    \centering
    \tiny
    \setlength\tabcolsep{0.1pt} 
    \renewcommand{\arraystretch}{1.15} 
\begin{tabular}{c|ccccc|ccccc|ccccc|ccccc|ccccc|ccccc}
    \hline
    \multicolumn{1}{c|}{\multirow{2}{*}{Method}} & \multicolumn{5}{c|}{BUSI \cite{al2020dataset}} & \multicolumn{5}{c|}{STU \cite{zhuang2019rdau}} & \multicolumn{5}{c|}{ISIC2018 \cite{gutman2016skin}} & \multicolumn{5}{c|}{PH2 \cite{mendoncca2013ph}} & \multicolumn{5}{c|}{COVID19-1 \cite{ma_jun_2020_3757476}}  & \multicolumn{5}{c}{COVID19-2 \cite{COVID19_2}}\\ \cline{2-31}
     & DSC & mIoU & $F_{\beta}^{w}$ & $S_{\alpha}$ & $E_{\phi}^{m}$ & DSC & mIoU & $F_{\beta}^{w}$ & $S_{\alpha}$ & $E_{\phi}^{m}$ & DSC & mIoU & $F_{\beta}^{w}$ & $S_{\alpha}$ & $E_{\phi}^{m}$ & DSC & mIoU & $F_{\beta}^{w}$ & $S_{\alpha}$ & $E_{\phi}^{m}$ & DSC & mIoU & $F_{\beta}^{w}$ & $S_{\alpha}$ & $E_{\phi}^{m}$ & DSC & mIoU & $F_{\beta}^{w}$ & $S_{\alpha}$ & $E_{\phi}^{m}$ \\
     \hline
     Baseline \cite{chen2018encoder} & 78.19 & 69.54 & 75.92 & 81.95 & 89.49 & 84.07 & 74.65 & 81.30 & 83.57 & 93.16 & 51.41 & 43.72 & 51.81 & 61.78 & 65.21 & 45.50 & 39.16 & 45.72 & 53.03 & 59.01 & 16.36 & 11.28 & 12.18 & 60.89 & 58.91 & 13.85 & 8.57 & 10.53 & 54.53 & 48.97 \\
     \hline
     \multicolumn{1}{c|}{\multirow{2}{*}{IN \cite{ulyanov2017improved}}} & 77.20 & 68.45 & 75.18 & 81.33 & 89.50
                                                                         & 85.49 & 75.92 & 83.50 & 84.15 & 93.70 
                                                                         & 59.83 & 48.38 & 59.93 & 63.66 & 72.60 
                                                                         & 56.89 & 50.02 & 56.54 & 58.77 & 67.35 
                                                                         & 11.15 & 7.04  & 8.42  & 58.87 & 60.55
                                                                         & 10.72 & 6.45 & 7.73 & 53.39 & 45.09 \\
\end{tabular}
    \caption{Segmentation results with training scheme of (C, U). We train each model on C (CVC-ClinicDB \cite{bernal2015wm} + Kvasir-SEG \cite{jha2020kvasir}) + U (BUSI \cite{al2020dataset}) train dataset and evaluate on test datasets of seen modalities (C \cite{bernal2015wm, jha2020kvasir, vazquez2017benchmark, tajbakhsh2015automated, silva2014toward} + U \cite{al2020dataset, zhuang2019rdau}) and unseen modality (D \cite{gutman2016skin, mendoncca2013ph} + R \cite{ma_jun_2020_3757476, COVID19_2}).}
    \label{tab:comparison_CU_other_metrics}
\end{table*}

\begin{table*}[h]
    \centering
    \tiny
    \setlength\tabcolsep{0.1pt} 
    \renewcommand{\arraystretch}{1.15} 
\begin{tabular}{c|ccccc|ccccc|ccccc|ccccc|ccccc|ccccc}
    \hline
    \multicolumn{1}{c|}{\multirow{2}{*}{Method}} & \multicolumn{5}{c|}{BUSI \cite{al2020dataset}} & \multicolumn{5}{c|}{STU \cite{zhuang2019rdau}} & \multicolumn{5}{c|}{ISIC2018 \cite{gutman2016skin}} & \multicolumn{5}{c|}{PH2 \cite{mendoncca2013ph}} & \multicolumn{5}{c|}{COVID19-1 \cite{ma_jun_2020_3757476}}  & \multicolumn{5}{c}{COVID19-2 \cite{COVID19_2}}\\ \cline{2-31}
     & DSC & mIoU & $F_{\beta}^{w}$ & $S_{\alpha}$ & $E_{\phi}^{m}$ & DSC & mIoU & $F_{\beta}^{w}$ & $S_{\alpha}$ & $E_{\phi}^{m}$ & DSC & mIoU & $F_{\beta}^{w}$ & $S_{\alpha}$ & $E_{\phi}^{m}$ & DSC & mIoU & $F_{\beta}^{w}$ & $S_{\alpha}$ & $E_{\phi}^{m}$ & DSC & mIoU & $F_{\beta}^{w}$ & $S_{\alpha}$ & $E_{\phi}^{m}$ & DSC & mIoU & $F_{\beta}^{w}$ & $S_{\alpha}$ & $E_{\phi}^{m}$ \\
     \hline
     Baseline \cite{chen2018encoder}    & 74.55 & 65.40 & 72.95 & 79.24 & 88.79 & 85.68 & 76.98 & 83.19 & 84.91 & 94.04 & 87.75 & 80.36 & 86.77 & 80.39 & 92.02 & 90.43 & 83.49 & 87.74 & 74.64 & 91.22 & 20.17 & 13.61 & 14.85 & 62.68 & 56.41 & 11.54 & 6.80 & 8.00 & 53.57 & 38.66 \\
     \hline
     \multicolumn{1}{c|}{\multirow{2}{*}{IN \cite{ulyanov2017improved}}} & 76.51 & 67.46 & 73.99 & 80.66 & 89.49 
                                                                         & 86.05 & 77.61 & 83.58 & 85.15 & 94.13 
                                                                         & 89.24 & 82.22 & 88.18 & 81.25 & 93.10 
                                                                         & 91.01 & 84.11 & 88.22 & 75.26 & 92.39
                                                                         & 24.32 & 17.00 & 16.23 & 62.92 & 49.69 
                                                                         & 12.89 & 7.69 & 8.40 & 54.32 & 33.74 \\
\end{tabular}
    \caption{Segmentation results with training scheme of (D, U). We train each model on D (ISIC2018 \cite{gutman2016skin}) + U (BUSI \cite{al2020dataset}) train dataset and evaluate on test datasets of seen modalities (D \cite{gutman2016skin, mendoncca2013ph} + U \cite{al2020dataset, zhuang2019rdau}) and unseen modality (C \cite{bernal2015wm, jha2020kvasir, vazquez2017benchmark, tajbakhsh2015automated, silva2014toward} + R \cite{ma_jun_2020_3757476, COVID19_2}).}
    \label{tab:comparison_DU_other_metrics}
\end{table*}

\begin{table*}[h]
    \centering
    \tiny
    \setlength\tabcolsep{0.1pt} 
    \renewcommand{\arraystretch}{1.15} 
\begin{tabular}{c|ccccc|ccccc|ccccc|ccccc|ccccc|ccccc}
    \hline
    \multicolumn{1}{c|}{\multirow{2}{*}{Method}} & \multicolumn{5}{c|}{BUSI \cite{al2020dataset}} & \multicolumn{5}{c|}{STU \cite{zhuang2019rdau}} & \multicolumn{5}{c|}{ISIC2018 \cite{gutman2016skin}} & \multicolumn{5}{c|}{PH2 \cite{mendoncca2013ph}} & \multicolumn{5}{c|}{COVID19-1 \cite{ma_jun_2020_3757476}}  & \multicolumn{5}{c}{COVID19-2 \cite{COVID19_2}}\\ \cline{2-31}
     & DSC & mIoU & $F_{\beta}^{w}$ & $S_{\alpha}$ & $E_{\phi}^{m}$ & DSC & mIoU & $F_{\beta}^{w}$ & $S_{\alpha}$ & $E_{\phi}^{m}$ & DSC & mIoU & $F_{\beta}^{w}$ & $S_{\alpha}$ & $E_{\phi}^{m}$ & DSC & mIoU & $F_{\beta}^{w}$ & $S_{\alpha}$ & $E_{\phi}^{m}$ & DSC & mIoU & $F_{\beta}^{w}$ & $S_{\alpha}$ & $E_{\phi}^{m}$ & DSC & mIoU & $F_{\beta}^{w}$ & $S_{\alpha}$ & $E_{\phi}^{m}$ \\
     \hline
     Baseline \cite{chen2018encoder} & 12.67 & 7.58  & 11.49 & 47.99 & 62.64 & 13.20 & 7.81  & 10.79 & 44.56 & 58.30 & 88.86 & 81.94 & 88.14 & 80.88 & 93.20 & 91.47 & 84.77 & 88.80 & 76.03 & 92.60 & 77.00 & 69.50 & 65.58 & 83.04 & 78.95  & 44.35 & 33.36 & 45.38 & 65.43 & 71.71 \\
     \hline
     \multicolumn{1}{c|}{\multirow{2}{*}{IN \cite{ulyanov2017improved}}} & 19.85 & 12.46 & 13.46 & 50.97 & 32.17 
                                                                         & 33.44 & 21.49 & 22.30 & 51.40 & 37.85 
                                                                         & 85.68 & 77.38 & 82.37 & 78.89 & 90.61 
                                                                         & 88.11 & 79.57 & 83.33 & 73.56 & 90.34 
                                                                         & 35.50 & 27.19 & 28.07 & 71.76 & 62.06 
                                                                         & 18.93 & 12.59 & 14.70 & 58.24 & 43.23 \\
\end{tabular}
    \caption{Segmentation results with training scheme of (D, R). We train each model on D (ISIC2018 \cite{gutman2016skin}) + R (COVID19-1 \cite{ma_jun_2020_3757476}) train dataset and evaluate on test datasets of seen modalities (D \cite{gutman2016skin, mendoncca2013ph} + R \cite{ma_jun_2020_3757476, COVID19_2}) and unseen modality (C \cite{bernal2015wm, jha2020kvasir, vazquez2017benchmark, tajbakhsh2015automated, silva2014toward} + U \cite{al2020dataset, zhuang2019rdau}).}
    \label{tab:comparison_DR_other_metrics}
\end{table*}

\begin{table*}[h]
    \centering
    \tiny
    \setlength\tabcolsep{0.1pt} 
    \renewcommand{\arraystretch}{1.15} 
\begin{tabular}{c|ccccc|ccccc|ccccc|ccccc|ccccc|ccccc}
    \hline
    \multicolumn{1}{c|}{\multirow{2}{*}{Method}} & \multicolumn{5}{c|}{BUSI \cite{al2020dataset}} & \multicolumn{5}{c|}{STU \cite{zhuang2019rdau}} & \multicolumn{5}{c|}{ISIC2018 \cite{gutman2016skin}} & \multicolumn{5}{c|}{PH2 \cite{mendoncca2013ph}} & \multicolumn{5}{c|}{COVID19-1 \cite{ma_jun_2020_3757476}}  & \multicolumn{5}{c}{COVID19-2 \cite{COVID19_2}}\\ \cline{2-31}
     & DSC & mIoU & $F_{\beta}^{w}$ & $S_{\alpha}$ & $E_{\phi}^{m}$ & DSC & mIoU & $F_{\beta}^{w}$ & $S_{\alpha}$ & $E_{\phi}^{m}$ & DSC & mIoU & $F_{\beta}^{w}$ & $S_{\alpha}$ & $E_{\phi}^{m}$ & DSC & mIoU & $F_{\beta}^{w}$ & $S_{\alpha}$ & $E_{\phi}^{m}$ & DSC & mIoU & $F_{\beta}^{w}$ & $S_{\alpha}$ & $E_{\phi}^{m}$ & DSC & mIoU & $F_{\beta}^{w}$ & $S_{\alpha}$ & $E_{\phi}^{m}$ \\
     \hline
     Baseline \cite{chen2018encoder} & 78.26 & 69.20 & 76.02 & 81.71 & 90.48 & 85.08 & 76.16 & 83.59 & 84.10 & 95.10  & 14.36 & 9.38  & 14.87 & 43.17 & 48.24 & 9.06  & 5.35  & 9.95  & 34.84 & 33.98 & 78.19 & 70.72 & 66.63 & 83.67 & 79.18 & 41.54 & 30.28 & 40.60 & 64.78 & 72.42 \\
     \hline
     \multicolumn{1}{c|}{\multirow{2}{*}{IN \cite{ulyanov2017improved}}} & 76.25 & 67.63 & 75.12 & 80.59 & 89.77 
                                                                         & 82.19 & 71.78 & 80.50 & 82.50 & 93.16 
                                                                         & 65.76 & 55.81 & 65.21 & 68.08 & 76.26 
                                                                         & 63.07 & 52.62 & 62.71 & 59.90 & 69.99
                                                                         & 76.57 & 69.18 & 64.62 & 83.08 & 77.96 
                                                                         & 46.94 & 35.25 & 46.82 & 66.51 & 75.97 \\
\end{tabular}
    \caption{Segmentation results with training scheme of (R, U). We train each model on R (COVID19-1 \cite{ma_jun_2020_3757476}) + U (BUSI (al2020dataset)) train dataset and evaluate on test datasets of seen modalities (R \cite{ma_jun_2020_3757476, COVID19_2} + U \cite{al2020dataset, zhuang2019rdau}) and unseen modality (C \cite{bernal2015wm, jha2020kvasir, vazquez2017benchmark, tajbakhsh2015automated, silva2014toward} + D \cite{gutman2016skin, mendoncca2013ph}).}
    \label{tab:comparison_RU_other_metrics}
\end{table*}

\end{document}